%% file: main.tex
\documentclass[10pt,twocolumn,letterpaper]{article}

\usepackage{iccv}
\usepackage{times}
\usepackage{epsfig}
\usepackage{graphicx}
\usepackage{amsmath}
\usepackage{amssymb}

\usepackage{xcolor}
\usepackage[title]{appendix}
\usepackage{kotex}
\usepackage{breqn}
\usepackage{float}
\usepackage{algorithm}
\usepackage{algorithmic}
\usepackage{capt-of}
\usepackage{pdfpages}
\usepackage{enumerate}
\usepackage{footnote}
\usepackage{multirow}
\usepackage{makecell}

\usepackage{subcaption}
\usepackage{multirow}

\definecolor{greyblue}{rgb}{0.1,0.6,0.5}

\newcommand{\fref}[1]{Figure \ref{#1}}
\newcommand{\Fref}[1]{Figure \ref{#1}}
\newcommand{\tref}[1]{Table \ref{#1}}
\newcommand{\Tref}[1]{Table \ref{#1}}
\newcommand{\sref}[1]{Section \ref{#1}}

\newcommand{\Rd}{{\mathbb R}}
\newcommand\blfootnote[1]{%
  \begingroup
  \renewcommand\thefootnote{}\footnote{#1}%
  \addtocounter{footnote}{-1}%
  \endgroup
}

\usepackage[pagebackref=true,breaklinks=true,letterpaper=true,colorlinks,bookmarks=false]{hyperref}

\iccvfinalcopy % *** Uncomment this line for the final submission

\ificcvfinal\fi
\begin{document}

%%%%%%%%% TITLE
\title{Photorealistic Style Transfer via Wavelet Transforms}

\author{Jaejun Yoo\footnotemark[1]\\
% Clova AI Research, NAVER Corp.\\
% {\tt\small jaejun.yoo@navercorp.com}
\and
Youngjung Uh\footnotemark[1]\footnotetext{* indicates equal contribution}\\
% {\tt\small youngjung.uh@navercorp.com}
\and
Sanghyuk Chun\footnotemark[1]\\
% {\tt\small sanghyuk.c@navercorp.com}
\and
Byeongkyu Kang\\
% {\tt\small bk.kang@navercorp.com}
\and
Jung-Woo Ha\\
% {\tt\small jungwoo.ha@navercorp.com}
\and
Clova AI Research, NAVER Corp.
\and
{\tt\small \{jaejun.yoo,youngjung.uh,sanghyuk.c,bk.kang,jungwoo.ha\}@navercorp.com}
}

\twocolumn[{
\renewcommand\twocolumn[1][]{#1}
\maketitle
\begin{center}
    \vspace{-0.5 cm}
    \centering
    \includegraphics[width=1.0\textwidth]{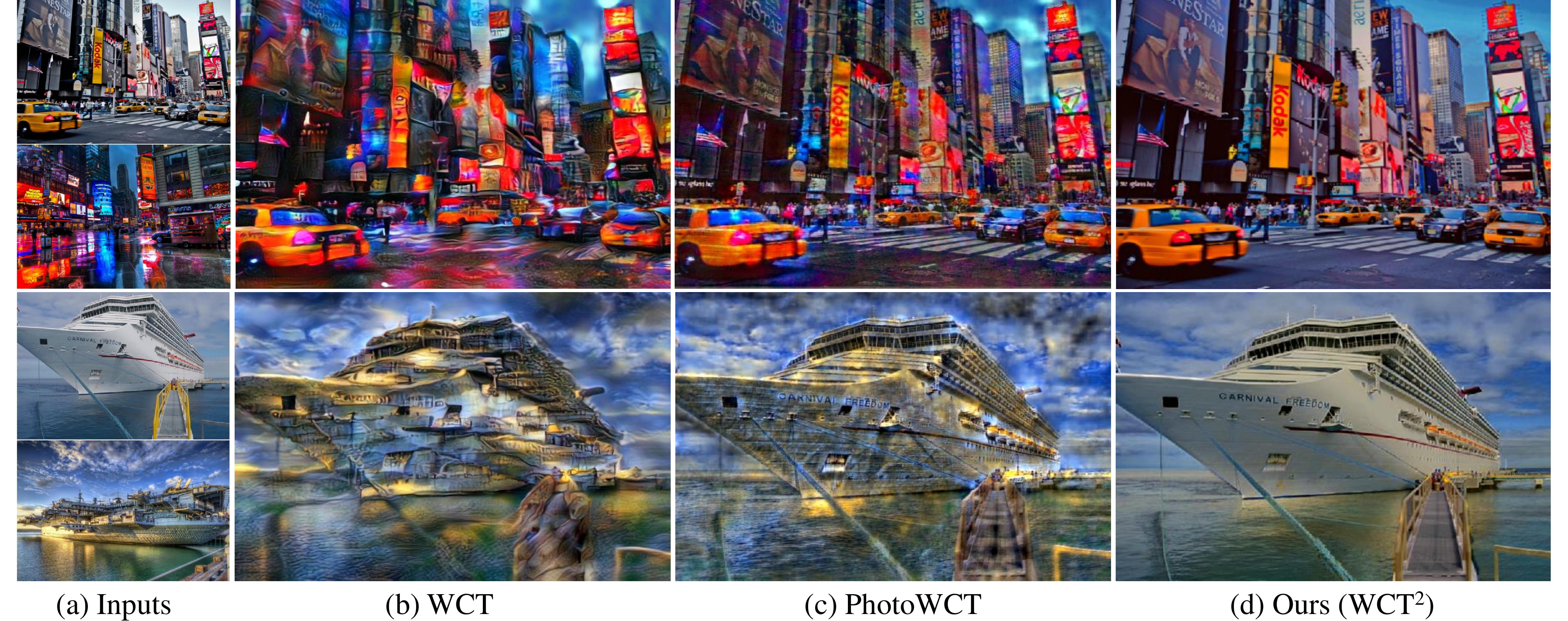}
    \captionof{figure}{Photorealistic stylization results. Given (a) an input pair (top: content, bottom: style), the results of (b) WCT \cite{wct}, (c) PhotoWCT \cite{photowct}, and (d) our model are shown. Every result is produced \textit{without} any post-processing.
    While WCT and PhotoWCT suffer from spatial distortions, our model successfully transfers the style and preserves the fine details. 
    }
    \label{teaser}
\end{center}
}]
\thispagestyle{empty}

%%%%%%%%% ABSTRACT
\begin{abstract}
\vspace{-0.5 cm}
    Recent style transfer models have provided promising artistic results. However, given a photograph as a reference style, existing methods are limited by spatial distortions or unrealistic artifacts, which should not happen in real photographs. We introduce a theoretically sound correction to the network architecture that remarkably enhances photorealism and faithfully transfers the style. The key ingredient of our method is wavelet transforms that naturally fits in deep networks. We propose a wavelet corrected transfer based on whitening and coloring transforms (WCT$^2$) that allows features to preserve their structural information and statistical properties of VGG feature space during stylization. \textbf{This is the first and the only end-to-end model that can stylize a 1024$\times$1024 resolution image in 4.7 seconds, giving a pleasing and photorealistic quality without any post-processing}. Last but not least, our model provides a stable video stylization without temporal constraints. Our code, generated images, and pre-trained models are all available at \href{https://github.com/ClovaAI/WCT2}{ClovaAI/WCT2}.
\end{abstract}

%%%%%%%%% BODY TEXT
\section{Introduction}
\begin{figure*}[!t]
\begin{center}
\includegraphics[width=1\textwidth]{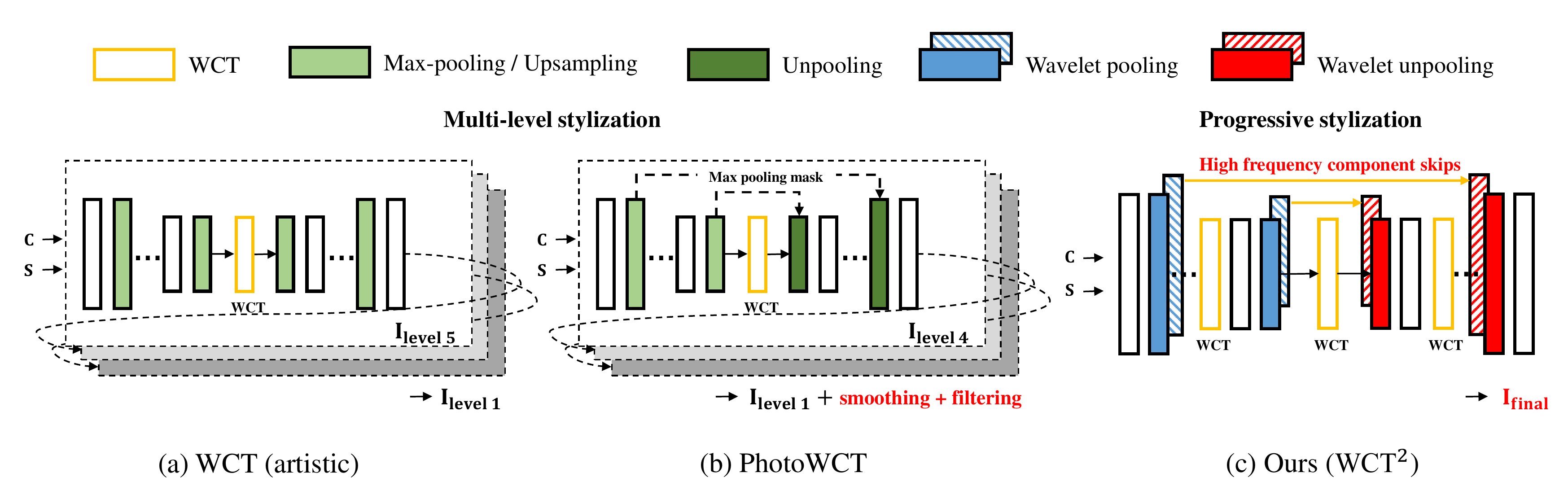}
\end{center}
\vspace{-0.5 cm}
   \caption{Comparison between previous style transfer models and our proposed model architecture (WCT$^2$). Unlike WCT \cite{wct} and PhotoWCT \cite{photowct} that use max-pooling and recursively stylize from coarse (level 5) to fine (level 1), WCT$^2$ replaces lossy operations (green) with wavelet pooling (blue) and unpooling (red), and employs the progressive stylization strategy in a single pass. Note that given the content (\textbf{c}) and style (\textbf{s}), WCT$^2$ outputs the final image ($I_{final}$) while the PhotoWCT output ($I_{level1}$) needs further post-processing steps (smoothing and filtering). %Only the low frequency features pass to the next layer and the high frequency components are directly skipped to the corresponding decoding layer. At the decoder, the components are aggregated by the wavelet unpooling (red). %A pair of encoder and decoder at same scale are shown. WCT is performed on the output of VGG {\tt convX\_1} layer followed by subsequent VGG layers and wavelet pooling. 
   }
\label{fig:netarch}
\vspace{-0.5 cm}
\end{figure*}
\blfootnote{* indicates equal contribution. Corresponding author: jaejun.yoo88@gmail.com}
Photorealistic style transfer has to satisfy contradictory objectives.
To be photorealistic, a model should apply the reference style on the scene without hurting the details of an image. 
In \fref{teaser}, for example, the general style (color and tone) of sky and sea should change, while the fine structures of the ship and the bridge remain intact. However, artistic style transfer methods (\eg, whitening and coloring transforms, WCT \cite{wct}) generally suffer from severe distortions due to their strong abstraction ability, which is not favored in the photorealistic stylization (\fref{teaser}b). (Please refer to our supplementary materials for more failure cases.)

Luan \etal \cite{luan2017deep} introduced a regularizer for photorealism on the traditional optimization-based method \cite{gatys2016image}. However, solving the optimization problem requires heavy computational costs, which limits their use in practice. 
To overcome this issue, Li \etal \cite{photowct} recently proposed a photorealistic variant of WCT (PhotoWCT) that replaced the upsampling components of the VGG decoder with unpooling.  By providing a max-pooling mask, PhotoWCT is designed to compensate for information loss during the encoding step and suppress the spatial distortion. Although their approach was valid, the introduction of the mask was not able to resolve the information loss that comes from the max-pooling of VGG network (\fref{teaser}c). To fix the remaining artifacts, they had to perform a series of post-processing steps, which require the original image to patch up the result. Not only do these post-processing steps retuire cumbersome computation and time but they entail another unfavorable blurry artifact and hyper-parameters to manually set. 

Instead of providing partial amendments, we address the fundamental problem by introducing a theoretically sound correction on the downsampling and upsampling operations. We propose a \textit{wavelet corrected transfer} based on whitening and coloring transforms (WCT$^2$) that substitutes the pooling and unpooling operations in the VGG encoder and decoder with wavelet pooling and unpooling. Our motivation is that the learned function by the network should have its inverse operation to enable exact signal recovery, and accordingly, photorealistic stylization. (We provide theoretical details in our supplementary materials.) It allows WCT$^2$ to fully reconstruct the signal without any post-processing steps, thanks to the favorable properties of wavelets providing minimal information loss~\cite{ye2018deep, yin2017tale}. The decomposed wavelet features provide interesting interpretations on the feature space as well, such as \textit{component-wise stylization} and \textit{why average pooling is known to give better stylization than max-pooling} (Section \ref{sec:analysis:wavelet}).

In addition, we propose \textit{progressive stylization} instead of following the multi-level strategy that is used in WCT \cite{wct} and PhotoWCT \cite{photowct}  (\fref{fig:netarch}). To maximize the stylization effect, WCT and PhotoWCT recursively transformed features in a multi-level manner from coarse to fine.
In contrast, we progressively transform features during a single pass. This allows two significant advantages over the others. 
First, our model is simple and efficient since we only have a single decoder during training as well as in the inference time. On the other hand, the multi-level strategy requires to train a decoder for each level without sharing parameters, which is inefficient in terms of the number of parameters and training procedure. This overhead remains in the inference time as well because the model requires to pass multiple encoder and decoder pairs to stylize an image. Second, by recursively encoding and decoding the signal with the lossy VGG networks, artifacts are amplified during the multi-level stylization. Because of wavelet operations and progressive stylization, our model does not have such a problem, and even more, it shows little error amplification when the multi-level strategy is employed (\fref{fig:analysis_moreSVD}). 

Our contributions are summarized as follows. We first show that the spatial distortions come from the network operations that cannot satisfy the reconstruction condition (\sref{sec:wct2}). By employing the wavelet corrected transfer and progressive stylization, we propose \textbf{the first end-to-end photorealistic style transfer model} that allows to remove the additional post-processing steps. Our model can \textbf{process a high resolution image (1024$\times$1024) in 4.7 seconds}, which is \textbf{830 times faster} than the state-of-the-art models, where PhotoWCT fails due to an out-of-memory issue and Deep Photo Style Transfer (DPST) \cite{luan2017deep} takes 3887.8 seconds. Our experimental results show quantitatively \textbf{better visual quality} in both SSIM and Gram loss (\fref{fig:statistics}), and qualitatively \textbf{being preferred by 62.21\%} in the user study (\Tref{table:user_study}). In addition, our model has \textbf{three times fewer} parameters than PhotoWCT and provides \textbf{temporally stable stylization} enabling video applications without additional constraints, such as optical flow (\fref{fig:video}). 

\section{Related Work}
\label{sec:related_work}

\paragraph{Style transfer.}
Starting from the seminal work of Gatys \etal \cite{gatys2016image}, many artistic style transfer studies have been proposed to synthesize stylized images through either iterative optimization \cite{gatys2017controlling}, finding dense correspondence \cite{analogy, sheng2018avatar, gu2018arbitrary} or manipulating features in pre-trained networks \cite{huang2017arbitrary,wct}. However, due to their powerful ability to abstract the features, they cannot be used in the photorealistic style transfer as they are.

Compared to artistic style transfer, photorealistic transfer has been overlooked. Classical methods mostly match the color and tone \cite{bae2006two,pitie2005n,reinhard2001color} of the images, which are restricted to specific usage. 
Luan \etal \cite{luan2017deep} proposed deep photo style transfer (DPST), which augments the neural style algorithm \cite{gatys2016image} with an additional photorealism regularization term and a semantic segmentation mask. 
However, DPST requires heavy computation to solve the regularized optimization problem. 

Recently, Li \etal \cite{photowct} proposed a photorealistic variant of WCT (PhotoWCT), which replaces the upsampling of the VGG decoder with unpooling. PhotoWCT showed that the spatial distortion could be relaxed by providing max-pooling masks to the decoder. Because the visual quality of the raw outputs of PhotoWCT was not satisfactory, the authors had to employ additional post-processing, such as \textit{smoothing} and \textit{filtering}. However, not only do these increase runtime exponentially to the image resolution, but blur final outputs.

Different from the existing methods, our method can preserve the fine structures of an image with little spatial distortion in an end-to-end manner, and thus removes the necessity of additional post-processing steps.  

\vspace{-0.2 cm}
\paragraph{Signal reconstruction using wavelets.}
Signal reconstruction using wavelets has been an extensive research topic in applied mathematics community due to its favorable characteristics such as proven convergence and compact representation of an arbitrary signal \cite{dong2017image, li2014wavelet}.
There have been several attempts to incorporate both classical signal processing and deep learning approaches, including feature reduction \cite{levinskis2013convolutional}, network compression \cite{gueguen2018faster, levinskis2013convolutional}, super-resolution \cite{bae2017beyond}, classification \cite{bruna2013invariant,fujieda2017wavelet, oyallon2017scaling, Ryu_2018_ECCV,williams2018wavelet} and image denoising \cite{kang2018deep}. Similarly, our approach augments wavelets as a component of the network architecture and provides an interpretable module that can enhance the photorealism of a style transfer model.

One closest related work \cite{williams2018wavelet} recently proposed to use wavelets as an alternative to traditional neighborhood pooling. However, their goal is to reduce feature dimensions by discarding the first-level sub-bands, while we exploits all sub-bands. In addition, we utilize both wavelet decomposition and reconstruction together to exactly recover the spatial information with minimal noise amplification. 
\section{WCT$^2$}
\label{sec:wct2}
To achieve photorealism, a model should recover the structural information of a given content image while it stylizes the image faithfully at the same time. 
To address this issue, we propose a Wavelet Corrected Transfer based on Whitening and Coloring Transforms, dubbed WCT$^2$. 
More specifically, we handle the first objective by employing wavelet pooling and unpooling, which preserve information of the content to the transfer network. We use progressive stylization within a single forward pass to tackle the second issue.

\subsection{Wavelet corrected transfer}
\label{subsec:wav_corr_transfer}

\paragraph{Haar wavelet pooling and unpooling.} We first explain the main components of our model using Haar wavelets, which we call wavelet pooling and unpooling. Haar wavelet pooling has four kernels, $\{$LL$^\top$ LH$^\top$ HL$^\top$ HH$^\top\}$, where the low (L) and high (H) pass filters are 
\begin{eqnarray} 
\label{eq:wavLHbasis}
L^\top =\frac{1}{\sqrt{2}}\left[
      \begin{array}{cc}
        1  &   1     \\
      \end{array}
    \right], \quad
H^\top =\frac{1}{\sqrt{2}}\left[
      \begin{array}{cc}
        -1  &   1
      \end{array}
    \right].
\end{eqnarray}
Thus, unlike common pooling operations, the output of the Haar wavelet pooling has four channels. 
Here, the low-pass filter captures smooth surface and texture while the high-pass filters extract vertical, horizontal, and diagonal edge-like information. For simplicity, we denote the output of each kernel as LL, LH, HL, and HH, respectively.

One important property of our wavelet pooling is that the original signal can be exactly reconstructed by mirroring its operation; \ie, wavelet unpooling. In detail, wavelet unpooling fully recovers the original signal by performing a component-wise transposed-convolution, followed by a summation. (Please see our supplementary materials for more details.) Thanks to this favorable property, our proposed model can stylize an image with minimal information loss and noise amplification. On the other hand, max-pooling does not have its exact inverse so that the encoder-decoder structured networks used in the WCT \cite{wct} and PhotoWCT \cite{photowct} cannot fully restore the signal. 

Note that Haar wavelet pooling and unpooling is not the only operation which can fully recover the original signal. However, we choose Haar wavelet because it splits the original signal into channels that capture different components, which leads to better stylization. 

\paragraph{Model architecture.}
To fully utilize the encoded information, we replace every max-pooling and unpooling of PhotoWCT with the wavelet pooling and unpooling (\fref{fig:netarch}). 
Specifically, we use the ImageNet \cite{deng2009imagenet} pre-trained VGG-19 network \cite{vgg} from {\tt conv1\_1} layer to {\tt conv4\_1} layer as the encoder. The max-pooling layers are replaced with wavelet pooling where the high frequency components (LH, HL, HH) are skipped to the decoder directly. Thus, only the low frequency component (LL) is passed to the next encoding layer. The decoder has a mirror structure of the encoder, and the wavelet unpooling aggregates the components. (Please refer to our supplementary materials for more details about the proposed network architecture)

\subsection{Stylization}
\label{subsec:stylization}

\paragraph{Whitening and coloring transforms (WCT).}
Since our method is built upon WCT \cite{wct}\footnote{Note that our wavelet corrected transfer is not limited to a specific stylization method. Here, we simply used WCT for better stylization. For example, at the expanse of slight image quality degradation, our model can be integrated with AdaIN \cite{huang2017arbitrary}, which further accelerates the model by avoiding SVD calculation. (Please refer to the supplemntary materials.)}, we first recap WCT briefly.
WCT can perform style transfer with arbitrary styles by directly matching the correlation between content and style in the VGG feature domain. It projects the content features to the eigenspace of style features by calculating singular value decomposition (SVD). The final stylized image is obtained by feeding the transferred features into the decoder. To provide better artistic style transfer, the authors employed a multi-level stylization framework by applying WCT to multiple encoder-decoder pairs (\fref{fig:netarch}b).

\paragraph{Progressive stylization.}
Instead of using the multi-level stylization used in WCT and PhotoWCT, we progressively transform features within a single forward-pass as illustrated in \fref{fig:netarch}. 
We sequentially apply WCT at each scale (\texttt{conv1\_X}, \texttt{conv2_X}, \texttt{conv3\_X} and \texttt{conv4\_X}) within a single encoder-decoder network. 
Note that the number of SVD computations of our model remains the same. We can add more WCTs on skip-connections and decoding layers to further strengthen the stylizing effect at the cost of time consumption. This will be covered in more detail in Section \ref{subsec:progressive}. 
There are several advantages of our proposed progressive stylization against the multi-level one. First, the multi-level strategy requires to train a decoder for each level without sharing parameters, which is inefficient. On the other hand, our training procedure is simple because we only have a single pair of encoder and decoder, which is advantageous in the inference time as well. Second, recursively encoding and decoding the signal with VGG network architecture amplifies errors causing unrealistic artifacts in the output. In the later section, we show that our proposed progressive stylization technique suffers less from the error amplification than the multi-level strategy.

\section{Analysis}
\label{sec:analysis}
\begin{figure}[t!]
    \centering
    \begin{subfigure}[t]{0.45\linewidth}
        \centering
        \includegraphics[width=\linewidth]{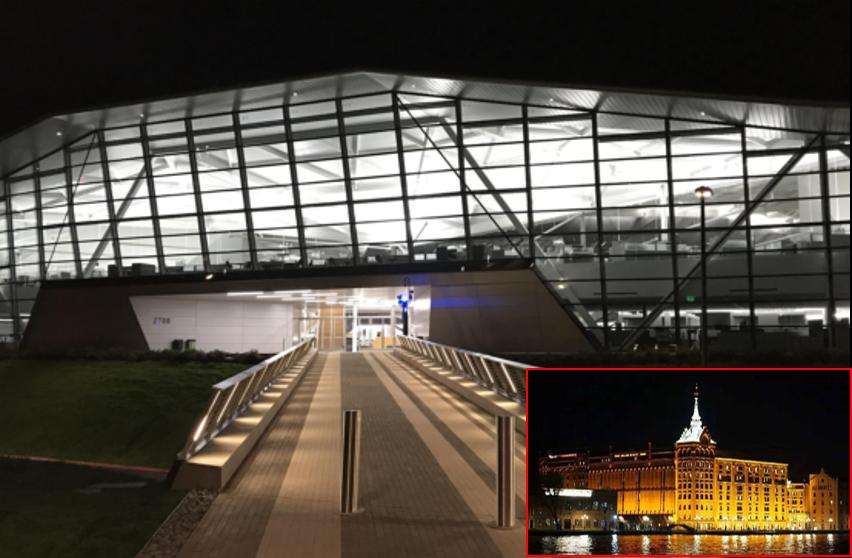}
        \caption{Input}
    \end{subfigure}
    \begin{subfigure}[t]{0.45\linewidth}
        \centering
        \includegraphics[width=\linewidth]{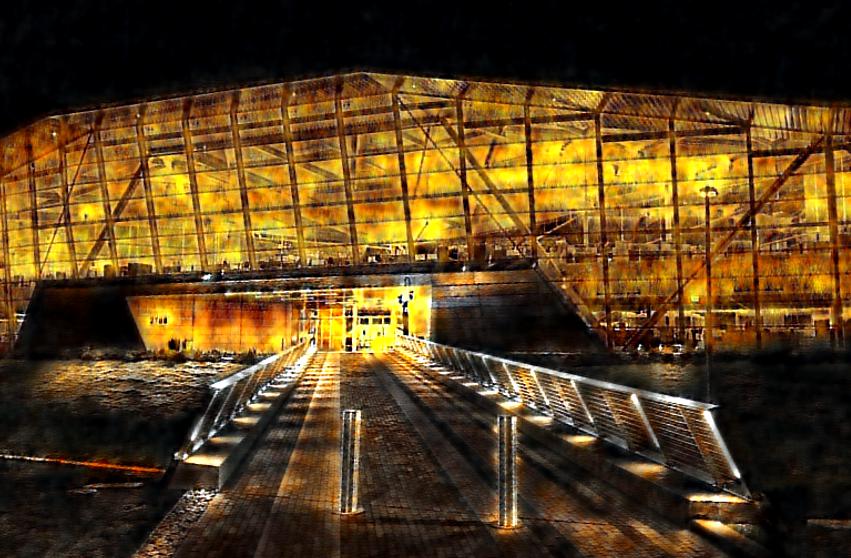}
        \caption{PhotoWCT \cite{photowct}}
        \label{fig:analysis_comp:photowct}
    \end{subfigure} \\
    \begin{subfigure}[t]{0.45\linewidth}
        \centering
        \includegraphics[width=\linewidth]{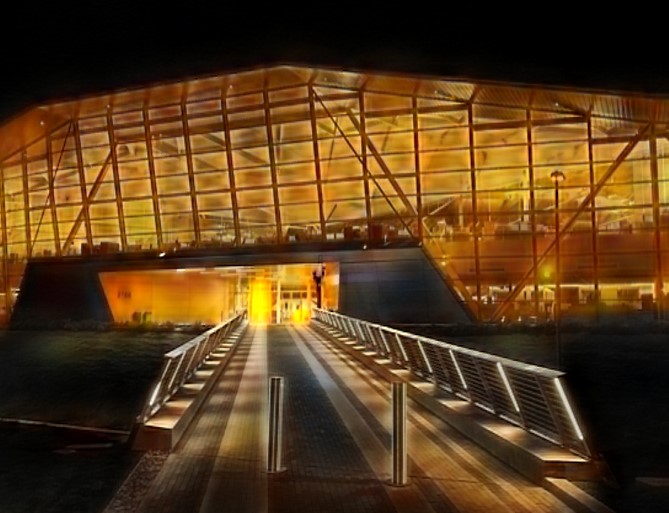}
        \caption{Ours}
        \label{fig:analysis_comp:ours}
    \end{subfigure}
    \begin{subfigure}[t]{0.45\linewidth}
        \centering
        \includegraphics[width=\linewidth]{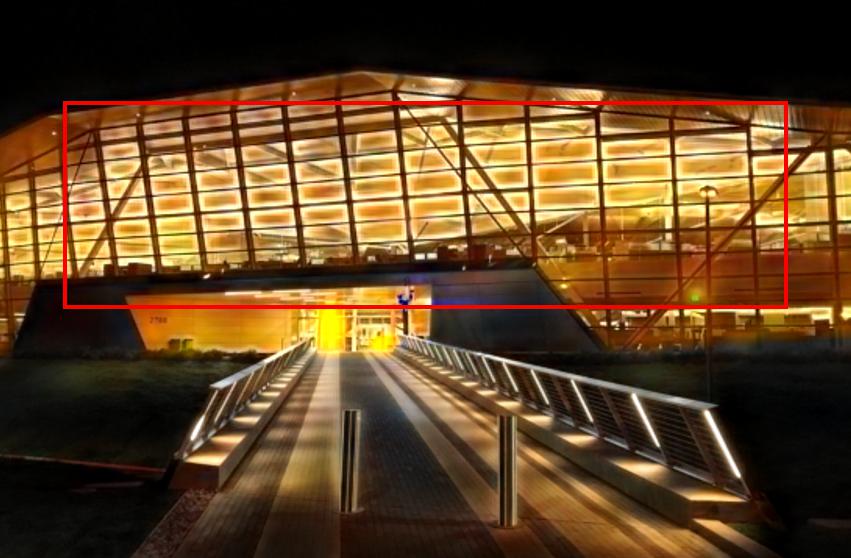}
        \caption{Ours (LL only)}%without edge stylization}
        \label{fig:analysis_comp:basic_e}
    \end{subfigure}\\
\caption{Comparison between max-pooling and wavelet pooling. Given (a) an input pair (inset: style), we compare the results of (b) PhotoWCT without post-processing, (c) ours  and (d) ours but stylize only the LL component. Note that the edges are left unstylized (inside the red box).}
\label{fig:analysis_comp}
\vspace{-0.5 cm}
\end{figure}
\subsection{Wavelet pooling}
\label{sec:analysis:wavelet}
We first examine the effects of using the wavelet pooling instead of max-pooling. As shown in  \fref{fig:analysis_comp:photowct} and \ref{fig:analysis_comp:ours}, PhotoWCT suffers from the loss of spatial information by max-pooling while ours preserves fine details. We recall that the low frequency component captures smooth surface and texture while the high frequency components detect edges. This enables our model to separately control the stylization effect by choosing a component. More specifically, it implies that applying WCT to LL of the encoder affects overall texture or surface while applying WCT to the high frequency components (\ie, LH, HL, HH) stylize edges. Indeed, when we stylize all components (\fref{fig:analysis_comp:ours}), our model transfers the given style to the entire building. In contrast, if we do not perform WCT on the high frequency components, the boundaries of windows remain unchanged (\fref{fig:analysis_comp:basic_e}). 

Note that using only the LL component of our wavelet pooling is equivalent to using the average pooling. Interestingly, since Gatys \etal \cite{gatys2016image}, many studies have consistently reported that replacing the max-pooling operation with average pooling yields slightly more appealing results. This can be explained in our framework that the model is using only the partial information (LL) of the wavelet decomposed feature domain. In addition, because each frequency component of the content feature is transformed into its corresponding component of style feature, we can obtain a similar advantage as we do by using spatial correspondences. 

\subsection{Ablation study}
To show that our model indeed benefits from the wavelet pooling, we compare the stylization results using other pooling variants. We unpool the features similar to the way we do for the wavelet unpooling; \ie, transposed-convolution and summation. 

\begin{figure}[t!]
    \centering
    \begin{subfigure}[t]{0.45\linewidth}
        \centering
        \includegraphics[width=\linewidth]{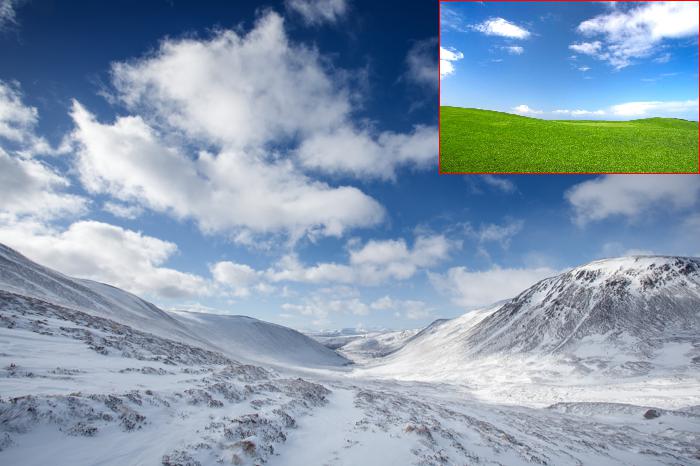}
        \caption{Input}
    \end{subfigure}
    \begin{subfigure}[t]{0.45\linewidth}
        \centering
        \includegraphics[width=\linewidth]{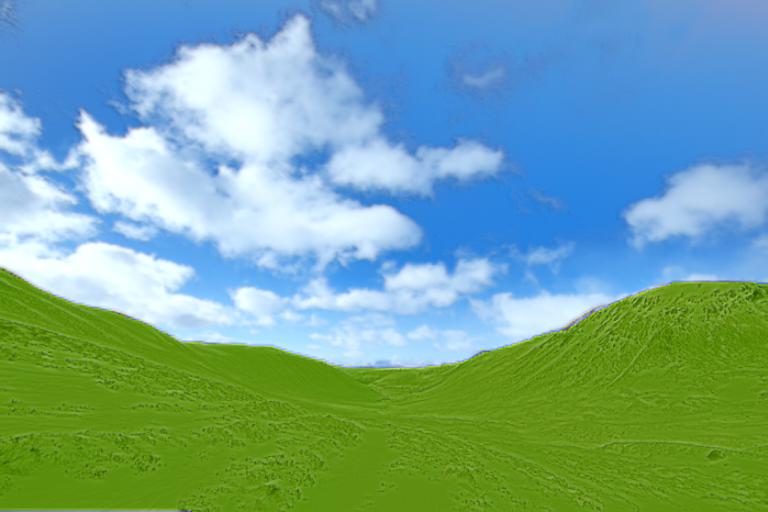}
        \caption{Split}
        \label{fig:analysis_pool:split}
    \end{subfigure}\\
    \begin{subfigure}[t]{0.45\linewidth}
        \centering
        \includegraphics[width=\linewidth]{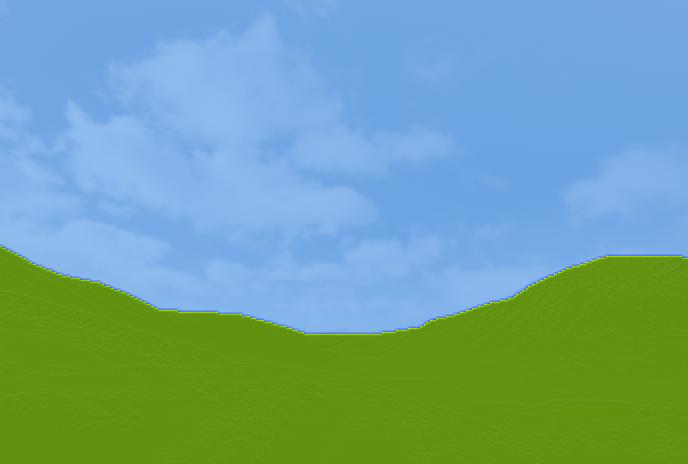}
        \caption{Learnable}
        \label{fig:analysis_pool:learnable}
    \end{subfigure}
    \begin{subfigure}[t]{0.45\linewidth}
        \centering
        \includegraphics[width=\linewidth]{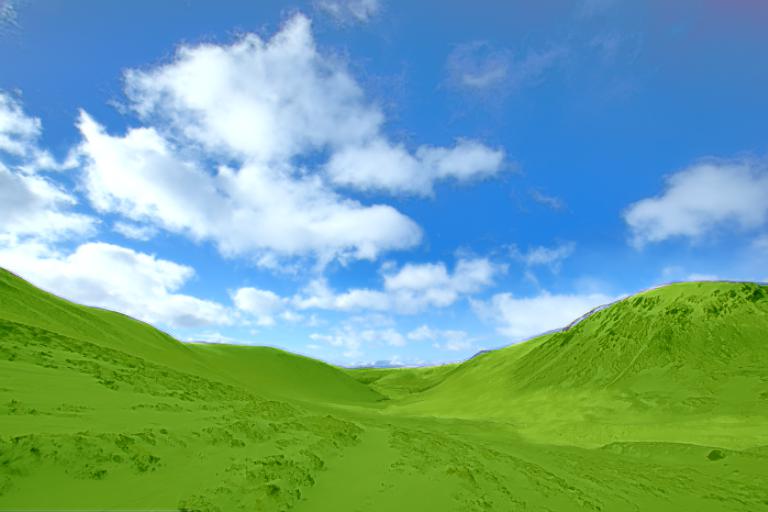}
        \caption{Ours (WCT$^2$)}
    \end{subfigure} \\
\caption{Ablation study on pooling methods. While split and learnable poolings suffer from the lack of representation power or altered feature statistics, wavelet pooling benefits from the compact representation of wavelets and retains the original VGG feature property intact.}
\label{fig:analysis_pool}
\end{figure}

\paragraph{Split pooling.}
Split pooling has $2\times 2$ filters with fixed weights, \ie, [1 0 0 0], [0 1 0 0], [0 0 1 0], and [0 0 0 1]. Split pooling has a similar property to wavelet pooling in that it can carry whole information. Here, we can see a similar effect but degradation in fine details, \eg, the grass (\fref{fig:analysis_pool:split}). We suspect that this is due to the lack of representation power.  

\paragraph{Learnable pooling.}
Learnable pooling is a trainable \texttt{conv} layer with a stride of two. As shown in \fref{fig:analysis_pool:learnable}, it does not preserve the content nor faithfully transfer the style. We suppose that this happens because the learnable pooling brings too much flexibility to the network. This ruins the original feature properties of VGG networks \cite{vgg}, which is known to be good at extracting styles \cite{gatys2016image}. 

\begin{figure}[t!]
    \centering
    \begin{subfigure}[t]{0.45\linewidth}
        \centering
        \includegraphics[width=\linewidth]{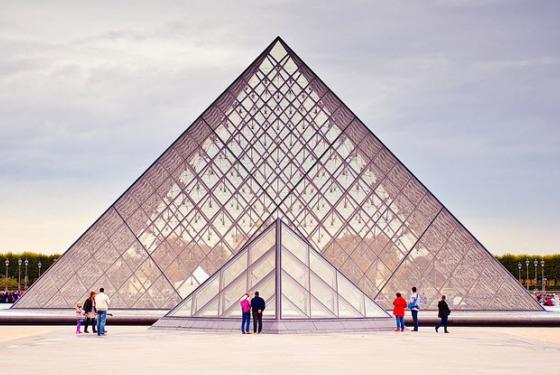}
        \caption{Content}
    \end{subfigure}
    \begin{subfigure}[t]{0.45\linewidth}
        \centering
        \includegraphics[width=\linewidth]{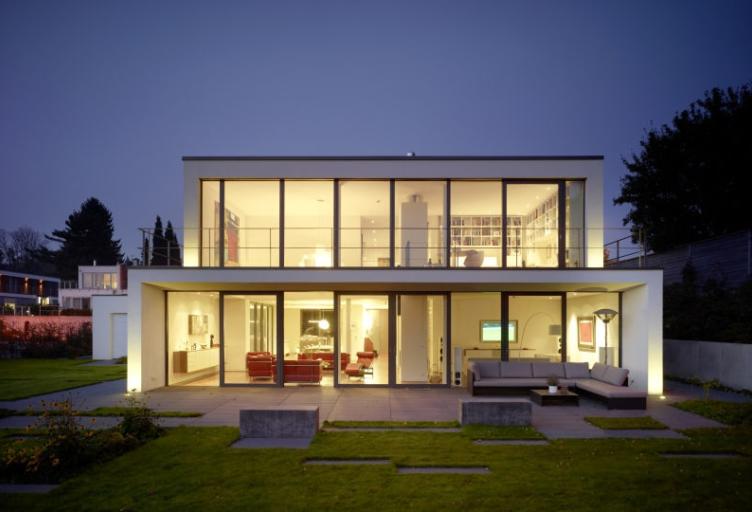}
        \caption{Style}
    \end{subfigure}\\
    \begin{subfigure}[t]{0.45\linewidth}
        \centering
        \includegraphics[width=\linewidth]{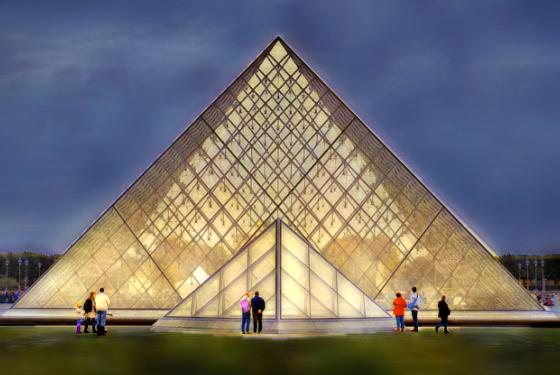}
        \caption{Sum-version}
    \end{subfigure}
    % \begin{subfigure}[t]{0.45\linewidth}
    %     \centering
    %     \includegraphics[width=\linewidth]{./images/analysis_unpool/cat4_in00_e1234_s_d_level_4_cropped.jpg}
    %     \caption{}
    % \end{subfigure}
    \begin{subfigure}[t]{0.45\linewidth}
        \centering
        \includegraphics[width=\linewidth]{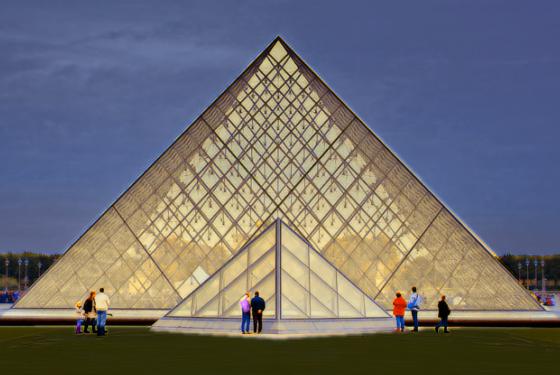}
        \caption{Concatenated-version}
    \end{subfigure}\\
\caption{Variation of the unpooling options (Section \ref{subsec:unpool}).}
\label{fig:analysis_unpool}
\end{figure}

\begin{figure}[t!]
    \centering
    \begin{subfigure}[t]{0.45\linewidth}
        \centering
        \includegraphics[width=\linewidth]{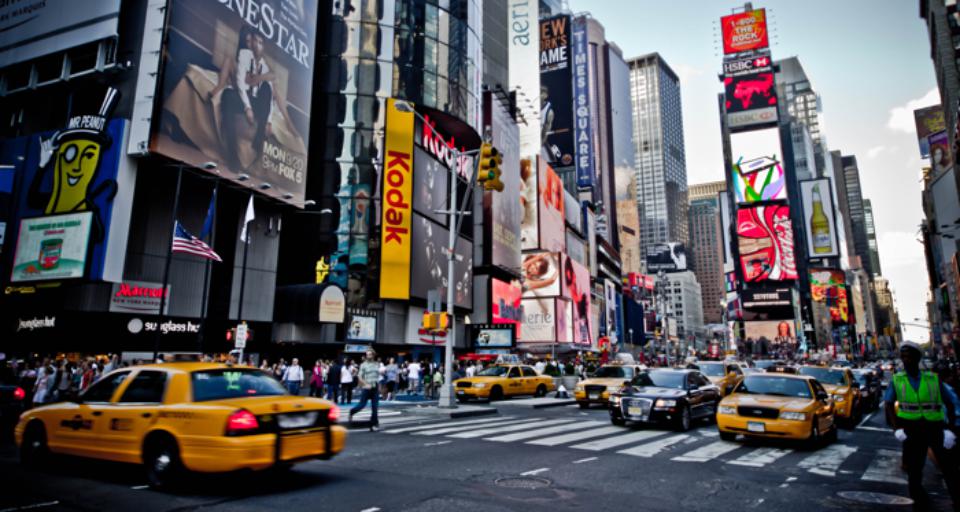}
        \caption{Input}
    \end{subfigure}
    \begin{subfigure}[t]{0.45\linewidth}
        \centering
        \includegraphics[width=\linewidth]{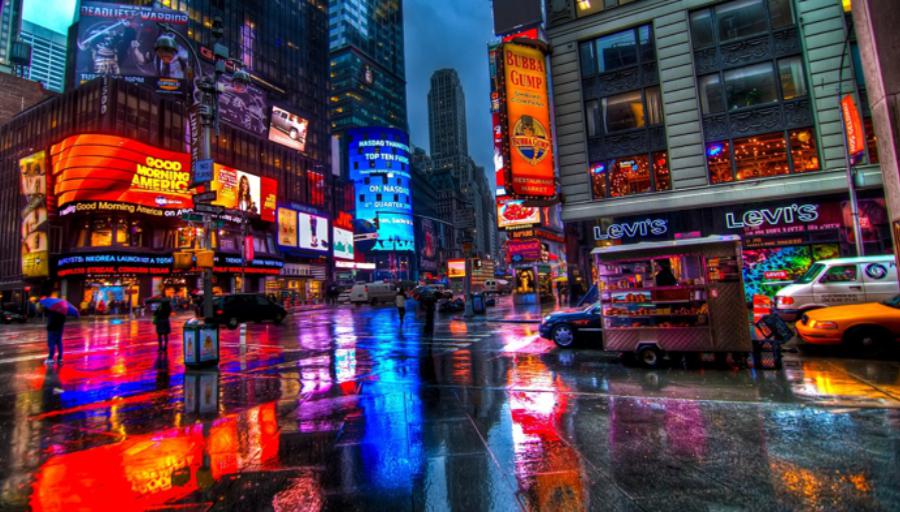}
        \caption{Style}
    \end{subfigure}\\
    \begin{subfigure}[t]{0.45\linewidth}
        \centering
        \includegraphics[width=\linewidth]{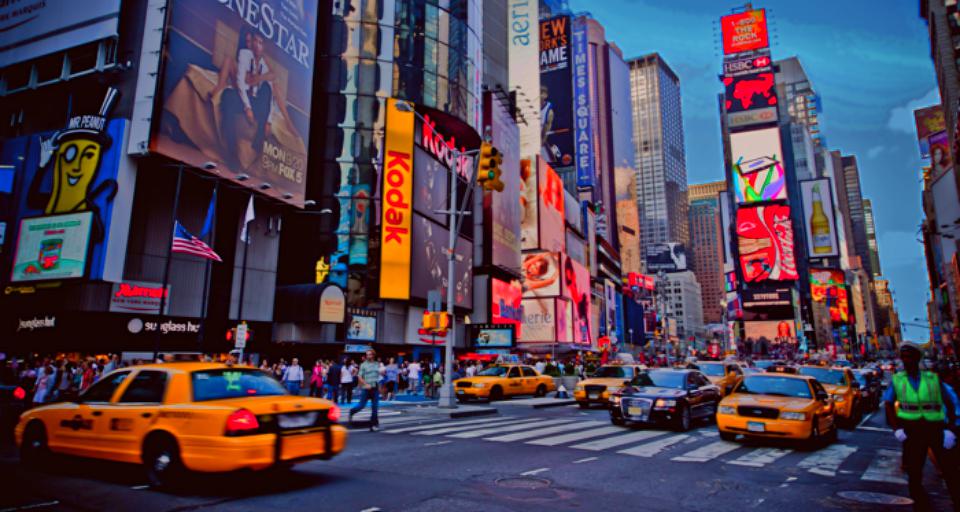}
        \caption{Single-pass (WCT$^2$)}
    \end{subfigure}
    \begin{subfigure}[t]{0.45\linewidth}
        \centering
        \includegraphics[width=\linewidth]{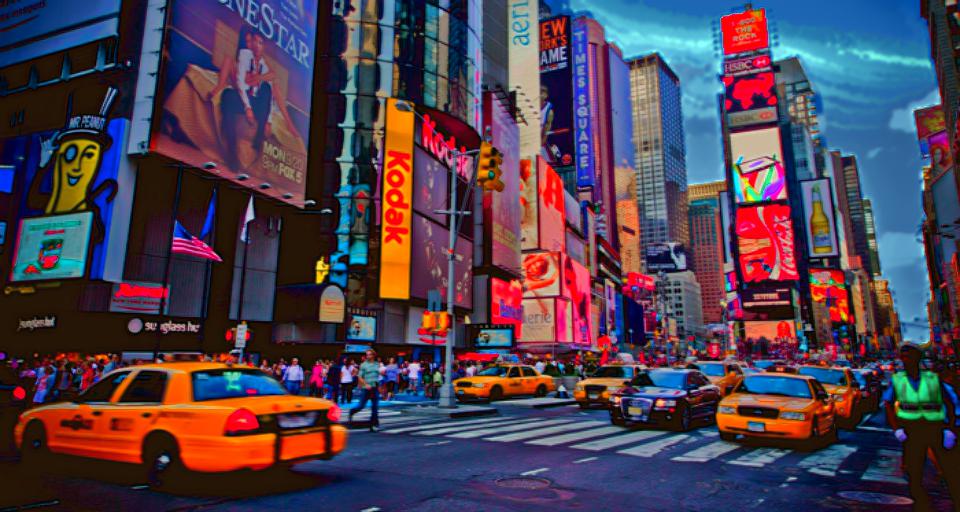}
        \caption{Multi-level}
        \label{fig:analysis_moreSVD:multi}
    \end{subfigure}\\
    % \begin{subfigure}[t]{0.45\linewidth}
    %     \centering
    %     \includegraphics[width=\linewidth]{./images/analysis_moreSVD/decoder.jpg}
    %     \caption{Single + decoder}
    % \end{subfigure} 
    % \begin{subfigure}[t]{0.45\linewidth}
    %     \centering
    %     \includegraphics[width=\linewidth]{./images/analysis_moreSVD/multi_decoder.jpg}
    %     \caption{Multi-level + decoder}
    % \end{subfigure} \\
\caption{Stylization strength with more whitening and coloring transforms. Single-pass is our baseline (Section \ref{subsec:progressive}).}
\label{fig:analysis_moreSVD}
\end{figure}

\input{./comparison_fig.tex}

\subsection{Unpooling options}
\label{subsec:unpool}
To achieve better reconstruction, we adopted concatenation instead of summation for unpooling, similar to U-Net structure \cite{framingunet, ronneberger2015u, yoo2018mathematical}. This enables the network to learn the weighted sum of components at the expense of interpretability and theoretical correctness. Specifically, our wavelet unpooling now performs channel-wise concatenation of four feature components from the corresponding scale plus feature output before the wavelet pooling.
Therefore, the number of parameters increases at the \texttt{conv} layer that comes right after the wavelet unpooling. This increases the total number of parameters to be $1.80\times$ of the sum-version of WCT$^2$ while PhotoWCT has $3.06\times$ parameters. As shown in \fref{fig:analysis_unpool}, spatial details are further improved.  The sum-version generally produces a more stylized output while the concatenated-version produces a clearer image. (Please see our supplementary materials for more results.)

\begin{figure*}[ht!]
    \centering
    \includegraphics[width=\linewidth]{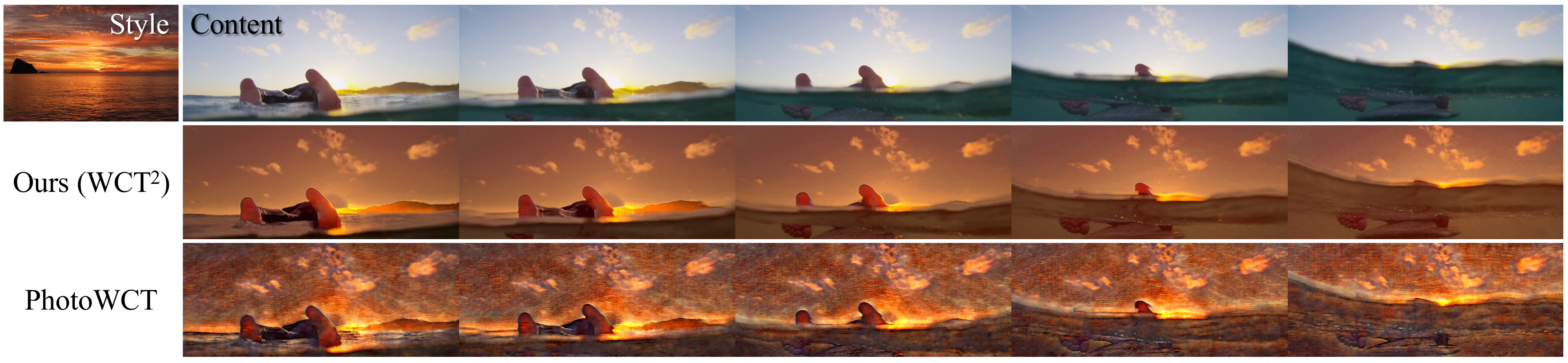}
\caption{Photorealistic video stylization results (from day-to-sunset). Given a style image and video frames (top), we show the results by (middle) WCT$^2$ and (bottom) PhotoWCT \cite{photowct} without providing semantic segmentation maps and post-processing steps. 
}
\label{fig:video}
\vspace{-0.5 cm}
\end{figure*}
\subsection{Progressive vs. multi-level strategy}
\label{subsec:progressive}
Owing to the exact reconstruction property of wavelet pooling, our model can adopt the multi-level strategy to increase the contrast in the transferred style with minimal noise amplification. As shown in \fref{fig:analysis_moreSVD:multi}, adopting the multi-level approach in addition to WCT$^2$ leads to more vivid results. Note that it maintains photorealism while PhotoWCT produces spotty artifacts due to the noise amplification (\fref{fig:photorealistic_comparison:photowct}). In addition, performing progressive stylization at the decoder as well, namely \texttt{conv3\_2}, \texttt{conv2\_2}, and \texttt{conv1\_2}, further increases stylization effect. Still, strengthening the style comes at the cost of photorealism and multiple SVD computations. (Please refer to the supplementary materials for more results)

\section{Experimental results}
\label{sec:experiment}

In this section, we show that our simple modification can remarkably enhance the performance of photorealistic style transfer. Here, every result is reported based on the concatenated version of our model. For a fair comparison and time-efficiency, we only perform whitening and coloring on LL components (\eg, \texttt{convX\_1} outputs of the encoder) progressively. Thus, the number of whitening and coloring procedure of our model matches with PhotoWCT. 

\subsection{Implementation details}
We use the encoder-decoder architecture with fixed VGG encoder weights. The decoder is trained
on Microsoft COCO dataset \cite{coco}, minimizing the L2 reconstruction loss and the additional feature Gram matching loss with the encoder. The training is done with NAVER Smart Machine Learning (NSML) platform~\cite{nsml2}. We use ADAM optimizer \cite{kingma2014adam} with a learning rate of $10^{-3}$.
Finally, similar to PhotoWCT and DPST, we utilize the semantic map to match the styles of corresponding image regions. 
The code and pre-trained models are available at \href{https://github.com/ClovaAI/WCT2}{ClovaAI/WCT2}.

\subsection{Qualitative evaluation}
\Fref{fig:photorealistic_comparison} shows the results of DPST, PhotoWCT and ours (WCT$^2$). 
DPST often generates ``staircasing" or ``cartoon" artifacts \cite{chan2005image} with an unrealistic color transfer, which severely hurts photorealism (\fref{fig:photorealistic_comparison:dpst}). PhotoWCT better reconstructs the details of the content image, while it shows spotty artifacts over entire images (\fref{fig:photorealistic_comparison:photowct}). 
Such artifacts can be removed by employing additional post-processing steps (\fref{fig:photorealistic_comparison:photowct_full}). However, it has three disadvantages that 1) optimization is slow, 2) hyper-parameters should be carefully tuned to trade-off between smoothness and fine details, and 3) the final image becomes blurry at the expense of removed artifacts. In contrast, our proposed method shows fewer artifacts while faithfully transferring the reference styles (\fref{fig:photorealistic_comparison:ours}). Note that we do not apply any post-processing after the network output.

\paragraph{Video stylization.}
To emphasize consistent feature representation of the wavelet pooling and unpooling, we separately stylize every video frame to target style without any semantic segmentation. \Fref{fig:video} shows that WCT$^2$ performs stable video style transfer without any temporal consistency regularization such as optical flow. On the other hand, PhotoWCT generates spotty and varying artifacts over frames, which harms the photorealism. (The link to the full video can be found in \href{https://github.com/ClovaAI/WCT2}{our project page}.)

\subsection{Quantitative evaluation}
\label{subsec:quant}
\begin{figure}[t!]
    \centering
    \includegraphics[width=1.0\linewidth]{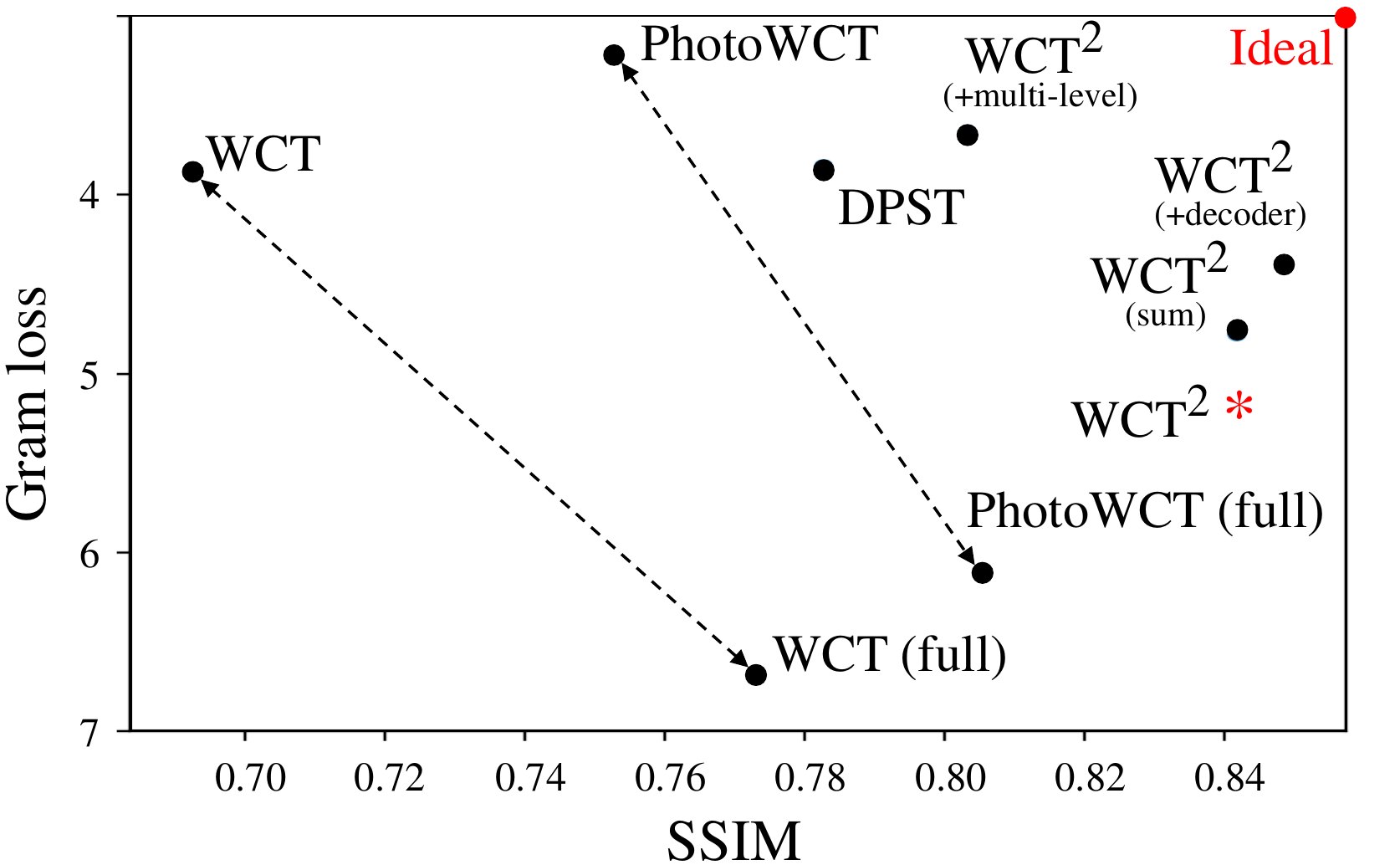}
\caption{SSIM index (higher is better) versus Style loss (lower is better). Ideal case is the top-right corner (red dot). Dashed lines depict the gap before and after the post-processing steps, \ie, smoothing. The baseline WCT$^2$ with concatenation is denoted by the red asterisk.}
\label{fig:statistics}
\end{figure}

\paragraph{Statistics.}
To measure photorealism, we employ two surrogate metrics for spatial recovery and stylization. We calculate the structural similarity (SSIM) index between edge responses \cite{xie2015holistically} of original contents and stylized images. Following WCT \cite{wct}, we report the covariance matrix difference (VGG style loss \cite{gatys2016image}) between the style image and the outputs of each model. 
\Fref{fig:statistics} shows SSIM (X-axis) against style loss (Y-axis). Our proposed model (WCT$^2$) remarkably outperforms other methods. 

Note that WCT$^2$ and its variants are located at the top-right corner, superior to PhotoWCT (full) and DPST that perform post-processing. Here, DPST has strength on the Gram-based score because it directly optimizes the style loss. 
Still, it is far from being practical due to its heavy optimization procedure (\tref{table:runtime}). As expected, when we compare the results of our variants, the multi-level approach adds more style (smaller Gram-based loss) at the expense of noise amplification (larger SSIM index), which is even better than the direct optimization (DPST). 

In addition, by comparing the gap before and after the post-processing steps (\fref{fig:statistics}, dashed lines), we can clearly see that \textbf{the final visual qualities of PhotoWCT majorly come from the powerful post-processing, especially the smoothing step, not the network itself}. The original WCT with smoothing already shows a comparable result to that of PhotoWCT. This demonstrates that the unpooling substitution of PhotoWCT did not fully address the information loss but the post-processing did. 

\paragraph{Runtime \& memory.}
\tref{table:runtime} shows the runtime comparison of DPST, PhotoWCT, and WCT$^2$. For PhotoWCT, we separately measured WCT and post-processing steps to better compare with ours. 
The reported runtime for each model is an average of ten-rounds run on a single NVIDIA P40 GPU. As expected, our model inherits the computational time of the original WCT.
Note that the concatenation in unpooling hardly increases the runtime of WCT$^2$. 
Because our model can remove the cumbersome post-processing steps, WCT$^2$ can deal with high resolution images, such as $1024\times1024$, maintaining a high quality of photorealistic style transfer. Compared to DPST, WCT$^2$ achieves a speed-up of about \textbf{830 times} in runtime. 
In addition, WCT$^2$ uses only $51\%$ GPU-memory of PhotoWCT, which uses a multi-level stylization requiring four encoder-decoder models (\sref{subsec:progressive}) since WCT$^2$ progressively stylize an image using a single encoder-decoder. 

\begin{table}[t!]
\centering
\small{
\begin{tabular}{cccccc}
 & & PhotoWCT (full) & \\
Image Size & DPST & (WCT $+$ post) & Ours \\ \hline
% $128 \times 128$   & $135.2$    & $2.7 + 2.5$  & \textbf{2.5} \\
$256 \times 256$   & $306.9$  & $3.2 + 9.2$ & \textbf{3.2} \\
% $384 \times 384$   & $607.5$  & $3.6 + 21.0$  & $3.6$ \\
$512 \times 512$   & $1020.7$ & $3.6 + 40.2$ & \textbf{3.8} \\
% $640 \times 640$   & $1553.7$ & $3.7 + 74.3$ & $3.9$ \\
% $768 \times 768$   & $2264.0$  & $3.8 + 101.8$         & \textbf{4.2} \\
$896 \times 896$   & $2988.6$ & $3.8 + \text{OOM}$ & \textbf{4.4} \\
$1024 \times 1024$ & $3887.8$ & $3.9+\text{OOM}$ & \textbf{4.7} \\  \hline
% Image Size & (phase 1 $+$ phase 2) & (WCT $+$ post) & Ours \\ \hline
% $128 \times 128$   & $37.4 + 97.8$    & $2.7 + 2.5$  & $2.5$ \\
% $256 \times 256$   & $151.6 + 155.3$  & $3.2 + 9.2$ & $3.2$ \\
% % $384 \times 384$   & $339.6 + 267.9$  & $3.6 + 21.0$  & $3.6$ \\
% $512 \times 512$   & $603.1 + 417.6$ & $3.6 + 40.2$ & $3.8$ \\
% % $640 \times 640$   & $939.7 + 614.0$ & $3.7 + 74.3$ & $3.9$ \\
% % $768 \times 768$   & $1385.9 + 878.1$  & $3.8 + 101.8$         & $4.2$ \\
% % $896 \times 896$   & $1843.8 + 1144.8$ & $3.8 + \text{OOM}$ & $4.4$ \\
% $1024 \times 1024$ & $2413.8 + 1474.0$ & $3.9 + \text{OOM}$ & $4.7$ \\  \hline
\end{tabular}
\caption{Runtime comparision of DPST, PhotoWCT (full) and ours in seconds. 
OOM denotes out-of-memory error. 
}
\label{table:runtime}
}
\end{table}
\begin{table}[t!]
\begin{tabular}{l|ccc}
\hline
                 & DPSP    & PhotoWCT (full) & Ours    \\ \hline
Fewest artifacts & 21.34\% & 9.33\%  & \textbf{69.33\%} \\
Best stylization & 30.49\% & 12.74\%  & \textbf{56.77\%} \\
Most preferred   & 24.63\% & 11.16\%  & \textbf{62.21\%} \\ \hline
\end{tabular}
\caption{User study results. The percentage indicates the preferred model outputs out of 1640 responses. 
}
\label{table:user_study}
\end{table}

\paragraph{User study.}
We conducted a user study to further evaluate the methods in terms of fewer artifacts, faithfulness to the style input, and overall qualities. Our benchmark dataset consists of content and style pairs provided by Luan \etal \cite{luan2017deep}. Total 40 sets of questions were asked to 41 subjects, in which subjects had to choose one among three stylized images from each model. The results are shown in random order with content and style images. Table \ref{table:user_study} shows the percentage of model outputs that are chosen out of 1640 ($=40\times41$) responses. Our method is preferred by human subjects against the other state-of-the-art methods by a large margin in all aspects. Note that we compare our results with PhotoWCT (full) that applies two post-processing steps proposed by the authors \cite{photowct} while we do not perform any post-processing for WCT$^2$. (Please see our supplementary materials for the images that are used for the user study)

\paragraph{Failure cases.}
Many photorealistic models \cite{luan2017deep,photowct} including ours require the semantic map and its accuracy is important for better stylization results. In  fact,  this  phenomenon is more prominent in our model because WCT$^2$ retains every fine detail unlike the others (Supplementary materials). The effect of misaligned map is visible in our result while PhotoWCT smooths it out unintentionally. Resolving the dependency on the semantic label map is an interesting future research direction.

\section{Conclusion}
\label{sec:conclusion}
In this paper, we proposed the first end-to-end photorealistic style transfer method, WCT$^2$. Based on the theoretical analysis, we specifically designed our model to satisfy the reconstruction condition. The exact recovery of the wavelet transforms allows our model to preserve structural information while providing stable stylization without any constraints. By employing progressive stylization, we achieved better results with less noise amplification. Compared to the other state-of-the-arts, our analysis and experimental results showed that WCT$^2$ is scalable, lighter, faster and achieves better photorealism quantitatively and qualitatively. Our results were preferred by human subjects in every aspect with a significant margin. Future study will include removing the necessity of semantic labels, which should be accurate for a flawless result so far.

\section{Acknowledgement}  
We would like to thank Clova AI Research team and advisory members, especially Yunjey Choi, Sangdoo Yun, Dongyoon Han, Youngjoon Yoo, and Jun-Yan Zhu for their helpful feedback and discussion. 
{\small
\bibliographystyle{ieee_fullname}
\bibliography{egbib}
}
% \newpage

\appendix
\onecolumn
\section{Frame-based signal reconstruction} 
\label{appendix}
Our proposed model WCT$^2$ is inspired by the recent theoretical advancement of frame-based signal reconstruction approaches~\cite{ye2018deep, yin2017tale}. %\blfootnote{* indicates equal contribution} 
To make the paper self-contained, we provide a brief introduction to the frame theory (Section \ref{subsec:frameRecon}), tightness of Haar wavelets (Section \ref{subsec:haartight}) and our theoretical motivation (Section \ref{subsec:dcf}).
\subsection{Perfect reconstruction condition}
\label{subsec:frameRecon}
\paragraph{}
Consider an  \textit{analysis operator} $\Phi = \begin{bmatrix} \phi_1 &\cdots & \phi_{m} \end{bmatrix}\in\Rd^{n\times m}$, where $\{\phi_k\}_{k=1}^m$ is a family of functions in a Hilbert space $H$. 
Then,  $\{\phi_k\}_{k=1}^m$ is called a {\em frame} if it satisfies the following inequality \cite{duffin1952class}:
\begin{eqnarray}\label{eq:framebound}
\alpha \|f\|^2 \leq  \|\Phi^\top f\|^2  \leq \beta\|f\|^2,\quad \forall f \in H ,
\end{eqnarray}
where $f\in \Rd^n$ is an input signal and $\alpha,\beta>0$ are called the frame bounds. 
% The frame bounds can be represented by where $\sigma_{min}(A)$ and $\sigma_{max}(A)$ denote the minimum and maximum singular values of $A$, respectively. 
 
The original signal $f$ can be exactly recovered from the frame coefficient $z=\Phi f$ when there is the \textit{dual frame} $\tilde \Phi$ (\ie, \textit{synthesis operator}) satisfying \textit{the perfect reconstruction (PR) condition}:
$\tilde \Phi \Phi^\top = I$, since $f = \tilde \Phi z = \tilde \Phi \Phi^\top f = f$. 
Here, we call such frame \textit{tight} (\ie, $\alpha=\beta$ in \eqref{eq:framebound}) which is equivalent to $\tilde \Phi =\Phi$ or $\Phi\Phi^\top = I$.
Note that a tight frame does not amplify the power of the input and thus it has the minimum noise amplification factor.  
To achieve the best reconstruction performance, frame bases should satisfy another property, called energy compaction. This is particularly important to parametric models, which have to adaptively deal with varying amounts of information with a fixed number of parameters, \eg, deep neural networks (DNNs). For example, singular value decomposition (SVD) provides both tight and energy compact bases given an arbitrary signal. However, SVD is data-dependent, which makes it hard to use for a large dataset.
\subsection{Wavelet frames}
\label{subsec:haartight}
Wavelets are known to compactly represent signals while maintaining important information such as edges, thus resulting in a good energy compaction \cite{chan2005image}. Therefore, by using a tight wavelet filter-bank, we can improve the reconstruction performance of encoder-decoder type of networks with minimal noise amplification. %frame is a redundant, stable way of representing a signal
Specifically, the non-local basis $\Phi^T$ is now composed of a filter bank:
\begin{equation}
    \Phi = [T_1 \cdots T_L],
\end{equation}
where $T_k$ denotes the $k$-th subband operator and the filter bank is tight, \ie
\begin{equation}
\label{eq:wavtight}
    \Phi\Phi^T=\sum_{k=1}^L T_kT_k^T = I.
\end{equation}

In this paper, we use Haar wavelets which is one of the simplest tight filter bank frames with low
and high sub-band decomposition. Here, $T_1\in \Rd^{\frac{n}{2}\times n}$ is the low-pass subband. This is equivalent to the average pooling: 
\begin{eqnarray} 
\label{eq:avrpool}
T_1^\top =\frac{1}{\sqrt{2}}\left[
      \begin{array}{ccccccc}
        1  &   1 &   0  &   0  &   \cdots  &   0    & 0 \\
        0  &   0 &   1  &   1  &   \cdots  &   0    & 0\\
        & \vdots &      &      &           &\ddots  & \vdots \\
        0  &   0 &  0   &   0  &   \cdots  &     1  &   1    \\
      \end{array}
    \right]. 
\end{eqnarray}
Then, $T_2$ is the high pass filtering given by 
\begin{eqnarray} 
\label{eq:wavHfilter}
T_2^\top =\frac{1}{\sqrt{2}}\left[
      \begin{array}{ccccccc}
        1  &   -1 &   0  &   0  &   \cdots  &   0    & 0 \\
        0  &   0 &   1  &   -1  &   \cdots  &   0    & 0\\
        & \vdots &      &      &           &\ddots  & \vdots \\
        0  &   0 &  0   &   0  &   \cdots  &     1  &   -1    \\
      \end{array}
    \right]
\end{eqnarray}
and we can easily see that 
\begin{eqnarray} 
\label{eq:haartight}
T_1T_1^T + T_2T_2^T = I,
\end{eqnarray}
so the Haar wavelet frame is tight. 

\subsection{Theoretical motivation} 
\label{subsec:dcf}
In the perspective of the frame-based signal reconstruction, the commonly used encoder-decoder convolution structure of deep neural networks (DNNs), such as U-net \cite{ronneberger2015u}, can be interpreted as the data-driven way of learning the local bases $\Psi$ (\eg, convolution filters) with hand-crafted global bases $\Phi$ (\eg, max-pooling) \cite{ye2018deep}. Recently, Ye \etal \cite{ye2018deep} interpreted training DNNs as finding a multi-layer realization of the convolution framelets \cite{yin2017tale}:
\begin{eqnarray}
    Z &=& \Phi^T(f\circledast\Psi)  \label{eq:enc} \\
    f &=& \left(\tilde\Phi Z\right) \circledast \tilde \Psi, \label{eq:dec}
\end{eqnarray}
where $\Phi=[\phi_1,\cdots,\phi_n]$ and $\tilde \Phi =[\tilde \phi_1,\cdots, \tilde\phi_n] \in \Rd^{n\times n}$ (resp. $ \Psi=[\psi_1,\cdots, \psi_q]$ and $\tilde \Psi =[\tilde \psi_1,\cdots,\tilde\psi_q] \in \Rd^{d\times q}$) are { frames and their duals}. 
Here, $\circledast$ stands for the convolution operation. 

Therefore, the convolutional layers of the encoder learns the signal representation with a global pooling operation. We refer to $\Phi$ as global bases because it observes the entire image dimension $n$ while $\Psi$ learns local features from the data by $d\times d$ convolution kernels of $q$ channels. % of VGG layer \texttt{ReLU4_1}) 
% The multi-scale structure of the encoder-decoder coincides with the signal decomposition and reconstruction in multiple times using wavelets, \eg. multi-scale wavelet decomposition and reconstruction. 
When these frames satisfy the PR condition:
\begin{eqnarray}
 \tilde \Phi \Phi^\top = I_{n\times n} ,&  \Psi \tilde \Psi^{\top} = I_{d\times d}, \label{eq:id}
\end{eqnarray}
the input signal $f$ can be exactly recovered from the learned representations. 
Note that the encoder-decoder architectures of WCT \cite{wct} and PhotoWCT \cite{photowct} cannot satisfy the perfect reconstruction condition because of the max-pooling, which does not have its exact inverse (\ie, not a frame). On the other hand, our model WCT$^2$ can fully exploit the information from the encoder due to the favorable property of the wavelet decomposition and reconstruction, \ie, Haar wavelet pooling and unpooling.

\subsection{Proposed network architecture} 
\begin{figure*}[h!]
\begin{center}
\includegraphics[width=0.8\textwidth]{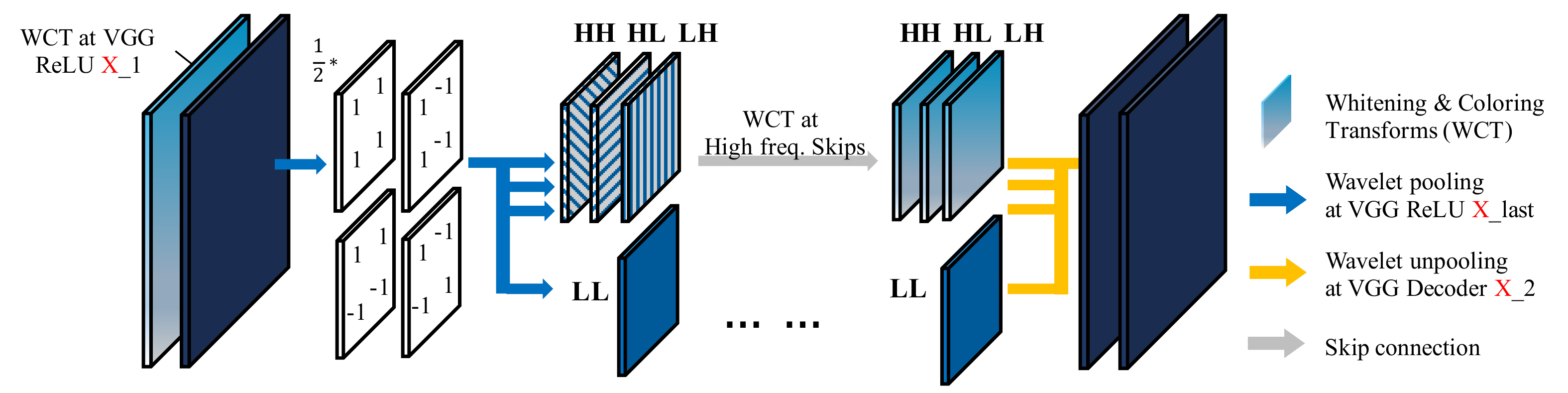}
\end{center}
  \caption{The proposed module using Haar wavelet pooling and unpooling. }
\label{fig:netarch1}
\end{figure*}

\begin{figure*}[h!]
    \centering
    \includegraphics[width=0.8\linewidth]{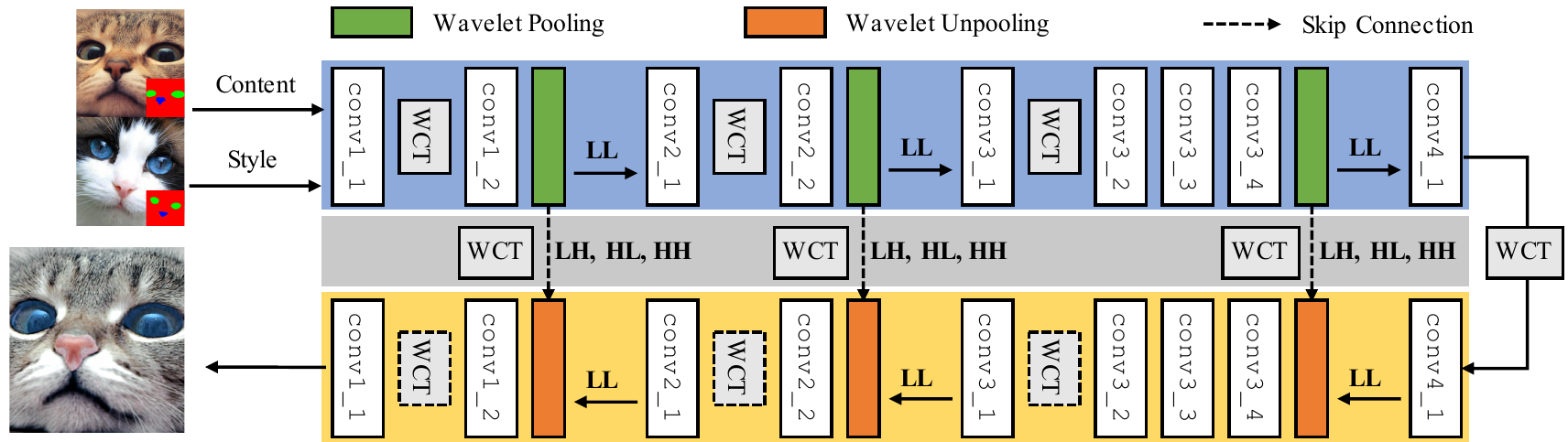}
\caption{Overview of the proposed progressive stylization. For the encoder, we perform WCT on the output of {\tt convX_1} layer and skip connections. For the decoder, we apply WCT on the output of {\tt convX_2} layer, which is optional.}
\label{fig:netarch2}
\end{figure*}
In Figure \ref{fig:netarch1}, a pair of encoder and decoder at same scale are shown. WCT is performed on the output of VGG {\tt convX\_1} layer followed by subsequent VGG layers and wavelet pooling. Only the low component passes to the next layer and the high frequency components are directly skipped to the corresponding decoding layer. At the decoder, the components are aggregated by the wavelet unpooling.

\subsection{Differences to PhotoWCT and Wavelet corrected transfer based on AdaIN (WCT-AdaIN)} 

\begin{figure*}[h]
\begin{center}
\includegraphics[width=1\textwidth]{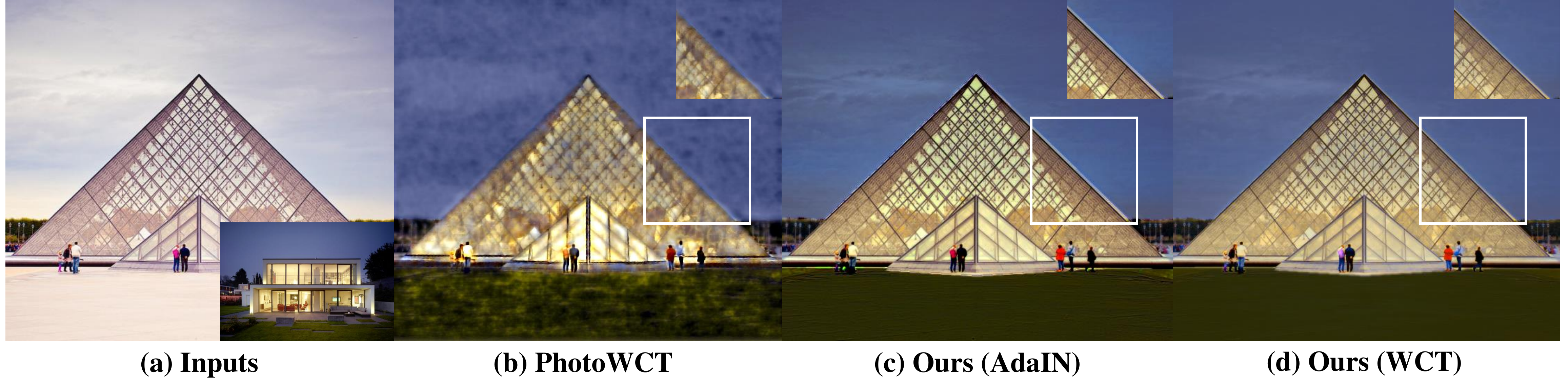}
\end{center}
  \caption{Photorealistic style transfer results of (a) input pairs using (b) PhotoWCT, (c) Ours (AdaIN) and (d) Ours (WCT). (c) is the results using our model architecture combined with AdaIN as the stylization method, and (d) is WCT$^2$ (proposed).}
\label{fig:wct_adain}
\end{figure*}

Our method shares the motivation with PhotoWCT but the way we posit the problem and reach to its solution is fundamentally different from PhotoWCT: \textit{i}) We showed that the reason why PhotoWCT fails in preserving spatial information is because \textbf{pooling and unpooling operations cannot satisfy the frame condition} \textit{ii}) Based on this theoretical analysis, our model architecture is \textbf{specifically designed to perfectly preserve spatial structure}, which is \textbf{proved effective in theory and practice.} This removes the necessity of post-processing, thus making our model far more practical and powerful than the previous methods. 
\textit{iii}) The wavelet corrected model is \textbf{by no means limited to a specific stylization method}. It can serve as a \textbf{general architecture for photorealistic style transfer}, which is compatible with various methods, \eg, AdaIN (Figure \ref{fig:wct_adain} (c)). Currently, our method (WCT$^2$) can process 1k resolution image in 4.7 seconds and this can be accelerated further ($\sim$1 second) by employing adaptive instance normalization (AdaIN) instead of time-consuming SVD procedure. 

\subsection{Qualitative comparison with artistic style transfer results}
\label{subsec:artistic_qual}

We compare our proposed WCT$^2$ with popular artistic style transfer methods including NeuralStyle \cite{gatys2016image}, AdaIN \cite{huang2017arbitrary} and WCT \cite{wct} in Figure~\ref{fig:artistic_comparison}.
To apply semantic segmentation map to the artistic style transfer methods, we followed the spatial control techniques proposed by the authors \cite{luan2017deep, huang2017arbitrary, wct} respectively.
In the figure, artistic style transfer methods generate undesired distortions and artifacts and often fail to maintain the structural information despite the spatial control with segmentation maps.
In comparison, because of the proposed wavelet corrected transfer, our proposed WCT$^2$ prevents unrealistic artifacts and preserve the structure information such as edges.

\subsection{Additional Qualitative comparison with photorealistic style transfer} 
\label{subsec:qual}
Additional qualitative results using WCT$^2$ and its variants are shown in Figure~\ref{fig:photorealistic_comparison_suppl1}, Figure~\ref{fig:photorealistic_comparison_suppl2} and Figure~\ref{fig:photorealistic_comparison_suppl3}. The video stylization results can be found in one of the other supplementary materials. 

\input{comparison_fig_artistic.tex}
\input{comparison_fig_supl.tex}

\newpage
\clearpage
\end{document}

%% file: comparison_fig.tex
\begin{figure*}[ht!]
    \centering
    \begin{subfigure}[h]{0.109\linewidth}
        \centering
        \includegraphics[width=\linewidth]{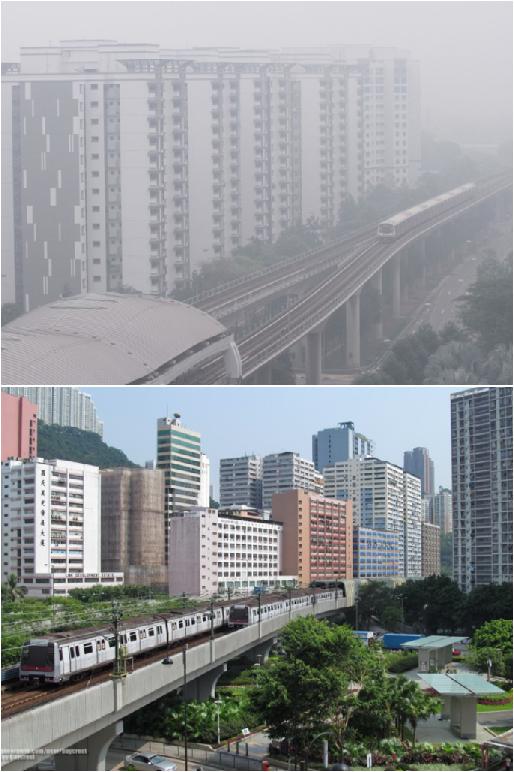}
    \end{subfigure}
    \begin{subfigure}[h]{0.218\linewidth}
        \centering
        \includegraphics[width=\linewidth]{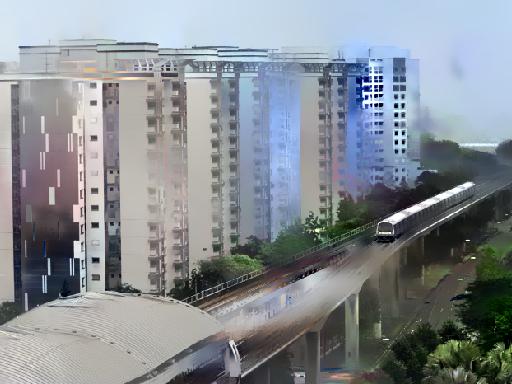}
    \end{subfigure}
    \begin{subfigure}[h]{0.218\linewidth}
        \centering
        \includegraphics[width=\linewidth]{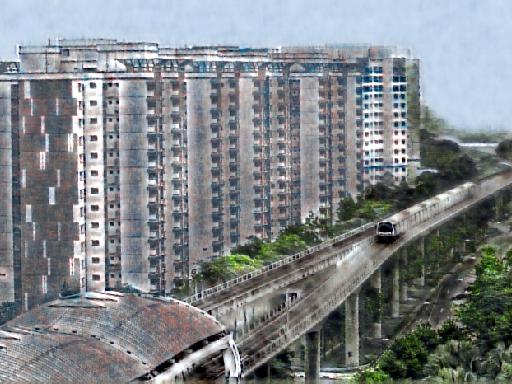}
    \end{subfigure}
    \begin{subfigure}[h]{0.218\linewidth}
        \centering
        \includegraphics[width=\linewidth]{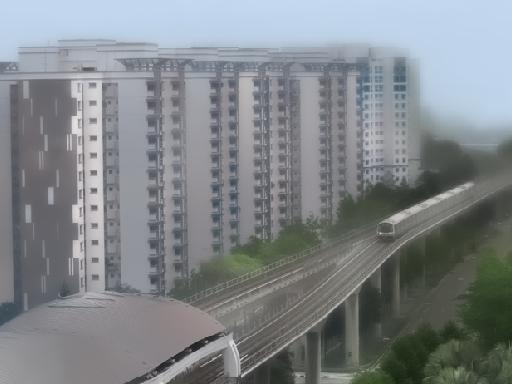}
    \end{subfigure}
    \begin{subfigure}[h]{0.218\linewidth}
        \centering
        \includegraphics[width=\linewidth]{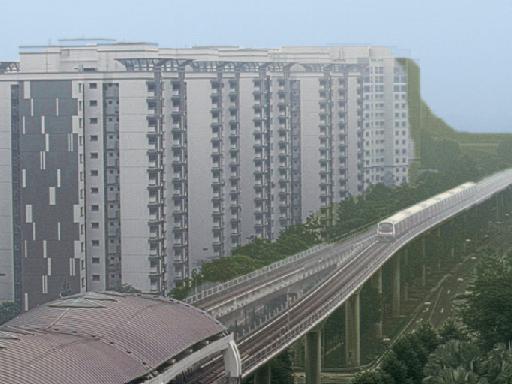}
    \end{subfigure}\\
    \begin{subfigure}[h]{0.109\linewidth}
        \centering
        \includegraphics[width=\linewidth]{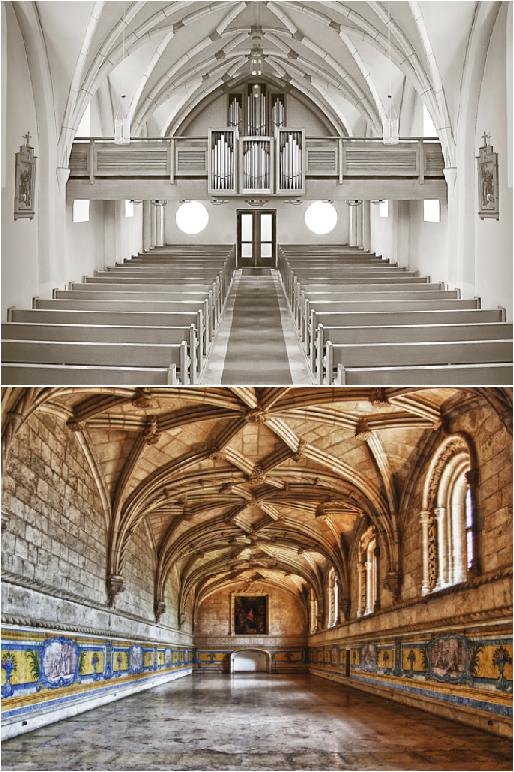}
    \end{subfigure}
    \begin{subfigure}[h]{0.218\linewidth}
        \centering
        \includegraphics[width=\linewidth]{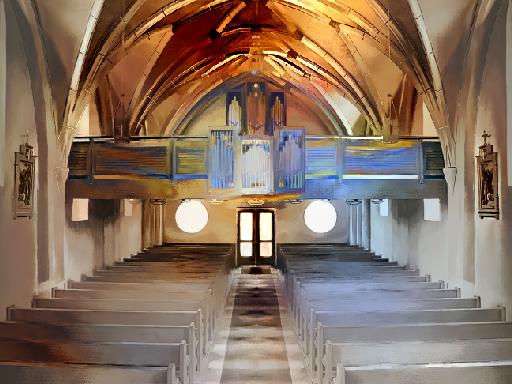}
    \end{subfigure}
    \begin{subfigure}[h]{0.218\linewidth}
        \centering
        \includegraphics[width=\linewidth]{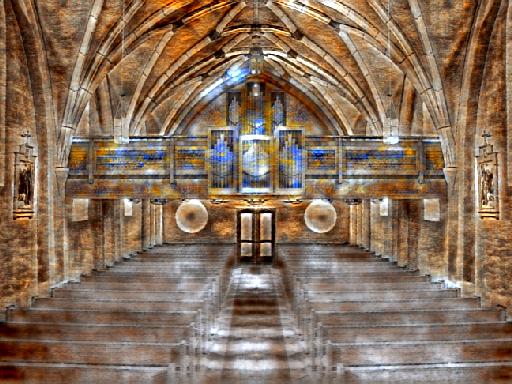}
    \end{subfigure}
    \begin{subfigure}[h]{0.218\linewidth}
        \centering
        \includegraphics[width=\linewidth]{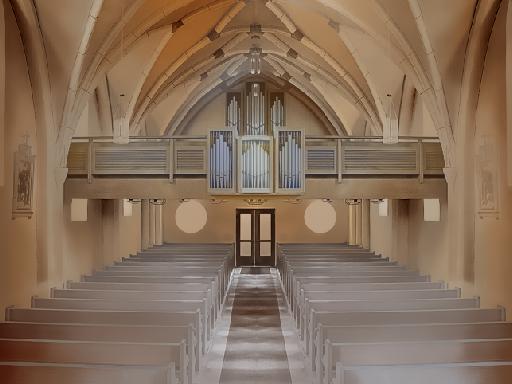}
    \end{subfigure}
    \begin{subfigure}[h]{0.218\linewidth}
        \centering
        \includegraphics[width=\linewidth]{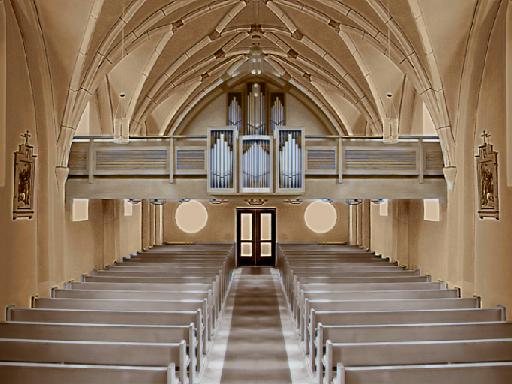}
    \end{subfigure}\\
    \begin{subfigure}[h]{0.109\linewidth}
        \centering
        \includegraphics[width=\linewidth]{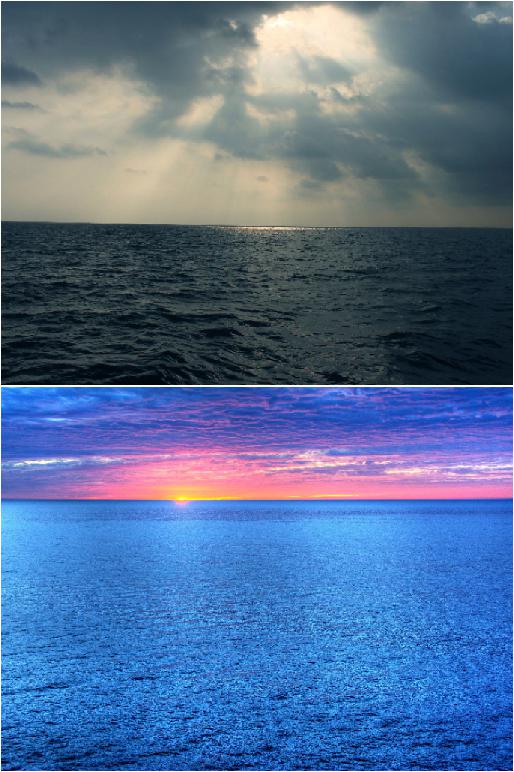}
    \end{subfigure}
    \begin{subfigure}[h]{0.218\linewidth}
        \centering
        \includegraphics[width=\linewidth]{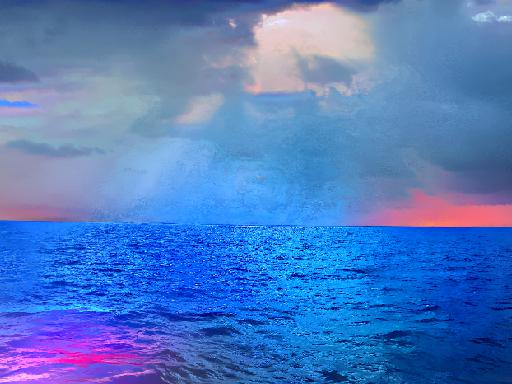}
    \end{subfigure}
    \begin{subfigure}[h]{0.218\linewidth}
        \centering
        \includegraphics[width=\linewidth]{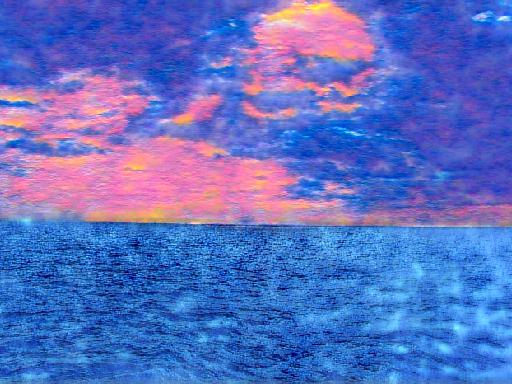}
    \end{subfigure}
    \begin{subfigure}[h]{0.218\linewidth}
        \centering
        \includegraphics[width=\linewidth]{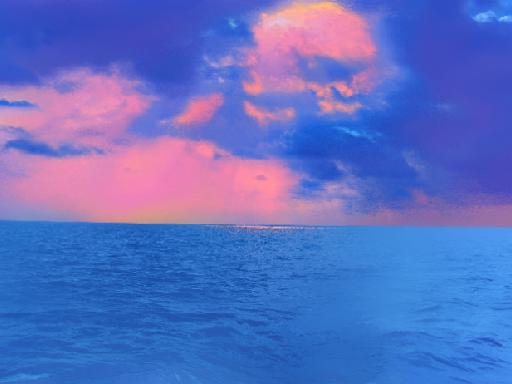}
    \end{subfigure}
    \begin{subfigure}[h]{0.218\linewidth}
        \centering
        \includegraphics[width=\linewidth]{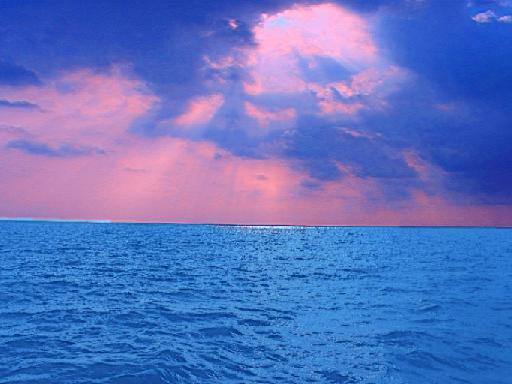}
    \end{subfigure}\\
    \begin{subfigure}[h]{0.109\linewidth}
        \centering
        \includegraphics[width=\linewidth]{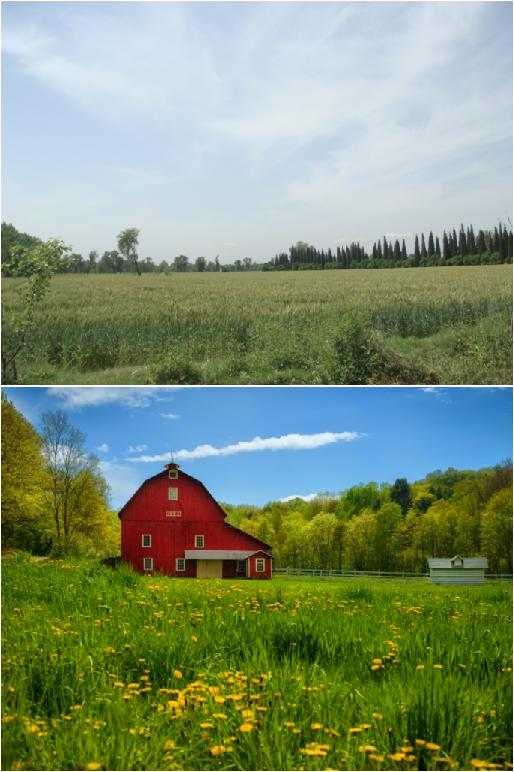}
    \end{subfigure}
    \begin{subfigure}[h]{0.218\linewidth}
        \centering
        \includegraphics[width=\linewidth]{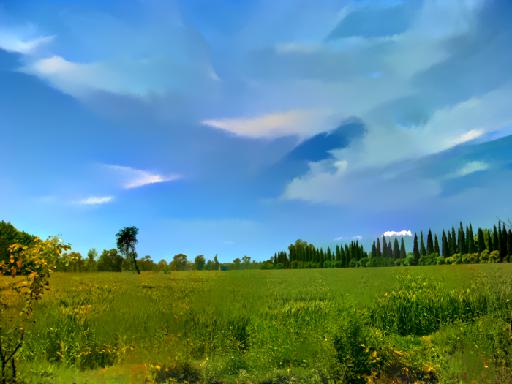}
    \end{subfigure}
    \begin{subfigure}[h]{0.218\linewidth}
        \centering
        \includegraphics[width=\linewidth]{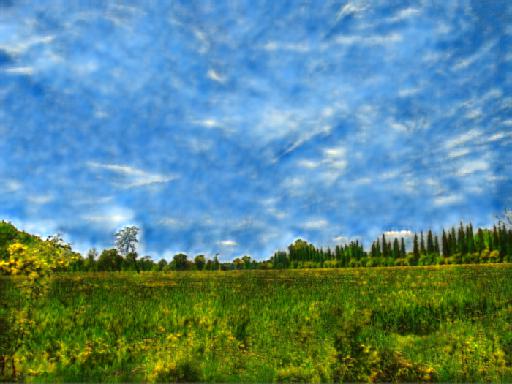}
    \end{subfigure}
    \begin{subfigure}[h]{0.218\linewidth}
        \centering
        \includegraphics[width=\linewidth]{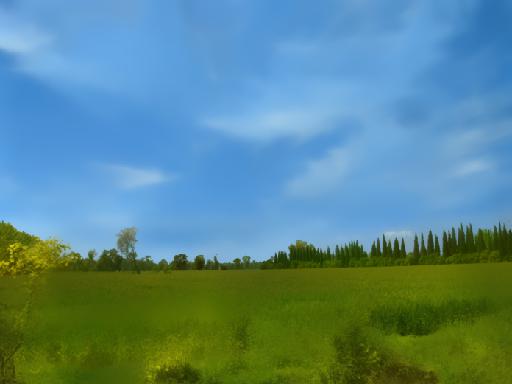}
    \end{subfigure}
    \begin{subfigure}[h]{0.218\linewidth}
        \centering
        \includegraphics[width=\linewidth]{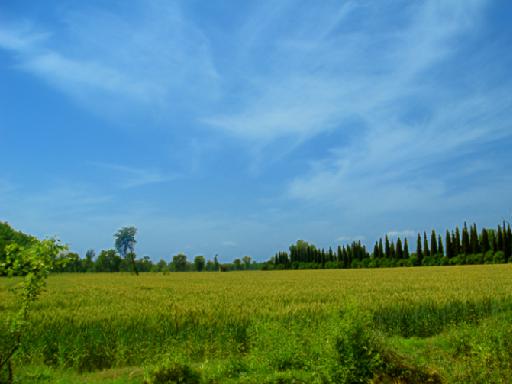}
    \end{subfigure}\\
    \begin{subfigure}[h]{0.109\linewidth}
        \centering
        \includegraphics[width=\linewidth]{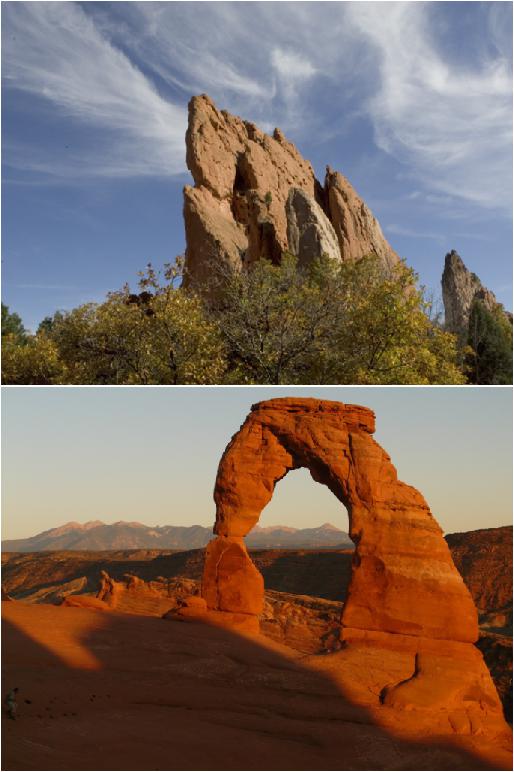}
    \end{subfigure}
    \begin{subfigure}[h]{0.218\linewidth}
        \centering
        \includegraphics[width=\linewidth]{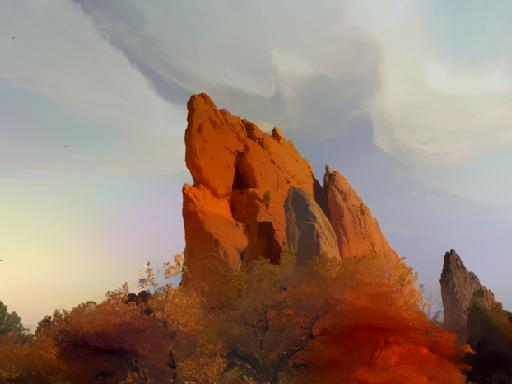}
    \end{subfigure}
    \begin{subfigure}[h]{0.218\linewidth}
        \centering
        \includegraphics[width=\linewidth]{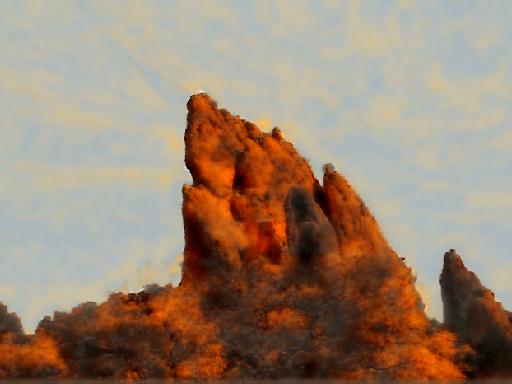}
    \end{subfigure}
    \begin{subfigure}[h]{0.218\linewidth}
        \centering
        \includegraphics[width=\linewidth]{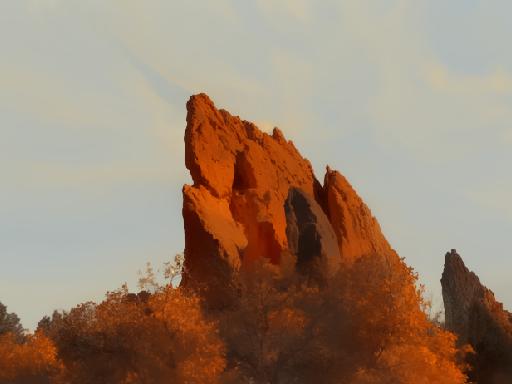}
    \end{subfigure}
    \begin{subfigure}[h]{0.218\linewidth}
        \centering
        \includegraphics[width=\linewidth]{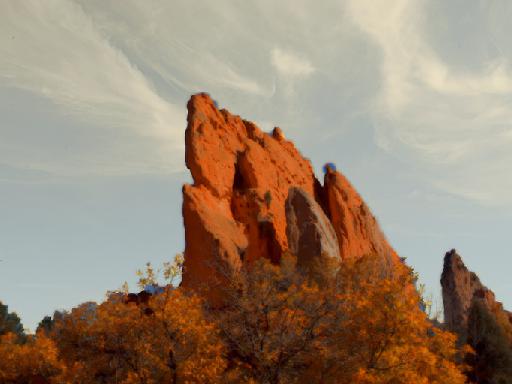}
    \end{subfigure}\\
    \begin{subfigure}[h]{0.109\linewidth}
        \centering
        \includegraphics[width=\linewidth]{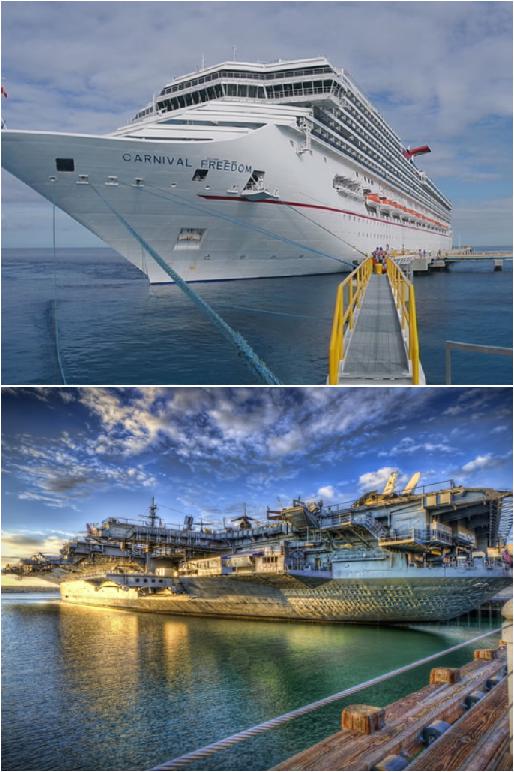}
    \end{subfigure}
    \begin{subfigure}[h]{0.218\linewidth}
        \centering
        \includegraphics[width=\linewidth]{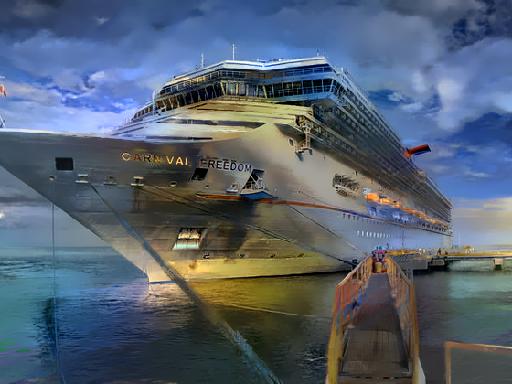}
    \end{subfigure}
    \begin{subfigure}[h]{0.218\linewidth}
        \centering
        \includegraphics[width=\linewidth]{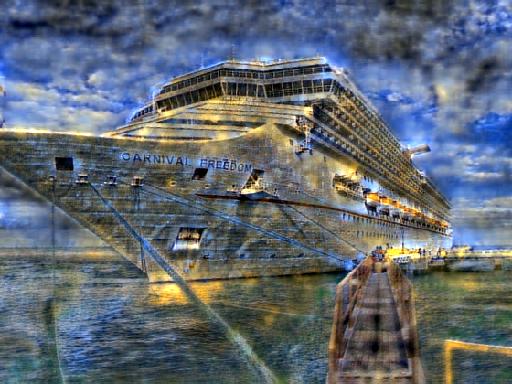}
    \end{subfigure}
    \begin{subfigure}[h]{0.218\linewidth}
        \centering
        \includegraphics[width=\linewidth]{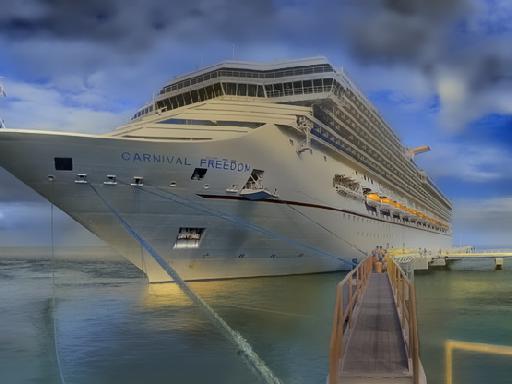}
    \end{subfigure}
    \begin{subfigure}[h]{0.218\linewidth}
        \centering
        \includegraphics[width=\linewidth]{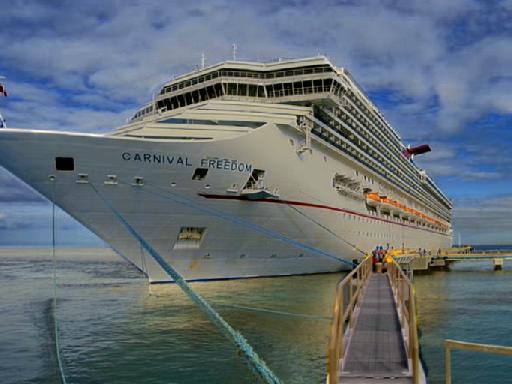}
    \end{subfigure}\\
\begin{subfigure}[h]{0.109\linewidth}
        \centering
        \includegraphics[width=\linewidth]{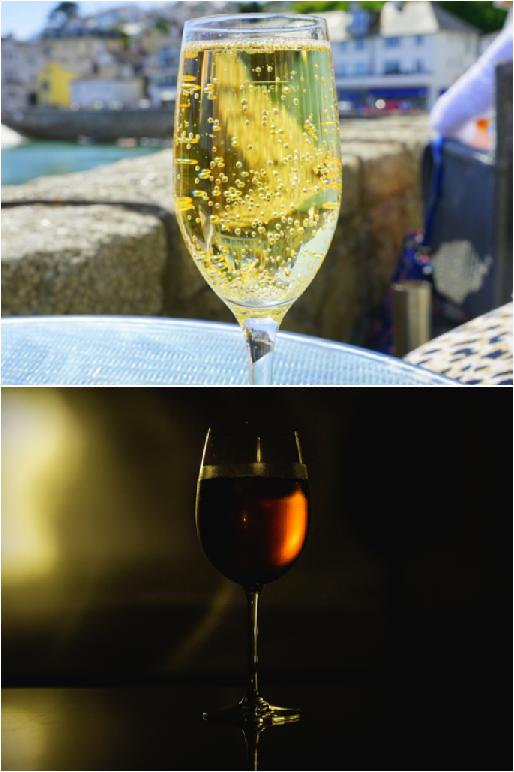}
        \caption{Input}
    \end{subfigure}
    \begin{subfigure}[h]{0.218\linewidth}
        \centering
        \includegraphics[width=\linewidth]{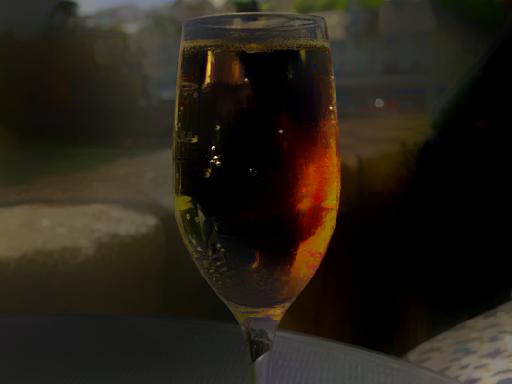}
        \caption{DPST \cite{luan2017deep}}
        \label{fig:photorealistic_comparison:dpst}
    \end{subfigure}
    \begin{subfigure}[h]{0.218\linewidth}
        \centering
        \includegraphics[width=\linewidth]{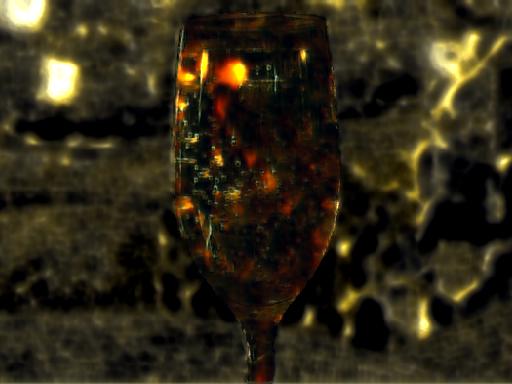}
        \caption{PhotoWCT \cite{photowct}}
        \label{fig:photorealistic_comparison:photowct}
    \end{subfigure}
    \begin{subfigure}[h]{0.218\linewidth}
        \centering
        \includegraphics[width=\linewidth]{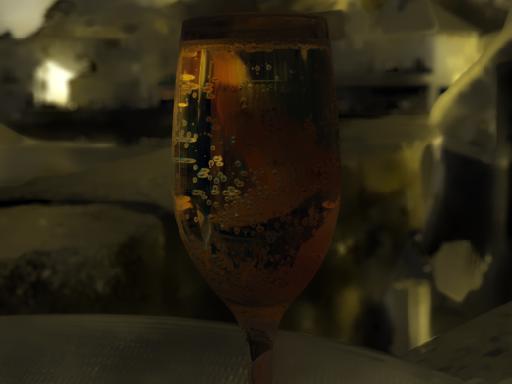}
        \caption{PhotoWCT (full) \cite{photowct}}
        \label{fig:photorealistic_comparison:photowct_full}
    \end{subfigure}
    \begin{subfigure}[h]{0.218\linewidth}
        \centering
        \includegraphics[width=\linewidth]{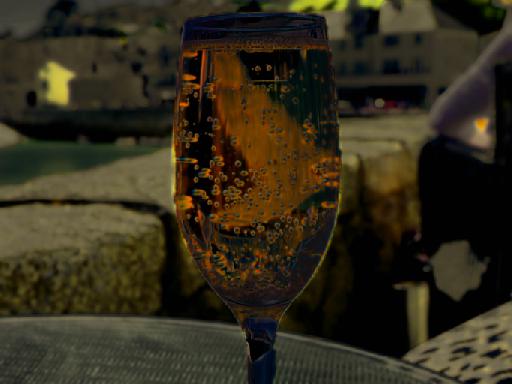}
        \caption{Ours (WCT$^2$)}
        \label{fig:photorealistic_comparison:ours}
    \end{subfigure}
\caption{Photorealistic stylization results. Given (a) an input pair (top: content, bottom: style), the results of (b) deep photo style transfer (DPST) \cite{luan2017deep}, (c) and (d) PhotoWCT \cite{photowct}, and (e) ours (WCT$^2$) are shown. PhotoWCT (full) denotes the results after applying two post-processing steps proposed by the authors \cite{photowct}. Note that WCT$^2$ \textbf{does not} need any post-processing.}
\label{fig:photorealistic_comparison}
\end{figure*}

%% file: comparison_fig_artistic.tex
\begin{figure}[ht!]
    \centering
    \begin{subfigure}[h]{0.109\linewidth}
        \centering
        \includegraphics[width=\linewidth]{./images/comparison/00_input2.jpg}
    \end{subfigure}
    \begin{subfigure}[h]{0.218\linewidth}
        \centering
        \includegraphics[width=\linewidth]{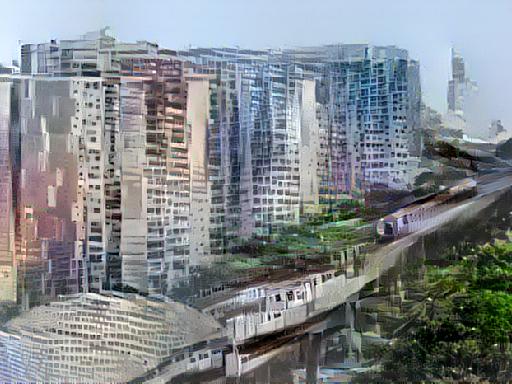}
    \end{subfigure}
    \begin{subfigure}[h]{0.218\linewidth}
        \centering
        \includegraphics[width=\linewidth]{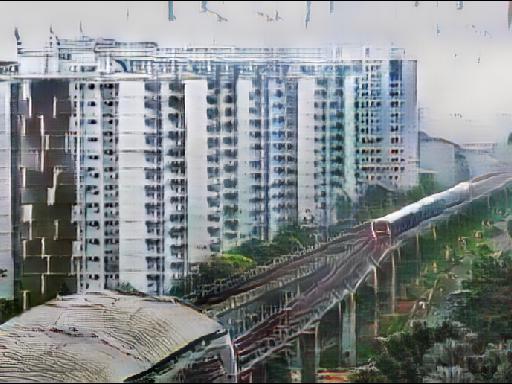}
    \end{subfigure}
    \begin{subfigure}[h]{0.218\linewidth}
        \centering
        \includegraphics[width=\linewidth]{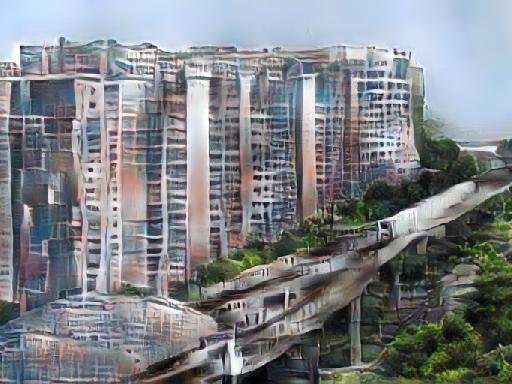}
    \end{subfigure}
    \begin{subfigure}[h]{0.218\linewidth}
        \centering
        \includegraphics[width=\linewidth]{./images/comparison/00_ours.jpg}
    \end{subfigure}\\
    \begin{subfigure}[h]{0.109\linewidth}
        \centering
        \includegraphics[width=\linewidth]{./images/comparison/01_input2.jpg}
    \end{subfigure}
    \begin{subfigure}[h]{0.218\linewidth}
        \centering
        \includegraphics[width=\linewidth]{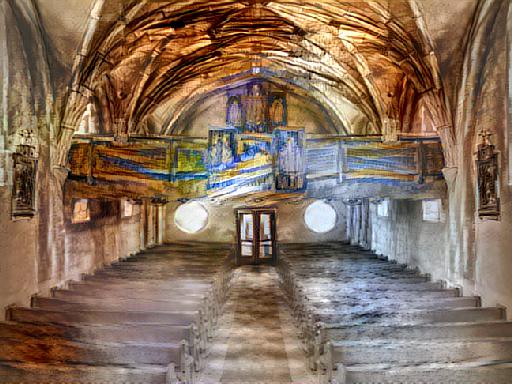}
    \end{subfigure}
    \begin{subfigure}[h]{0.218\linewidth}
        \centering
        \includegraphics[width=\linewidth]{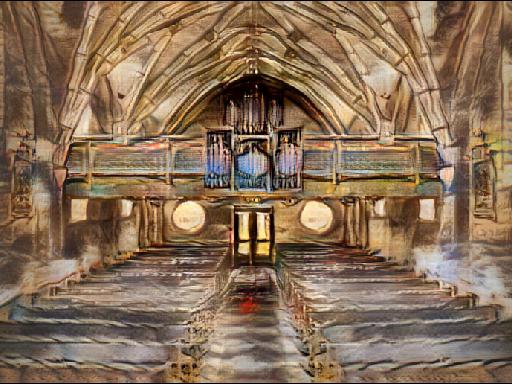}
    \end{subfigure}
    \begin{subfigure}[h]{0.218\linewidth}
        \centering
        \includegraphics[width=\linewidth]{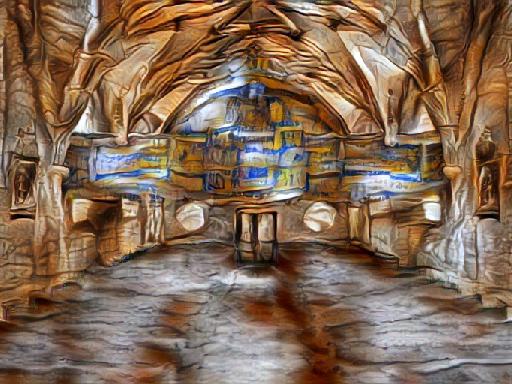}
    \end{subfigure}
    \begin{subfigure}[h]{0.218\linewidth}
        \centering
        \includegraphics[width=\linewidth]{./images/comparison/01_ours.jpg}
    \end{subfigure}\\
    \begin{subfigure}[h]{0.109\linewidth}
        \centering
        \includegraphics[width=\linewidth]{./images/comparison/02_input2.jpg}
    \end{subfigure}
    \begin{subfigure}[h]{0.218\linewidth}
        \centering
        \includegraphics[width=\linewidth]{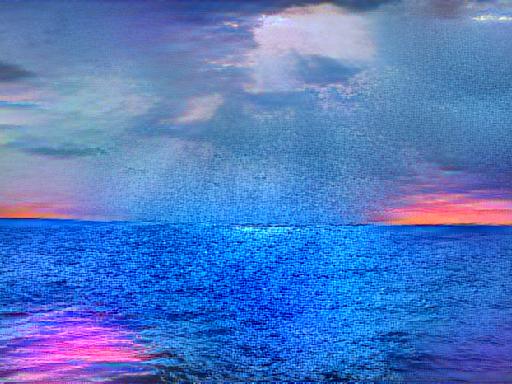}
    \end{subfigure}
    \begin{subfigure}[h]{0.218\linewidth}
        \centering
        \includegraphics[width=\linewidth]{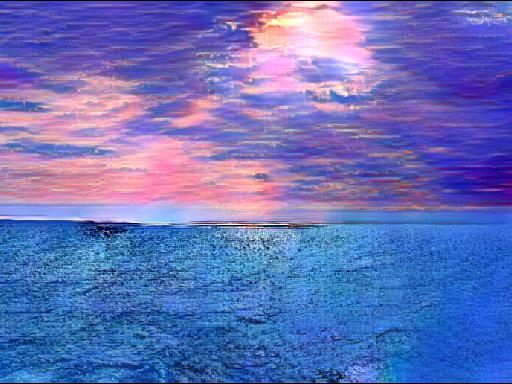}
    \end{subfigure}
    \begin{subfigure}[h]{0.218\linewidth}
        \centering
        \includegraphics[width=\linewidth]{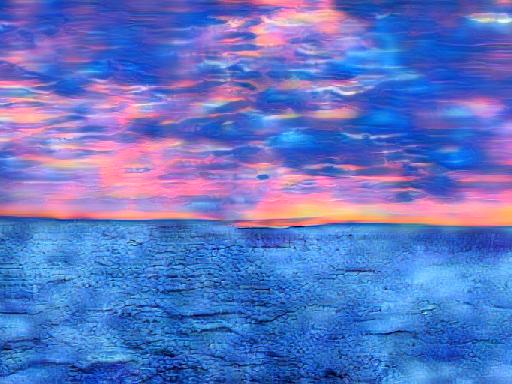}
    \end{subfigure}
    \begin{subfigure}[h]{0.218\linewidth}
        \centering
        \includegraphics[width=\linewidth]{./images/comparison/02_ours.jpg}
    \end{subfigure}\\
    \begin{subfigure}[h]{0.109\linewidth}
        \centering
        \includegraphics[width=\linewidth]{./images/comparison/03_input2.jpg}
    \end{subfigure}
    \begin{subfigure}[h]{0.218\linewidth}
        \centering
        \includegraphics[width=\linewidth]{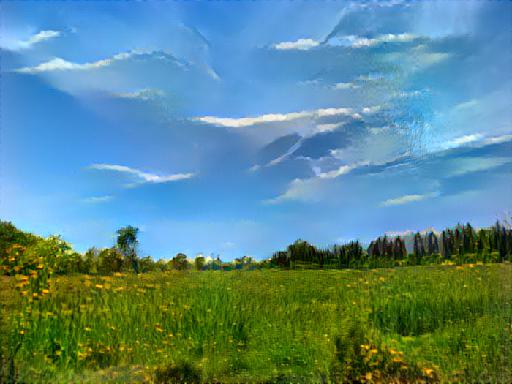}
    \end{subfigure}
    \begin{subfigure}[h]{0.218\linewidth}
        \centering
        \includegraphics[width=\linewidth]{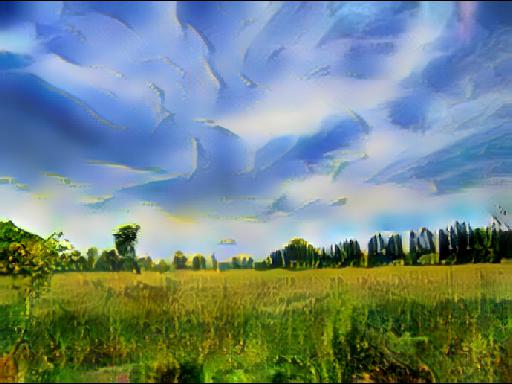}
    \end{subfigure}
    \begin{subfigure}[h]{0.218\linewidth}
        \centering
        \includegraphics[width=\linewidth]{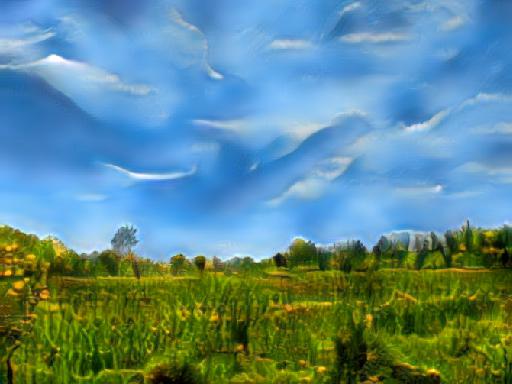}
    \end{subfigure}
    \begin{subfigure}[h]{0.218\linewidth}
        \centering
        \includegraphics[width=\linewidth]{./images/comparison/03_ours.jpg}
    \end{subfigure}\\
    \begin{subfigure}[h]{0.109\linewidth}
        \centering
        \includegraphics[width=\linewidth]{./images/comparison/04_input2.jpg}
    \end{subfigure}
    \begin{subfigure}[h]{0.218\linewidth}
        \centering
        \includegraphics[width=\linewidth]{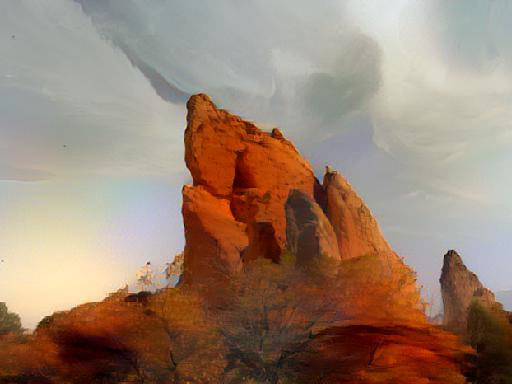}
    \end{subfigure}
    \begin{subfigure}[h]{0.218\linewidth}
        \centering
        \includegraphics[width=\linewidth]{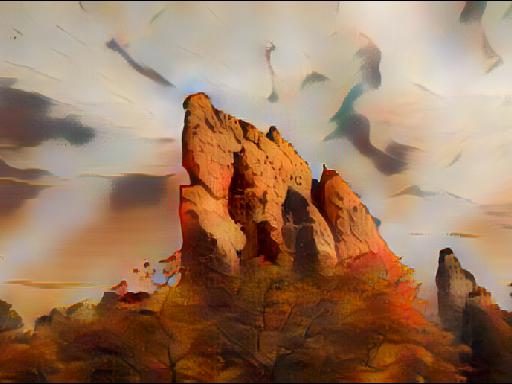}
    \end{subfigure}
    \begin{subfigure}[h]{0.218\linewidth}
        \centering
        \includegraphics[width=\linewidth]{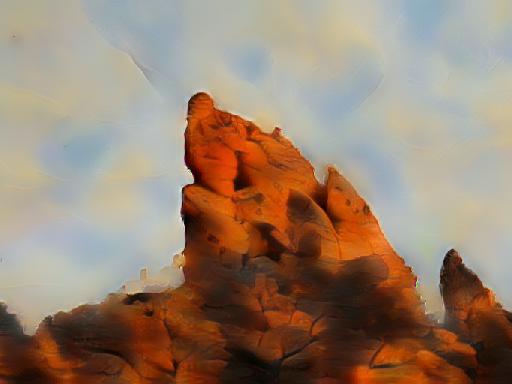}
    \end{subfigure}
    \begin{subfigure}[h]{0.218\linewidth}
        \centering
        \includegraphics[width=\linewidth]{./images/comparison/04_ours.jpg}
    \end{subfigure}\\
    \begin{subfigure}[h]{0.109\linewidth}
        \centering
        \includegraphics[width=\linewidth]{./images/comparison/05_input2.jpg}
    \end{subfigure}
    \begin{subfigure}[h]{0.218\linewidth}
        \centering
        \includegraphics[width=\linewidth]{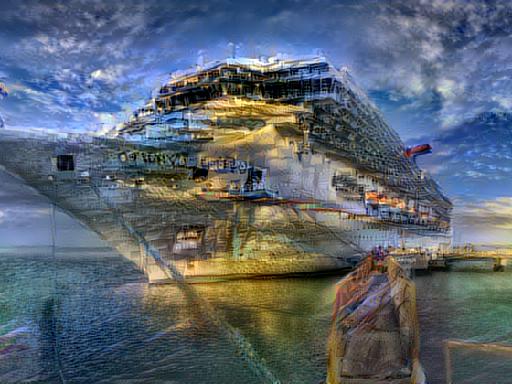}
    \end{subfigure}
    \begin{subfigure}[h]{0.218\linewidth}
        \centering
        \includegraphics[width=\linewidth]{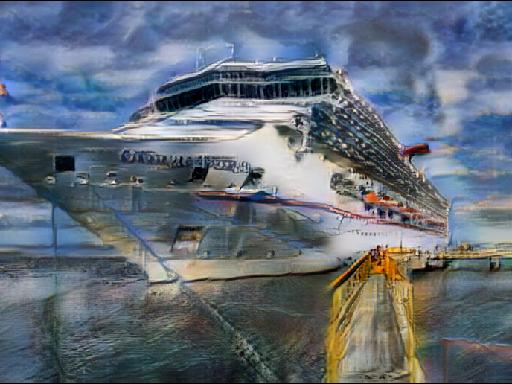}
    \end{subfigure}
    \begin{subfigure}[h]{0.218\linewidth}
        \centering
        \includegraphics[width=\linewidth]{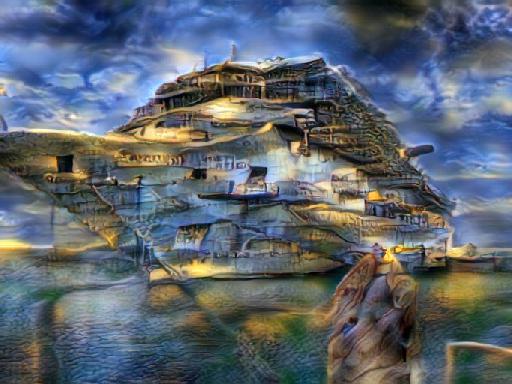}
    \end{subfigure}
    \begin{subfigure}[h]{0.218\linewidth}
        \centering
        \includegraphics[width=\linewidth]{./images/comparison/05_ours.jpg}
    \end{subfigure}\\
    \begin{subfigure}[h]{0.109\linewidth}
        \centering
        \includegraphics[width=\linewidth]{./images/comparison/06_input2.jpg}
        \caption{Input}
    \end{subfigure}
    \begin{subfigure}[h]{0.218\linewidth}
        \centering
        \includegraphics[width=\linewidth]{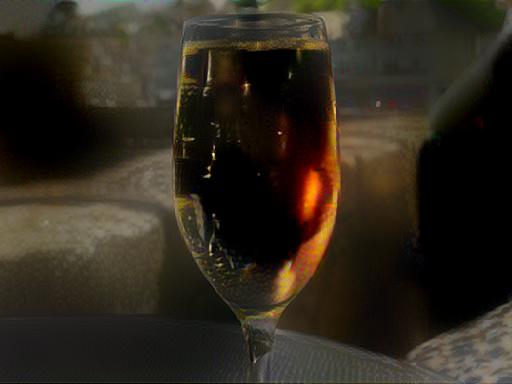}
        \caption{NeuralStyle \cite{gatys2016image}}
    \end{subfigure}
    \begin{subfigure}[h]{0.218\linewidth}
        \centering
        \includegraphics[width=\linewidth]{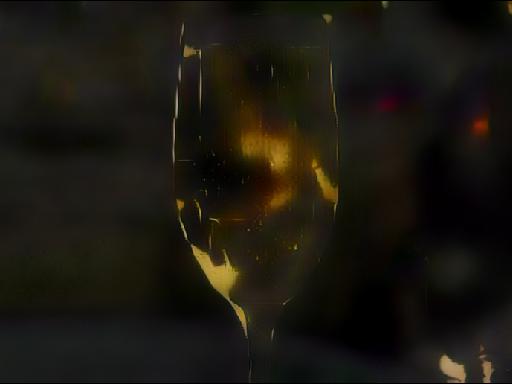}
        \caption{AdaIN \cite{huang2017arbitrary}}
    \end{subfigure}
    \begin{subfigure}[h]{0.218\linewidth}
        \centering
        \includegraphics[width=\linewidth]{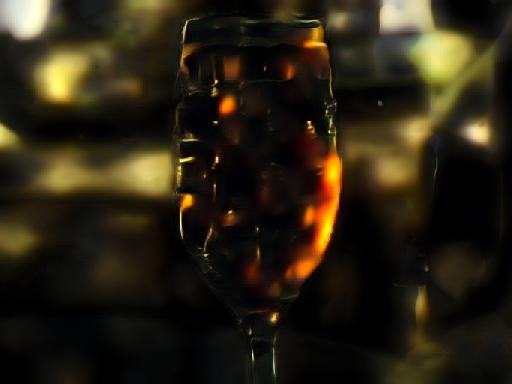}
        \caption{WCT \cite{wct}}
    \end{subfigure}
    \begin{subfigure}[h]{0.218\linewidth}
        \centering
        \includegraphics[width=\linewidth]{./images/comparison/06_ours.jpg}
        \caption{Ours (WCT$^2$)}
    \end{subfigure}
\caption{Qualitative comparison with artistic style transfer results. Given (a) an input pair (top: content, bottom: style), we compare the results of (b) NeuralStyle \cite{gatys2016image}, (c) AdaIN \cite{huang2017arbitrary} (d) WCT \cite{wct} and (e) ours (WCT$^2$).}
\label{fig:artistic_comparison}
\end{figure}

%% file: comparison_fig_supl.tex
\begin{figure}[ht!]
    \centering
    \begin{subfigure}[ht!]{0.109\linewidth}
        \centering
        \includegraphics[width=\linewidth]{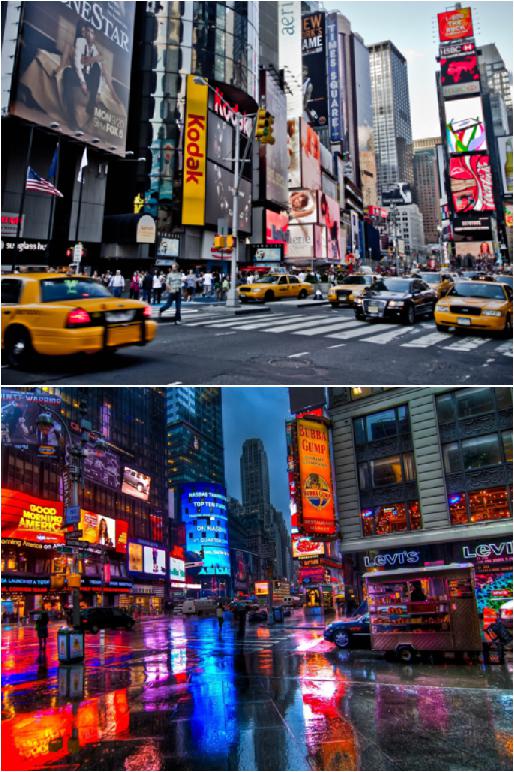}
    \end{subfigure}
    \begin{subfigure}[ht!]{0.218\linewidth}
        \centering
        \includegraphics[width=\linewidth]{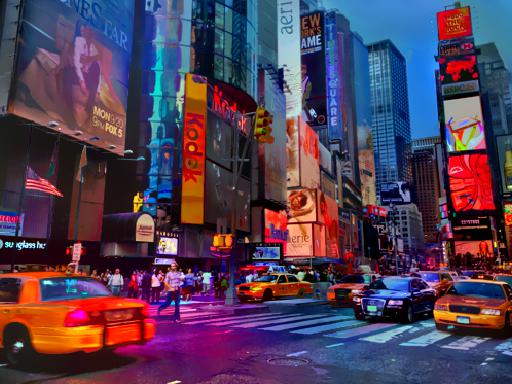}
    \end{subfigure}
    \begin{subfigure}[ht!]{0.218\linewidth}
        \centering
        \includegraphics[width=\linewidth]{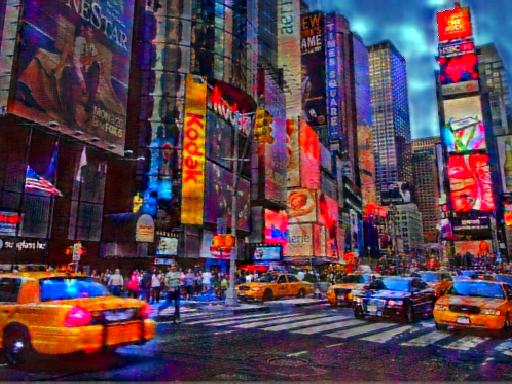}
    \end{subfigure}
    \begin{subfigure}[ht!]{0.218\linewidth}
        \centering
        \includegraphics[width=\linewidth]{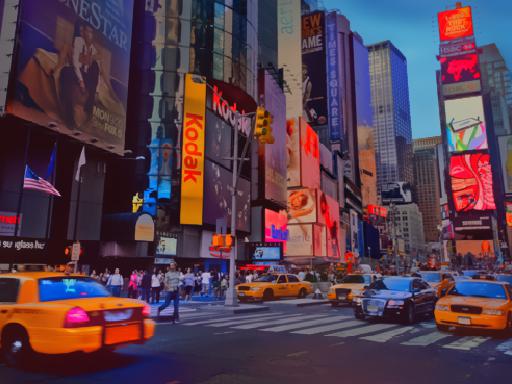}
    \end{subfigure}
    \begin{subfigure}[ht!]{0.218\linewidth}
        \centering
        \includegraphics[width=\linewidth]{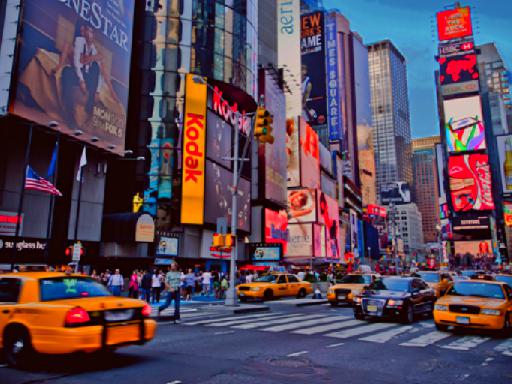}
    \end{subfigure}\\
    \begin{subfigure}[ht!]{0.109\linewidth}
        \centering
        \includegraphics[width=\linewidth]{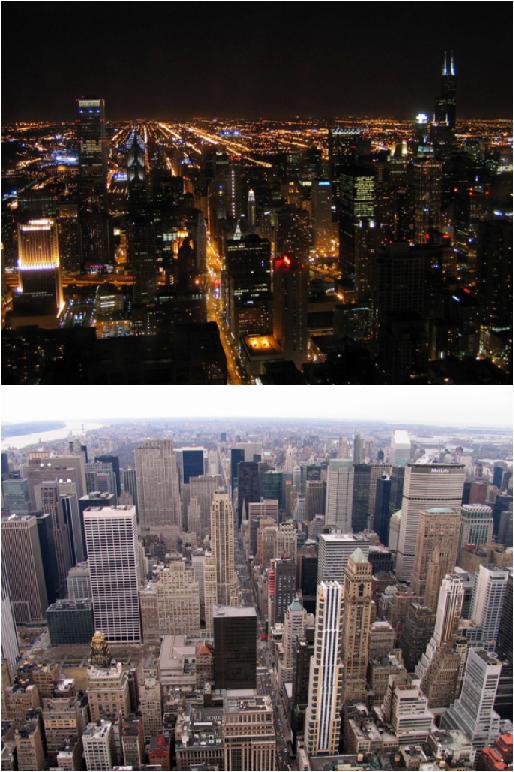}
    \end{subfigure}
    \begin{subfigure}[ht!]{0.218\linewidth}
        \centering
        \includegraphics[width=\linewidth]{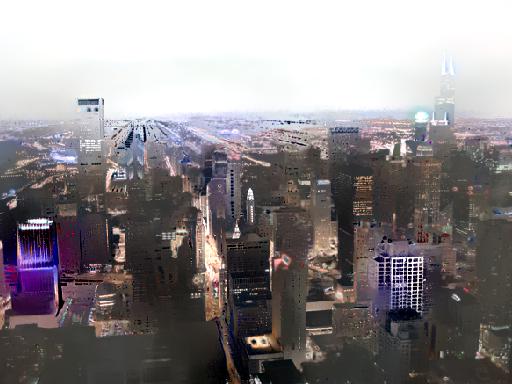}
    \end{subfigure}
    \begin{subfigure}[ht!]{0.218\linewidth}
        \centering
        \includegraphics[width=\linewidth]{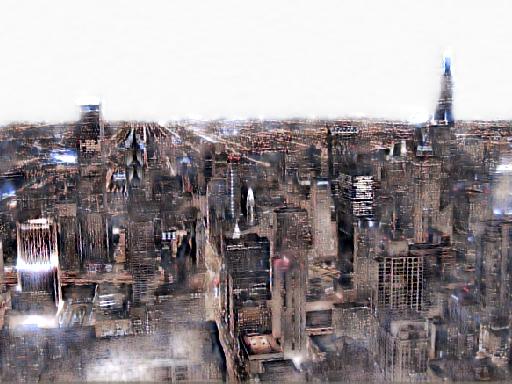}
    \end{subfigure}
    \begin{subfigure}[ht!]{0.218\linewidth}
        \centering
        \includegraphics[width=\linewidth]{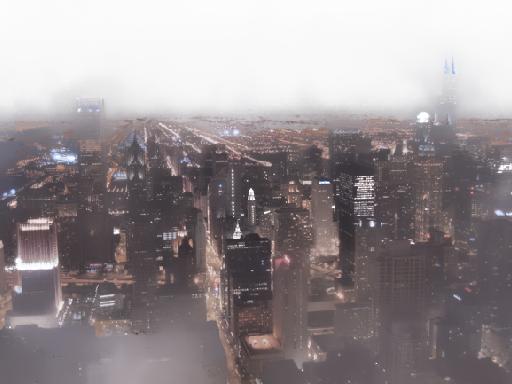}
    \end{subfigure}
    \begin{subfigure}[ht!]{0.218\linewidth}
        \centering
        \includegraphics[width=\linewidth]{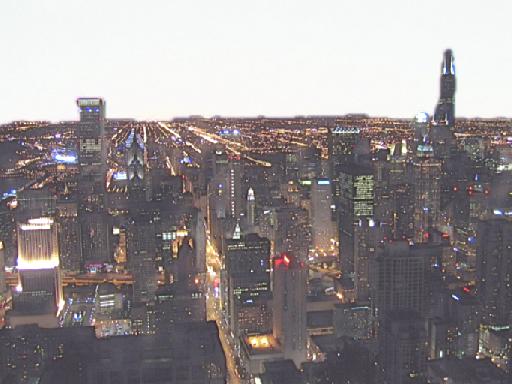}
    \end{subfigure}\\
    \begin{subfigure}[ht!]{0.109\linewidth}
        \centering
        \includegraphics[width=\linewidth]{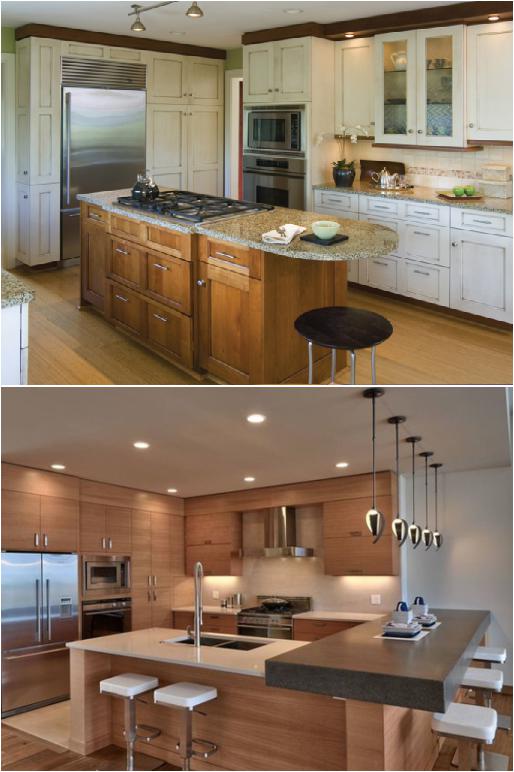}
    \end{subfigure}
    \begin{subfigure}[ht!]{0.218\linewidth}
        \centering
        \includegraphics[width=\linewidth]{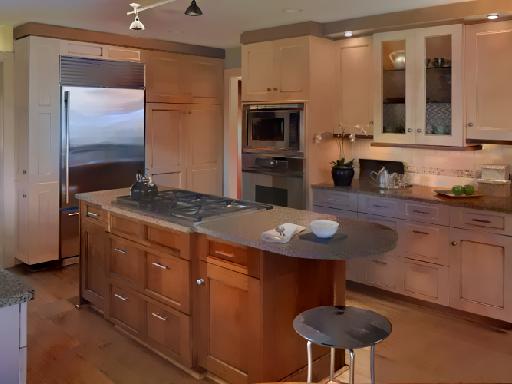}
    \end{subfigure}
    \begin{subfigure}[ht!]{0.218\linewidth}
        \centering
        \includegraphics[width=\linewidth]{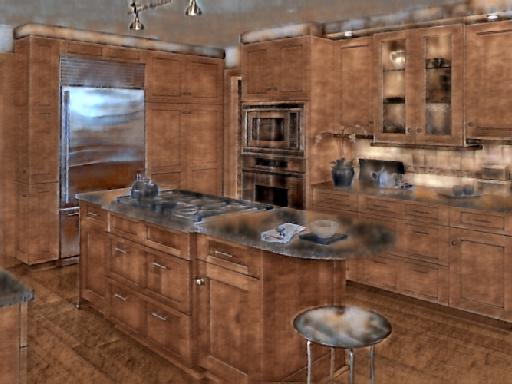}
    \end{subfigure}
    \begin{subfigure}[ht!]{0.218\linewidth}
        \centering
        \includegraphics[width=\linewidth]{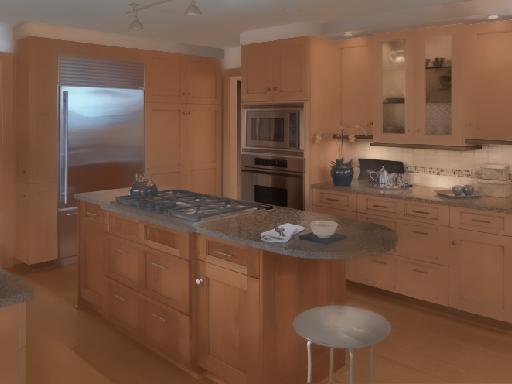}
    \end{subfigure}
    \begin{subfigure}[ht!]{0.218\linewidth}
        \centering
        \includegraphics[width=\linewidth]{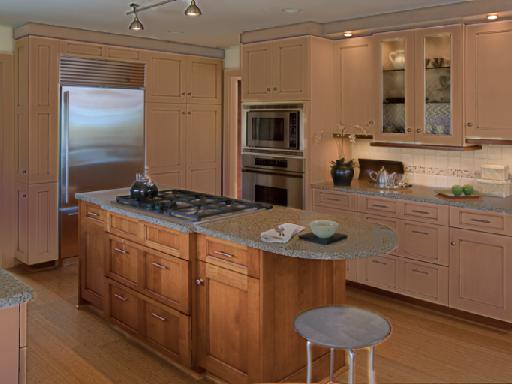}
    \end{subfigure}\\
    \begin{subfigure}[ht!]{0.109\linewidth}
        \centering
        \includegraphics[width=\linewidth]{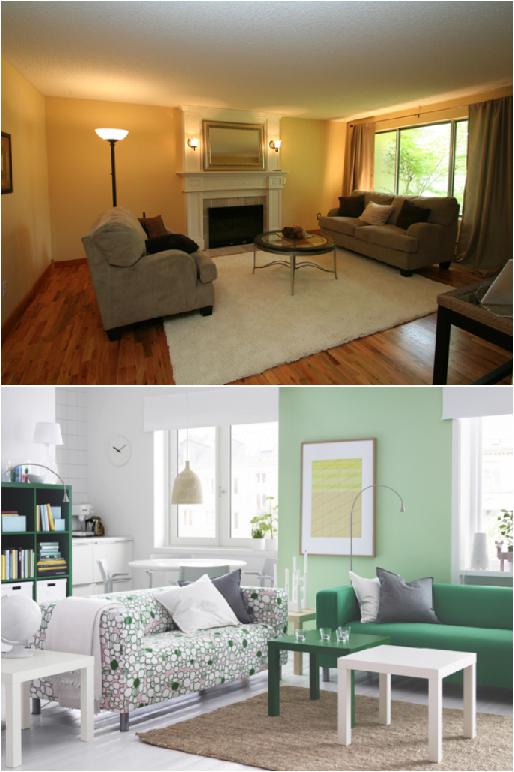}
    \end{subfigure}
    \begin{subfigure}[ht!]{0.218\linewidth}
        \centering
        \includegraphics[width=\linewidth]{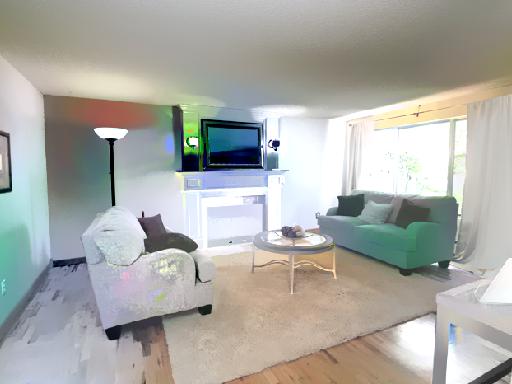}
    \end{subfigure}
    \begin{subfigure}[ht!]{0.218\linewidth}
        \centering
        \includegraphics[width=\linewidth]{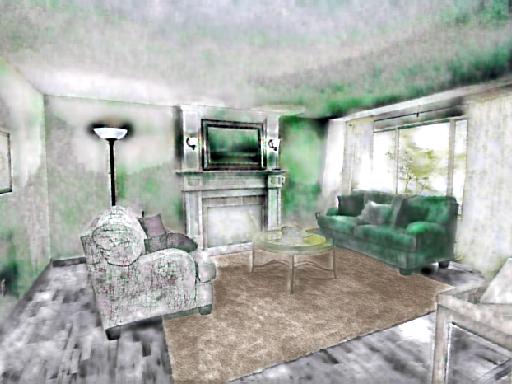}
    \end{subfigure}
    \begin{subfigure}[ht!]{0.218\linewidth}
        \centering
        \includegraphics[width=\linewidth]{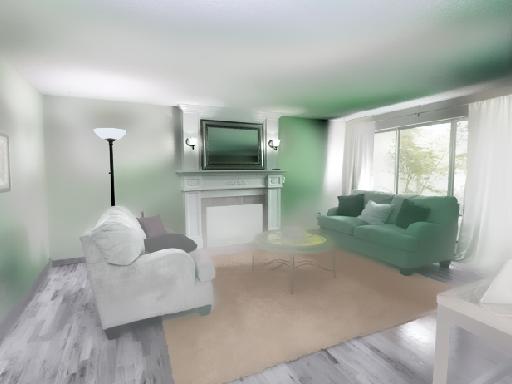}
    \end{subfigure}
    \begin{subfigure}[ht!]{0.218\linewidth}
        \centering
        \includegraphics[width=\linewidth]{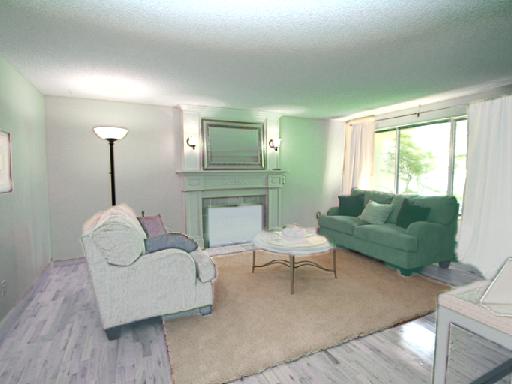}
    \end{subfigure}\\
    \begin{subfigure}[ht!]{0.109\linewidth}
        \centering
        \includegraphics[width=\linewidth]{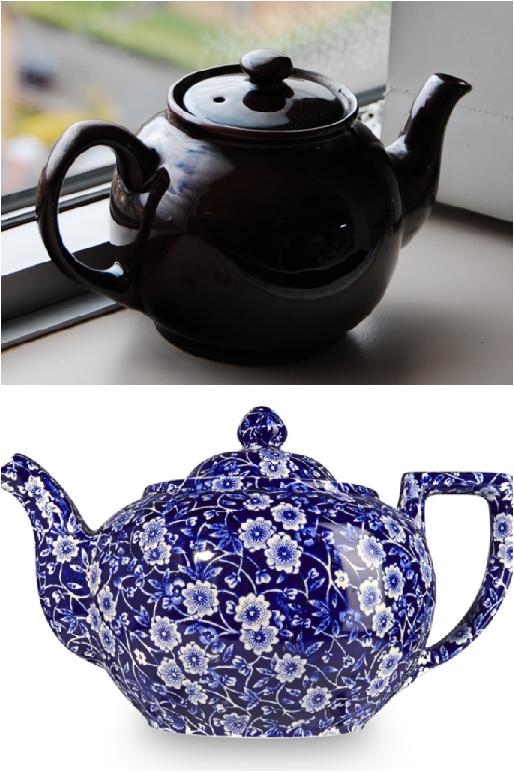}
    \end{subfigure}
    \begin{subfigure}[ht!]{0.218\linewidth}
        \centering
        \includegraphics[width=\linewidth]{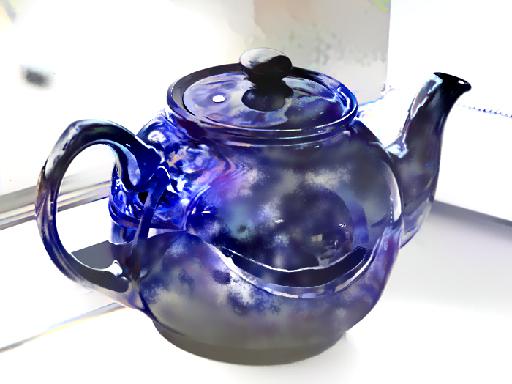}
    \end{subfigure}
    \begin{subfigure}[ht!]{0.218\linewidth}
        \centering
        \includegraphics[width=\linewidth]{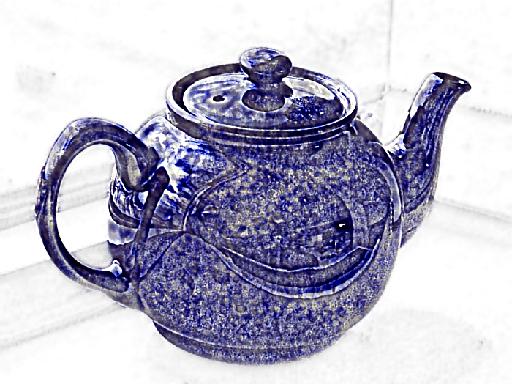}
    \end{subfigure}
    \begin{subfigure}[ht!]{0.218\linewidth}
        \centering
        \includegraphics[width=\linewidth]{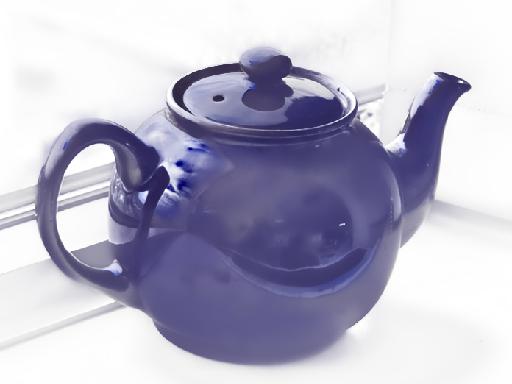}
    \end{subfigure}
    \begin{subfigure}[ht!]{0.218\linewidth}
        \centering
        \includegraphics[width=\linewidth]{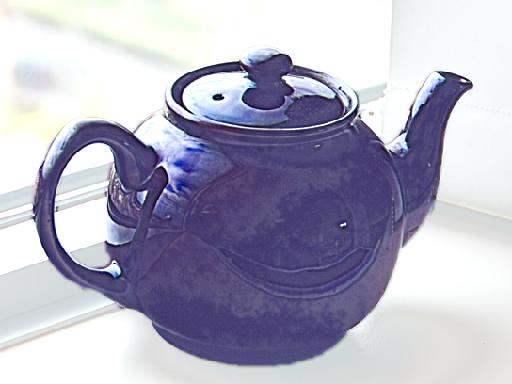}
    \end{subfigure}\\
    \begin{subfigure}[ht!]{0.109\linewidth}
        \centering
        \includegraphics[width=\linewidth]{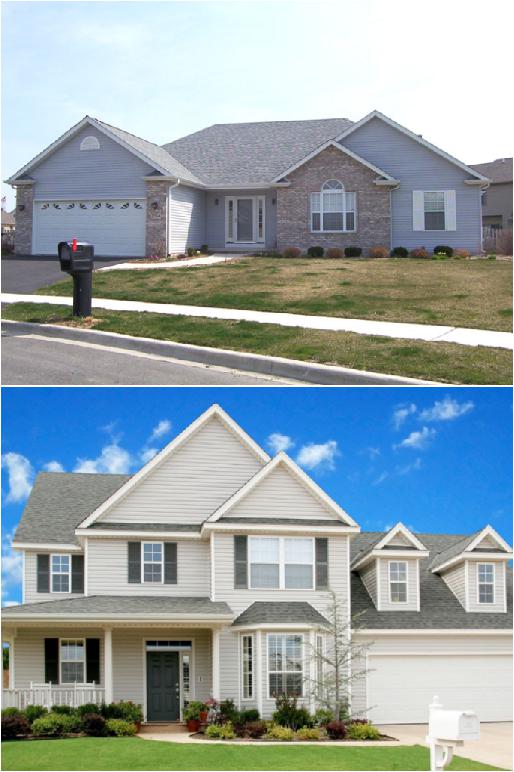}
    \end{subfigure}
    \begin{subfigure}[ht!]{0.218\linewidth}
        \centering
        \includegraphics[width=\linewidth]{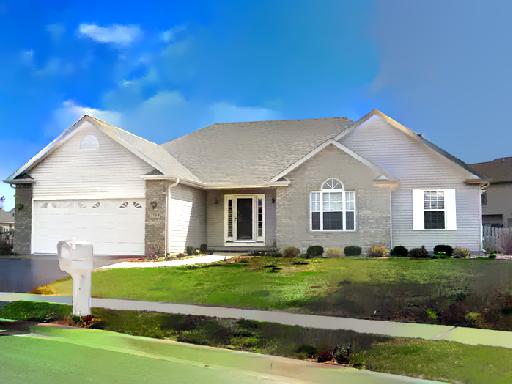}
    \end{subfigure}
    \begin{subfigure}[ht!]{0.218\linewidth}
        \centering
        \includegraphics[width=\linewidth]{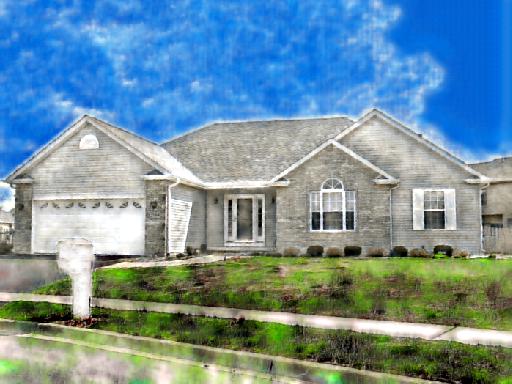}
    \end{subfigure}
    \begin{subfigure}[ht!]{0.218\linewidth}
        \centering
        \includegraphics[width=\linewidth]{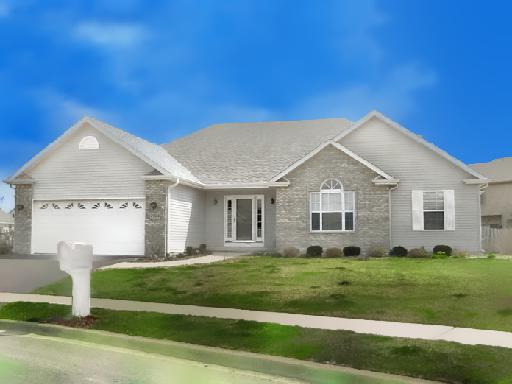}
    \end{subfigure}
    \begin{subfigure}[ht!]{0.218\linewidth}
        \centering
        \includegraphics[width=\linewidth]{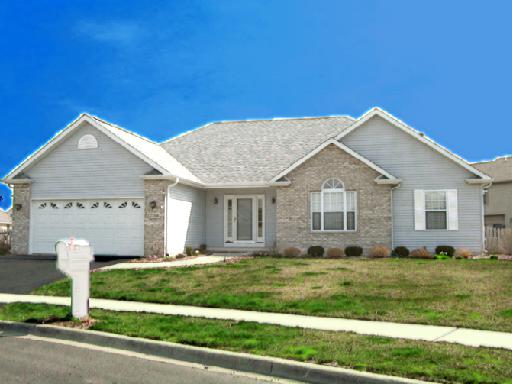}
    \end{subfigure}\\
    \begin{subfigure}[ht!]{0.109\linewidth}
        \centering
        \includegraphics[width=\linewidth]{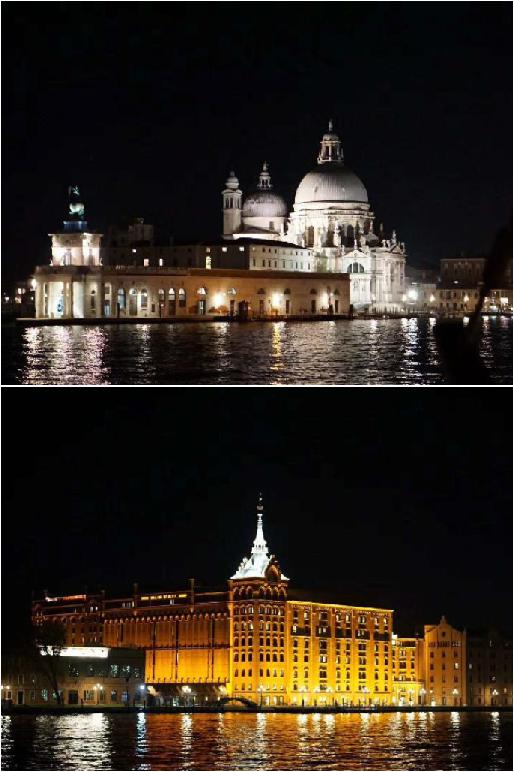}
        \caption{Input}
    \end{subfigure}
    \begin{subfigure}[ht!]{0.218\linewidth}
        \centering
        \includegraphics[width=\linewidth]{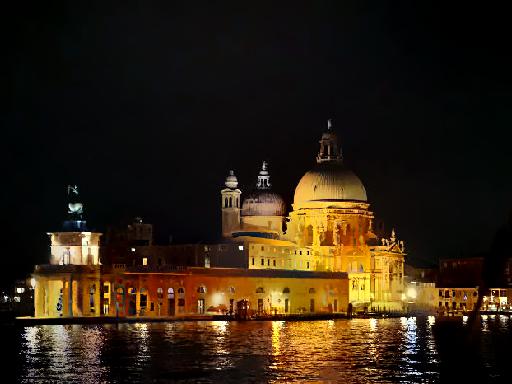}
        \caption{DPST \cite{luan2017deep}}
    \end{subfigure}
    \begin{subfigure}[ht!]{0.218\linewidth}
        \centering
        \includegraphics[width=\linewidth]{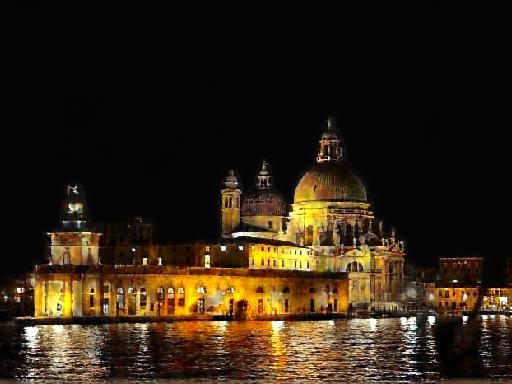}
        \caption{PhotoWCT \cite{photowct}}
    \end{subfigure}
    \begin{subfigure}[ht!]{0.218\linewidth}
        \centering
        \includegraphics[width=\linewidth]{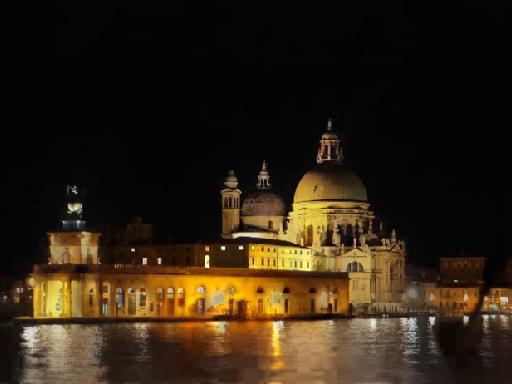}
        \caption{PhotoWCT (full) \cite{photowct}}
    \end{subfigure}
    \begin{subfigure}[ht!]{0.218\linewidth}
        \centering
        \includegraphics[width=\linewidth]{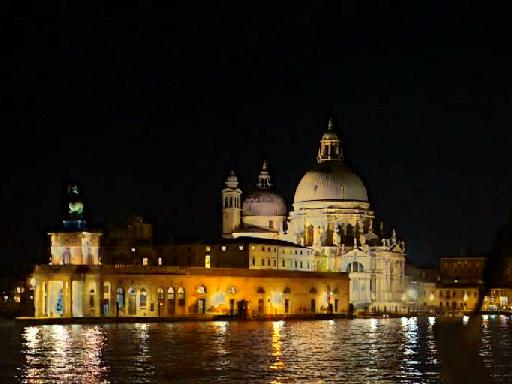}
        \caption{Ours (WCT$^2$)}
    \end{subfigure}
\caption{Photorealistic style transfer results. Given (a) an input pair (top: content, bottom: style), we compare the results of (b) deep photo style transfer (DPST) \cite{luan2017deep}, (c) and (d) PhotoWCT \cite{photowct} and (e) ours (WCT$^2$). (c) is the results of PhotoWCT without any post-processing and (d) shows the results after applying two post-processing steps proposed by the authors \cite{photowct}.}
\label{fig:photorealistic_comparison_suppl1}
\end{figure}

\begin{figure}[ht!]
    \centering
    \begin{subfigure}[ht!]{0.109\linewidth}
        \centering
        \includegraphics[width=\linewidth]{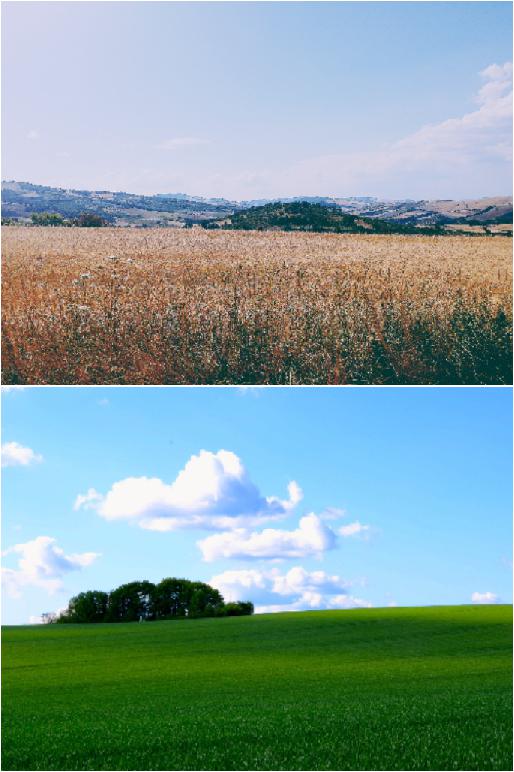}
    \end{subfigure}
    \begin{subfigure}[ht!]{0.218\linewidth}
        \centering
        \includegraphics[width=\linewidth]{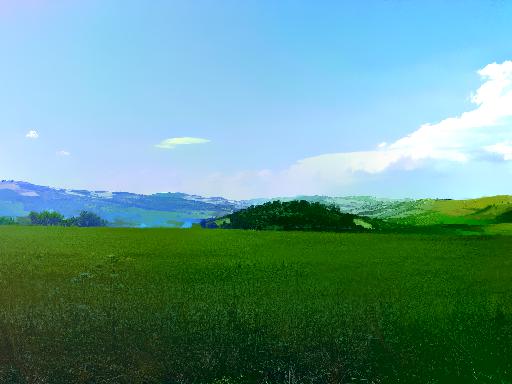}
    \end{subfigure}
    \begin{subfigure}[ht!]{0.218\linewidth}
        \centering
        \includegraphics[width=\linewidth]{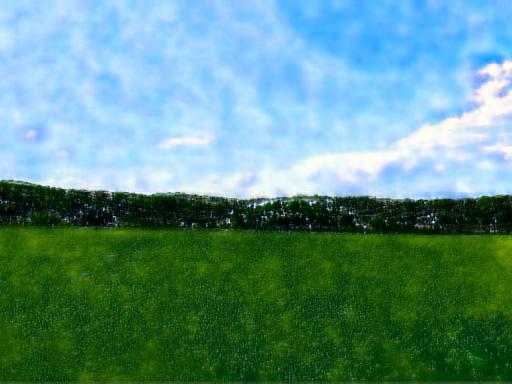}
    \end{subfigure}
    \begin{subfigure}[ht!]{0.218\linewidth}
        \centering
        \includegraphics[width=\linewidth]{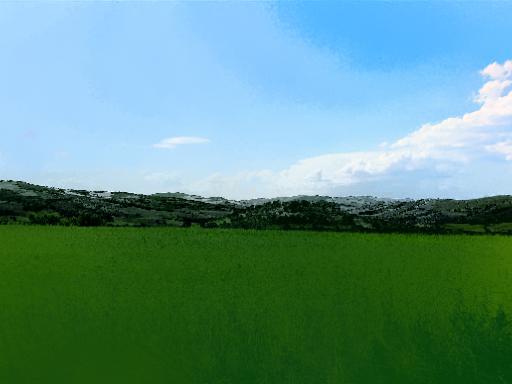}
    \end{subfigure}
    \begin{subfigure}[ht!]{0.218\linewidth}
        \centering
        \includegraphics[width=\linewidth]{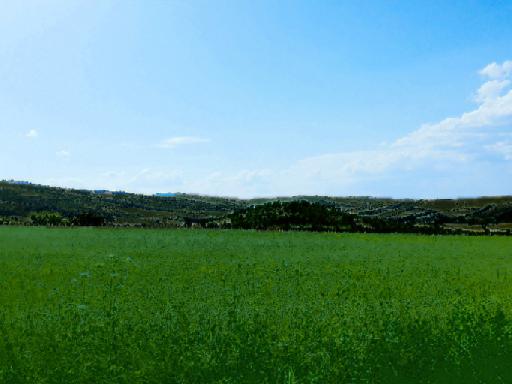}
    \end{subfigure}\\
    \begin{subfigure}[ht!]{0.109\linewidth}
        \centering
        \includegraphics[width=\linewidth]{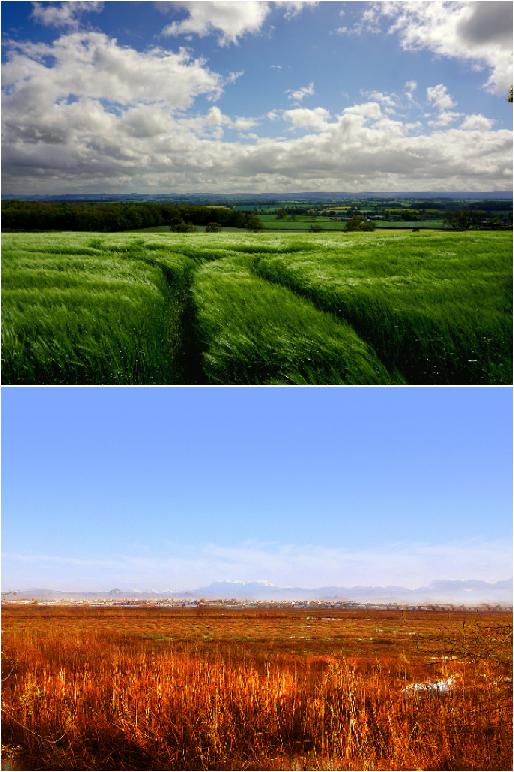}
    \end{subfigure}
    \begin{subfigure}[ht!]{0.218\linewidth}
        \centering
        \includegraphics[width=\linewidth]{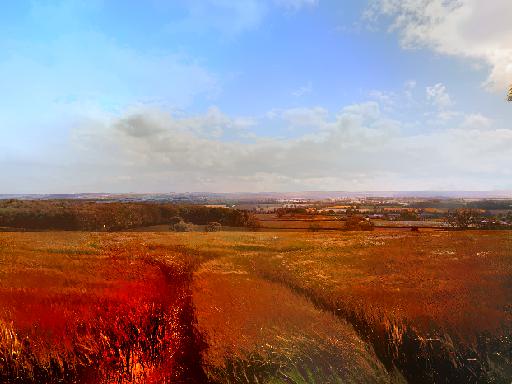}
    \end{subfigure}
    \begin{subfigure}[ht!]{0.218\linewidth}
        \centering
        \includegraphics[width=\linewidth]{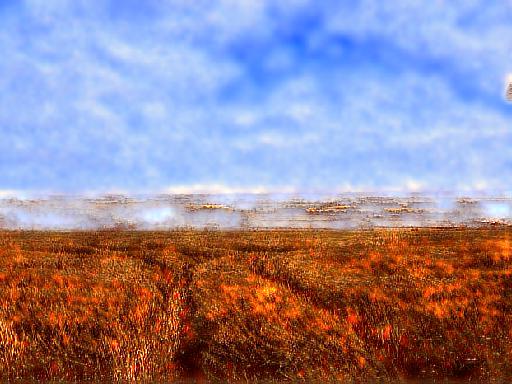}
    \end{subfigure}
    \begin{subfigure}[ht!]{0.218\linewidth}
        \centering
        \includegraphics[width=\linewidth]{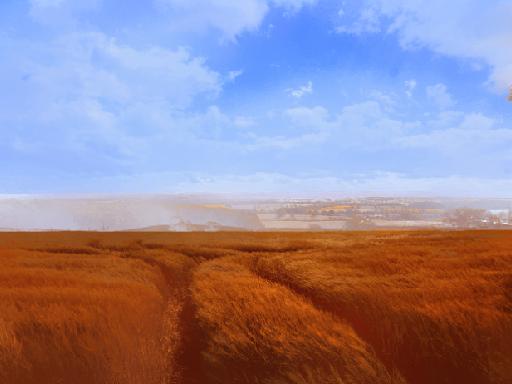}
    \end{subfigure}
    \begin{subfigure}[ht!]{0.218\linewidth}
        \centering
        \includegraphics[width=\linewidth]{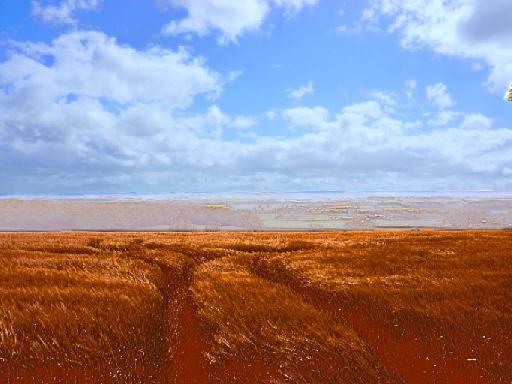}
    \end{subfigure}\\
    \begin{subfigure}[ht!]{0.109\linewidth}
        \centering
        \includegraphics[width=\linewidth]{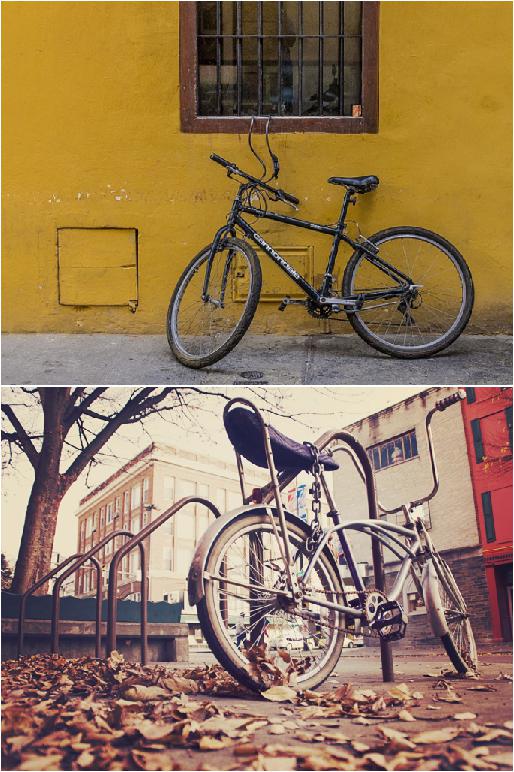}
    \end{subfigure}
    \begin{subfigure}[ht!]{0.218\linewidth}
        \centering
        \includegraphics[width=\linewidth]{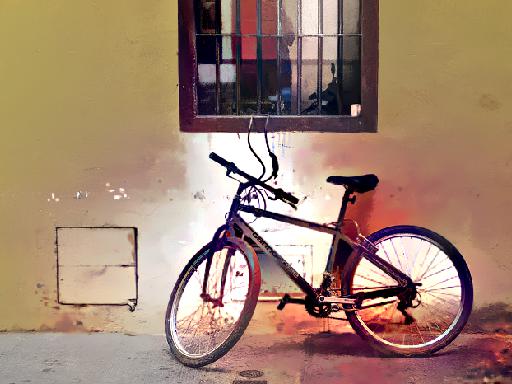}
    \end{subfigure}
    \begin{subfigure}[ht!]{0.218\linewidth}
        \centering
        \includegraphics[width=\linewidth]{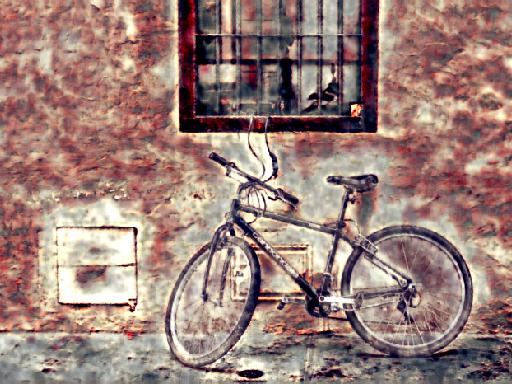}
    \end{subfigure}
    \begin{subfigure}[ht!]{0.218\linewidth}
        \centering
        \includegraphics[width=\linewidth]{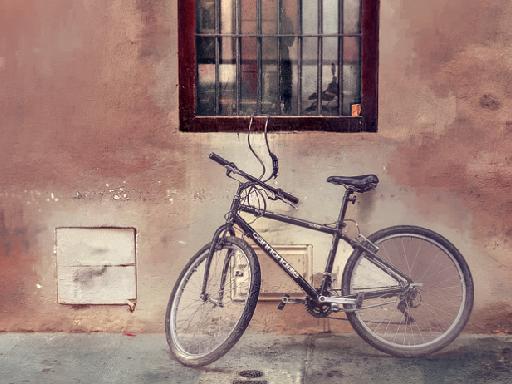}
    \end{subfigure}
    \begin{subfigure}[ht!]{0.218\linewidth}
        \centering
        \includegraphics[width=\linewidth]{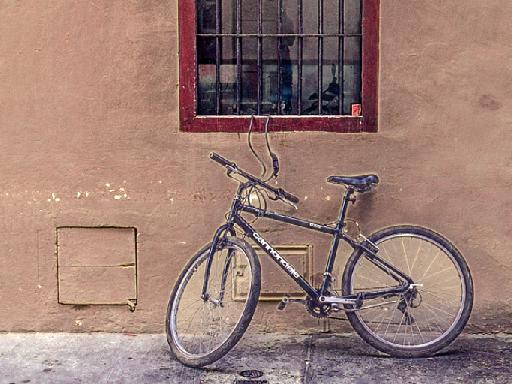}
    \end{subfigure}\\
    \begin{subfigure}[ht!]{0.109\linewidth}
        \centering
        \includegraphics[width=\linewidth]{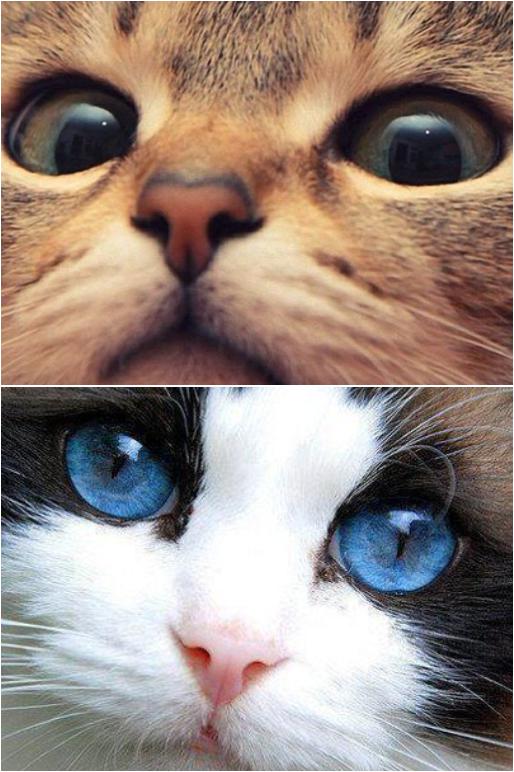}
    \end{subfigure}
    \begin{subfigure}[ht!]{0.218\linewidth}
        \centering
        \includegraphics[width=\linewidth]{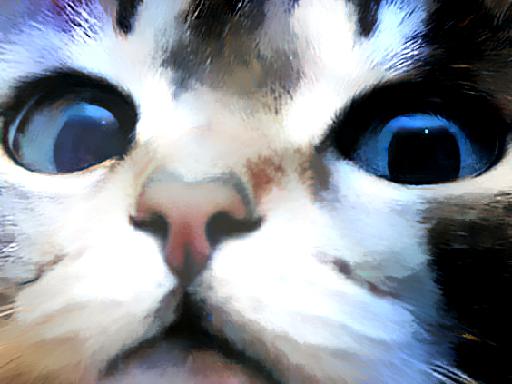}
    \end{subfigure}
    \begin{subfigure}[ht!]{0.218\linewidth}
        \centering
        \includegraphics[width=\linewidth]{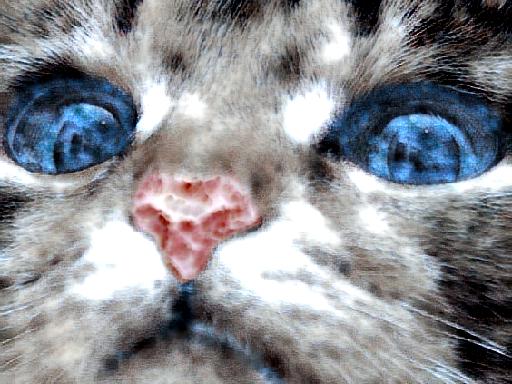}
    \end{subfigure}
    \begin{subfigure}[ht!]{0.218\linewidth}
        \centering
        \includegraphics[width=\linewidth]{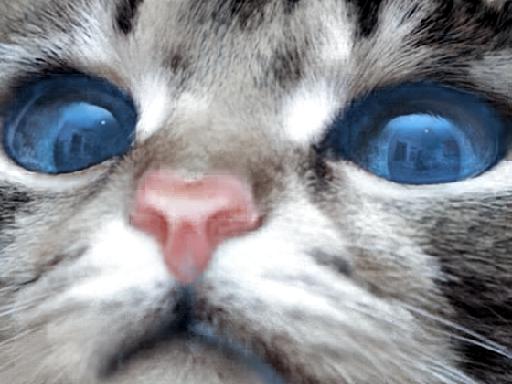}
    \end{subfigure}
    \begin{subfigure}[ht!]{0.218\linewidth}
        \centering
        \includegraphics[width=\linewidth]{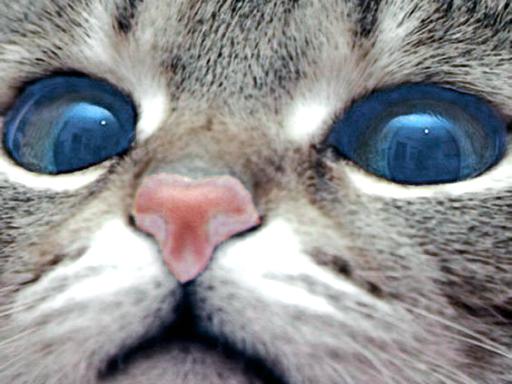}
    \end{subfigure}\\
    \begin{subfigure}[ht!]{0.109\linewidth}
        \centering
        \includegraphics[width=\linewidth]{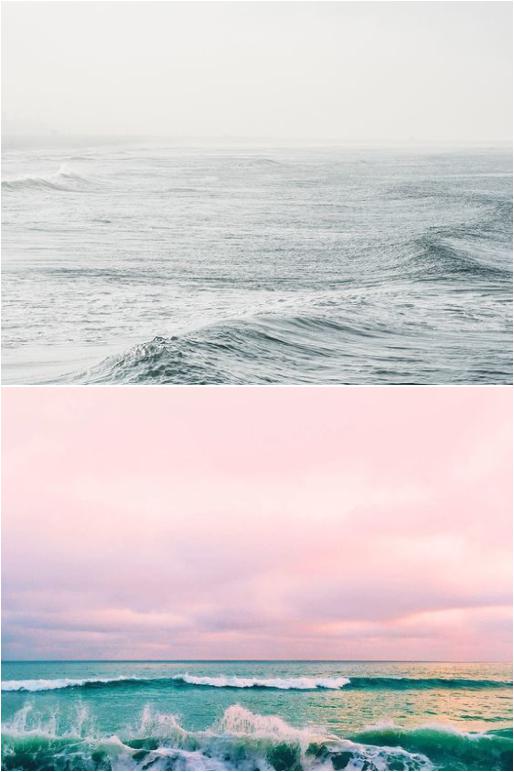}
    \end{subfigure}
    \begin{subfigure}[ht!]{0.218\linewidth}
        \centering
        \includegraphics[width=\linewidth]{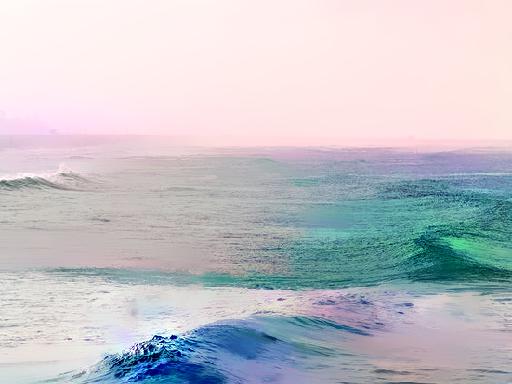}
    \end{subfigure}
    \begin{subfigure}[ht!]{0.218\linewidth}
        \centering
        \includegraphics[width=\linewidth]{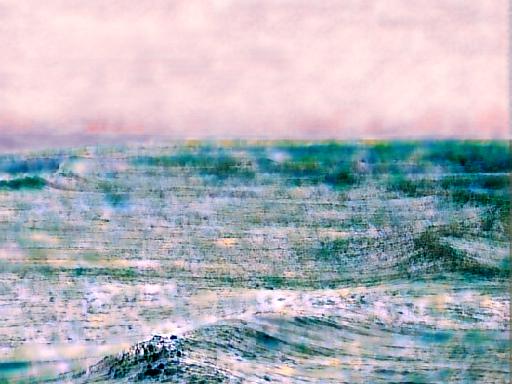}
    \end{subfigure}
    \begin{subfigure}[ht!]{0.218\linewidth}
        \centering
        \includegraphics[width=\linewidth]{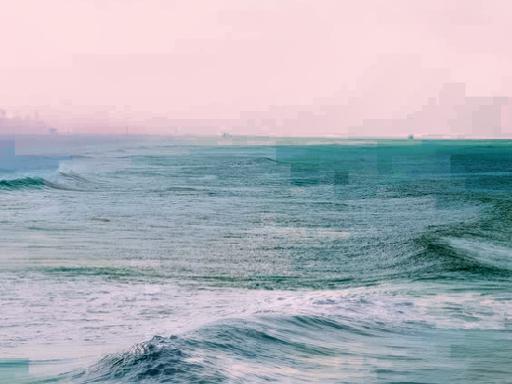}
    \end{subfigure}
    \begin{subfigure}[ht!]{0.218\linewidth}
        \centering
        \includegraphics[width=\linewidth]{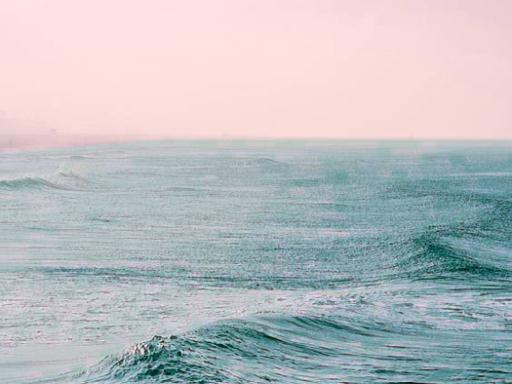}
    \end{subfigure}\\
    \begin{subfigure}[ht!]{0.109\linewidth}
        \centering
        \includegraphics[width=\linewidth]{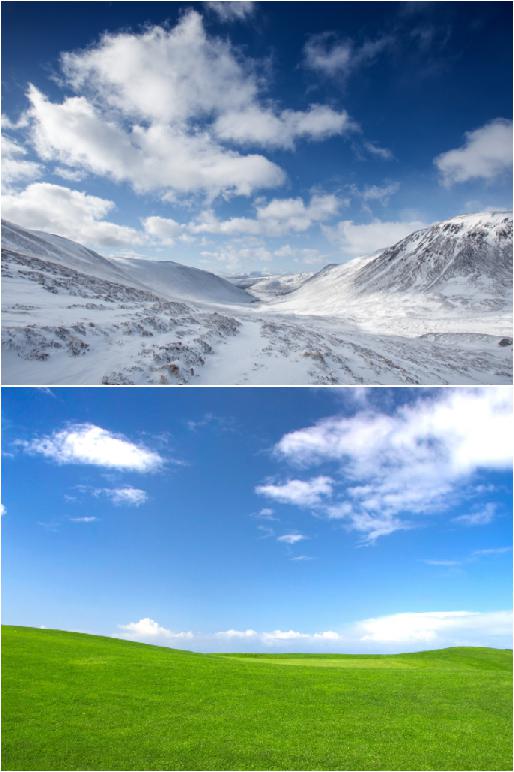}
        \caption{Input}
    \end{subfigure}
    \begin{subfigure}[ht!]{0.218\linewidth}
        \centering
        \includegraphics[width=\linewidth]{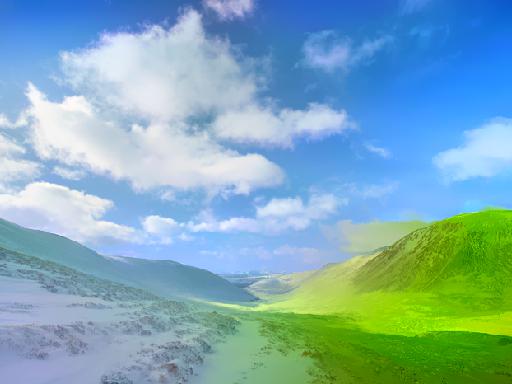}
        \caption{DPST \cite{luan2017deep}}
    \end{subfigure}
    \begin{subfigure}[ht!]{0.218\linewidth}
        \centering
        \includegraphics[width=\linewidth]{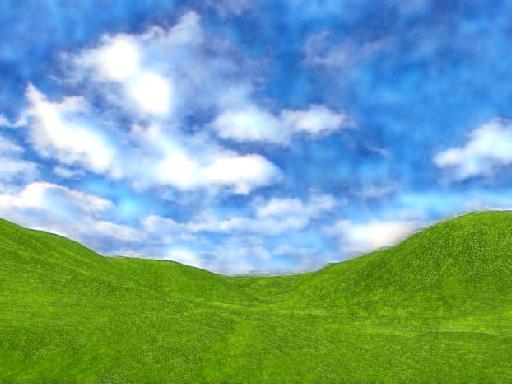}
        \caption{PhotoWCT \cite{photowct}}
    \end{subfigure}
    \begin{subfigure}[ht!]{0.218\linewidth}
        \centering
        \includegraphics[width=\linewidth]{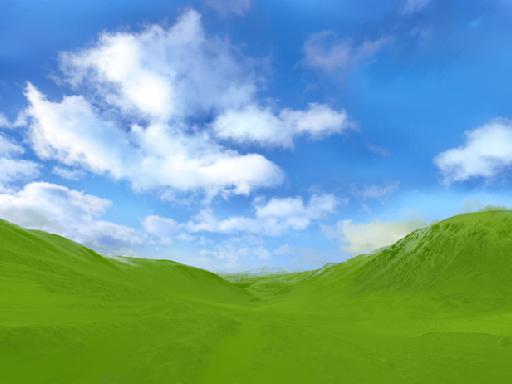}
        \caption{PhotoWCT (full) \cite{photowct}}
    \end{subfigure}
    \begin{subfigure}[ht!]{0.218\linewidth}
        \centering
        \includegraphics[width=\linewidth]{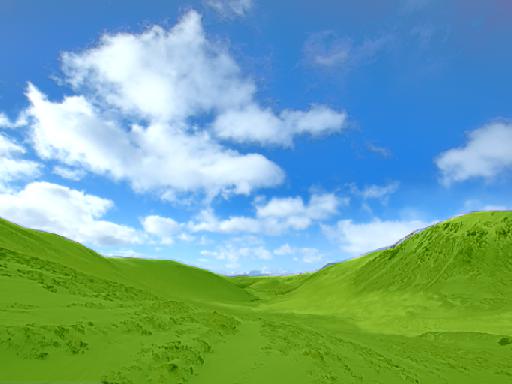}
        \caption{Ours (WCT$^2$)}
    \end{subfigure}%\\

% \begin{subfigure}[ht!]{0.109\linewidth}
%         \centering
%         \includegraphics[width=\linewidth]{./images/suppl/comparison_suppl/06_input.jpg}
%         \caption{Input}
%     \end{subfigure}
%     \begin{subfigure}[ht!]{0.218\linewidth}
%         \centering
%         \includegraphics[width=\linewidth]{./images/suppl/comparison_suppl/06_dps.jpg}
%         \caption{DPST \cite{luan2017deep}}
%     \end{subfigure}
%     \begin{subfigure}[ht!]{0.218\linewidth}
%         \centering
%         \includegraphics[width=\linewidth]{./images/suppl/comparison_suppl/06_photowct_basic.jpg}
%         \caption{PhotoWCT \cite{photowct}}
%     \end{subfigure}
%     \begin{subfigure}[ht!]{0.218\linewidth}
%         \centering
%         \includegraphics[width=\linewidth]{./images/suppl/comparison_suppl/06_photowct.jpg}
%         \caption{PhotoWCT (full) \cite{photowct}}
%     \end{subfigure}
%     \begin{subfigure}[ht!]{0.218\linewidth}
%         \centering
%         \includegraphics[width=\linewidth]{./images/suppl/comparison_suppl/06_ours.jpg}
%         \caption{Ours (WCT$^2$)}
%     \end{subfigure}
\caption{Photorealistic style transfer results. Given (a) an input pair (top: content, bottom: style), we compare the results of (b) deep photo style transfer (DPST) \cite{luan2017deep}, (c) and (d) PhotoWCT \cite{photowct} and (e) ours (WCT$^2$). (c) is the results of PhotoWCT without any post-processing and (d) shows the results after applying two post-processing steps proposed by the authors \cite{photowct}.}
\label{fig:photorealistic_comparison_suppl2}
\end{figure}

\begin{figure}[ht!]
    \centering
    \begin{subfigure}[ht!]{0.109\linewidth}
        \centering
        \includegraphics[width=\linewidth]{./images/comparison/00_input2.jpg}
    \end{subfigure}
    \begin{subfigure}[ht!]{0.218\linewidth}
        \centering
        \includegraphics[width=\linewidth]{./images/comparison/00_ours.jpg}
    \end{subfigure}
    \begin{subfigure}[ht!]{0.218\linewidth}
        \centering
        \includegraphics[width=\linewidth, height=0.75\linewidth]{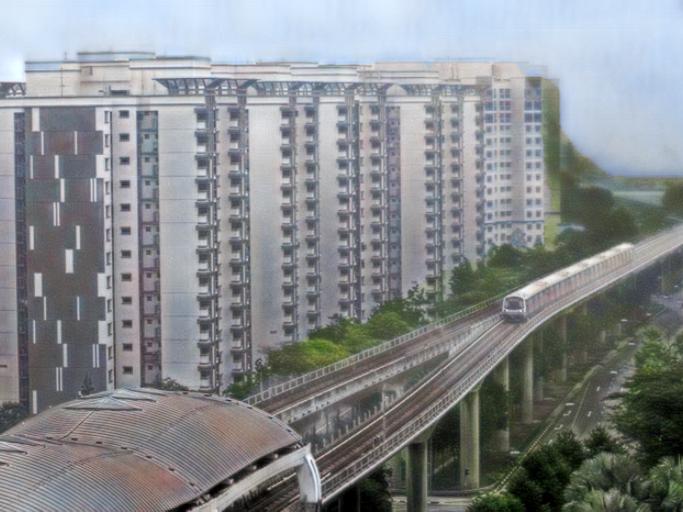}
    \end{subfigure}
    \begin{subfigure}[ht!]{0.218\linewidth}
        \centering
        \includegraphics[width=\linewidth, height=0.75\linewidth]{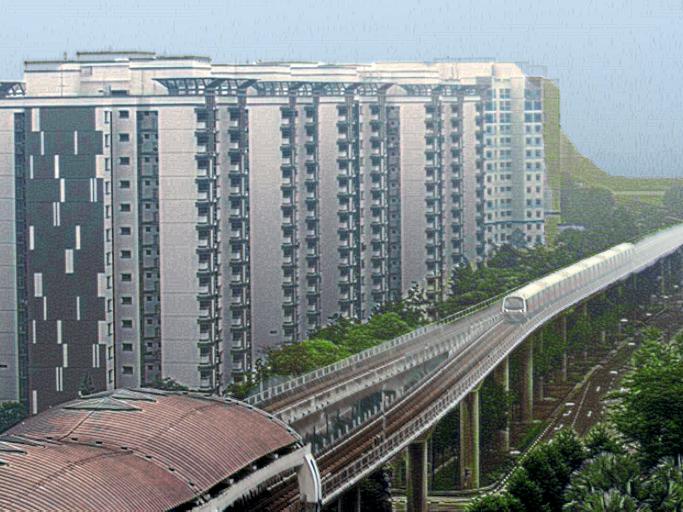}
    \end{subfigure}
    \begin{subfigure}[ht!]{0.218\linewidth}
        \centering
        \includegraphics[width=\linewidth, height=0.75\linewidth]{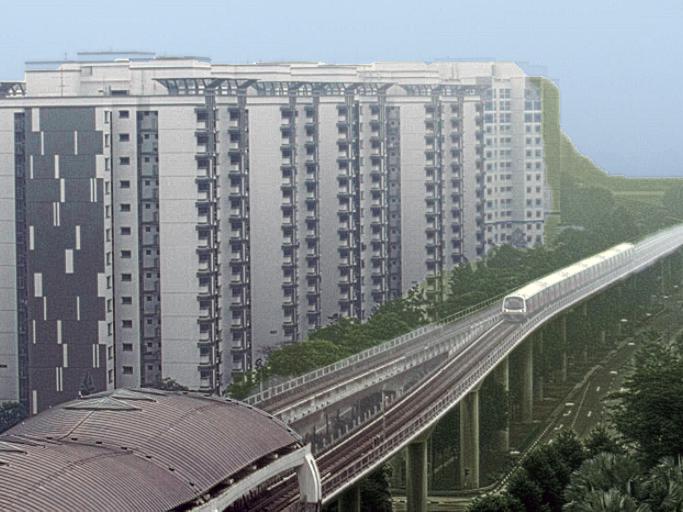}
    \end{subfigure}\\
    \begin{subfigure}[ht!]{0.109\linewidth}
        \centering
        \includegraphics[width=\linewidth]{./images/comparison/01_input2.jpg}
    \end{subfigure}
    \begin{subfigure}[ht!]{0.218\linewidth}
        \centering
        \includegraphics[width=\linewidth]{./images/comparison/01_ours.jpg}
    \end{subfigure}
    \begin{subfigure}[ht!]{0.218\linewidth}
        \centering
        \includegraphics[width=\linewidth, height=0.75\linewidth]{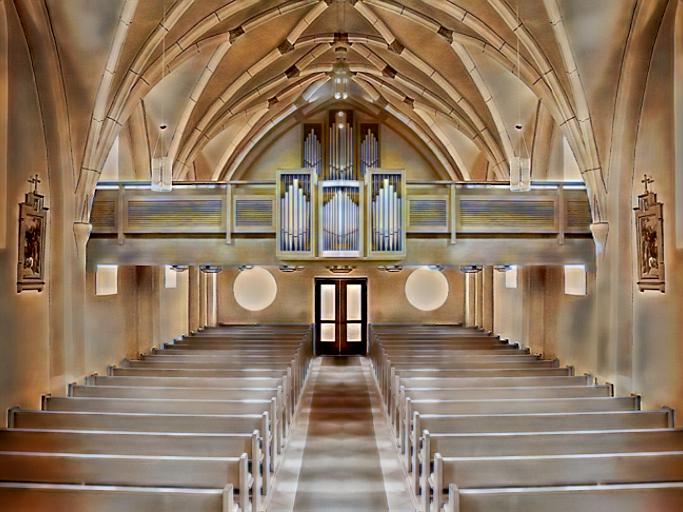}
    \end{subfigure}
    \begin{subfigure}[ht!]{0.218\linewidth}
        \centering
        \includegraphics[width=\linewidth, height=0.75\linewidth]{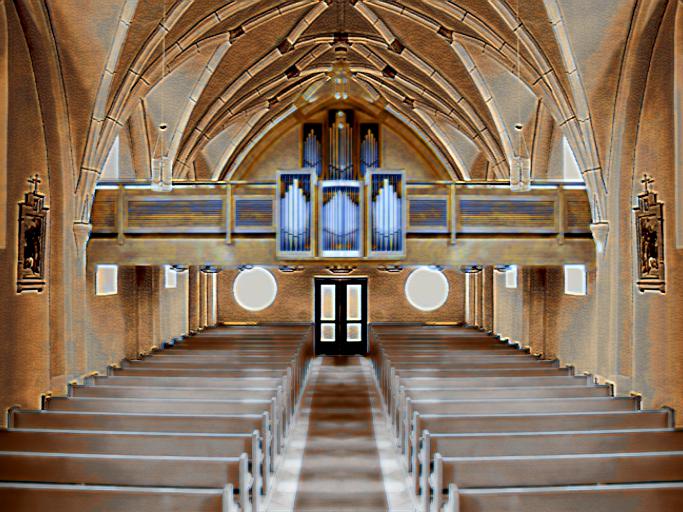}
    \end{subfigure}
    \begin{subfigure}[ht!]{0.218\linewidth}
        \centering
        \includegraphics[width=\linewidth, height=0.75\linewidth]{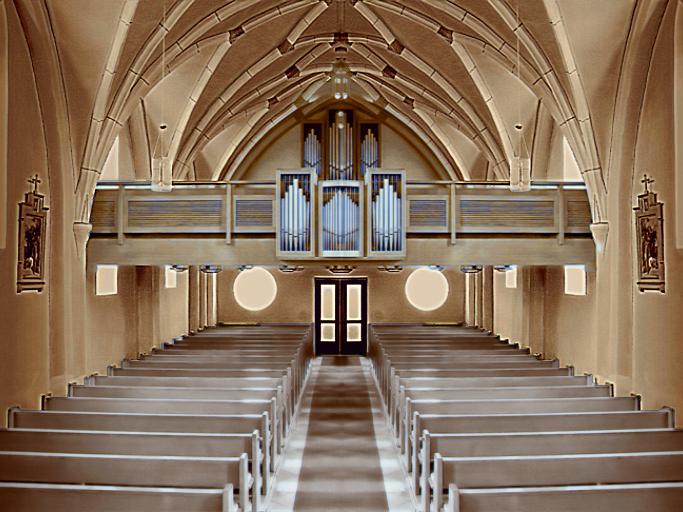}
    \end{subfigure}\\
    \begin{subfigure}[ht!]{0.109\linewidth}
        \centering
        \includegraphics[width=\linewidth]{./images/comparison/02_input2.jpg}
    \end{subfigure}
    \begin{subfigure}[ht!]{0.218\linewidth}
        \centering
        \includegraphics[width=\linewidth]{./images/comparison/02_ours.jpg}
    \end{subfigure}
    \begin{subfigure}[ht!]{0.218\linewidth}
        \centering
        \includegraphics[width=\linewidth, height=0.75\linewidth]{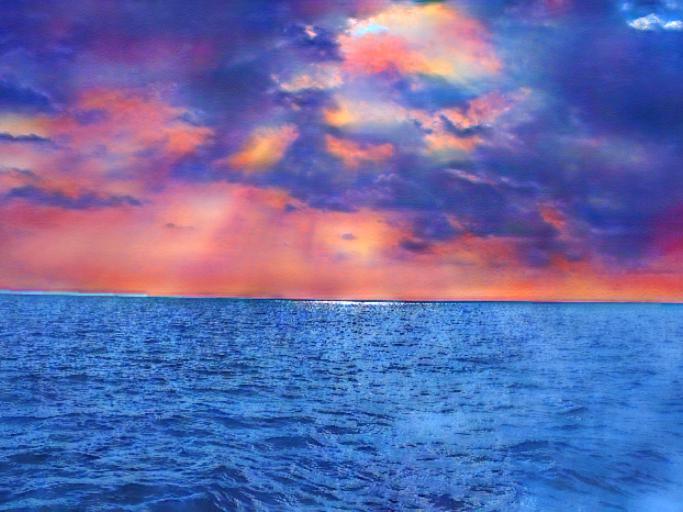}
    \end{subfigure}
    \begin{subfigure}[ht!]{0.218\linewidth}
        \centering
        \includegraphics[width=\linewidth, height=0.75\linewidth]{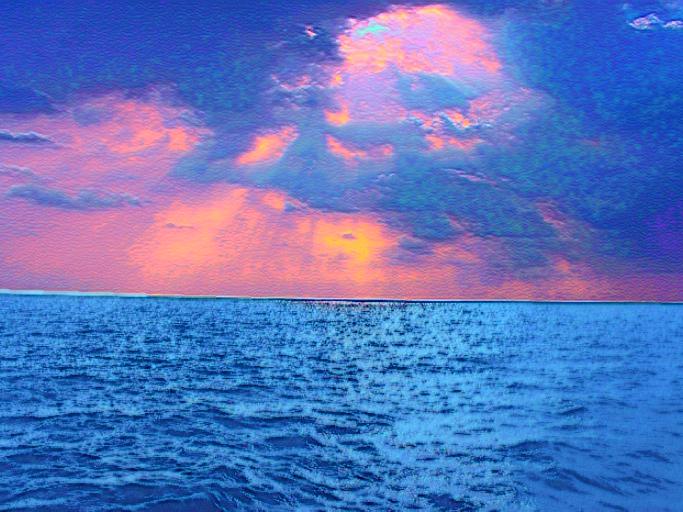}
    \end{subfigure}
    \begin{subfigure}[ht!]{0.218\linewidth}
        \centering
        \includegraphics[width=\linewidth, height=0.75\linewidth]{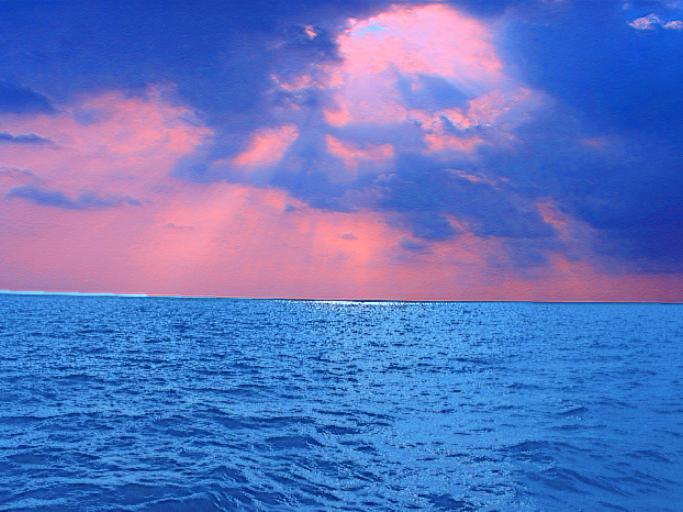}
    \end{subfigure}\\
    \begin{subfigure}[ht!]{0.109\linewidth}
        \centering
        \includegraphics[width=\linewidth]{./images/comparison/03_input2.jpg}
    \end{subfigure}
    \begin{subfigure}[ht!]{0.218\linewidth}
        \centering
        \includegraphics[width=\linewidth]{./images/comparison/03_ours.jpg}
    \end{subfigure}
    \begin{subfigure}[ht!]{0.218\linewidth}
        \centering
        \includegraphics[width=\linewidth, height=0.75\linewidth]{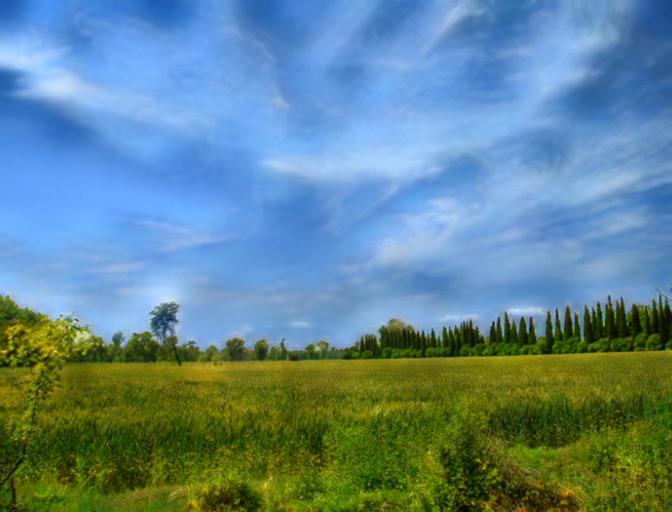}
    \end{subfigure}
    \begin{subfigure}[ht!]{0.218\linewidth}
        \centering
        \includegraphics[width=\linewidth, height=0.75\linewidth]{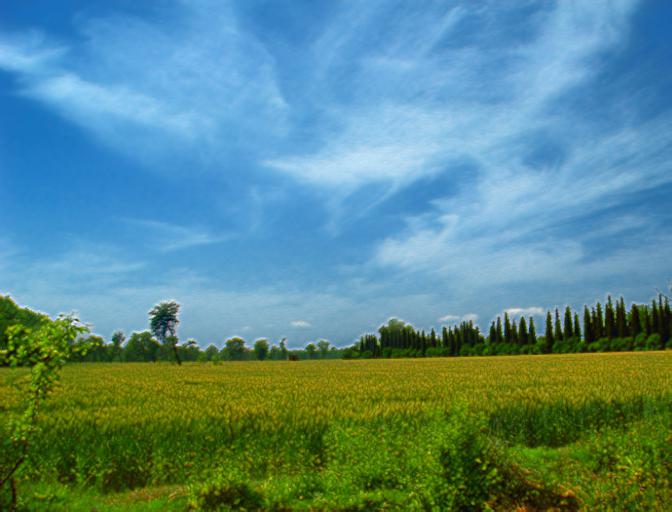}
    \end{subfigure}
    \begin{subfigure}[ht!]{0.218\linewidth}
        \centering
        \includegraphics[width=\linewidth, height=0.75\linewidth]{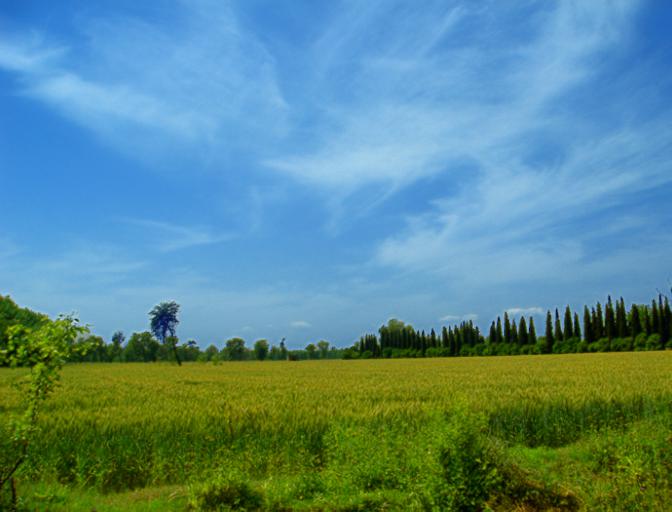}
    \end{subfigure}\\
    \begin{subfigure}[ht!]{0.109\linewidth}
        \centering
        \includegraphics[width=\linewidth]{./images/comparison/04_input2.jpg}
    \end{subfigure}
    \begin{subfigure}[ht!]{0.218\linewidth}
        \centering
        \includegraphics[width=\linewidth]{./images/comparison/04_ours.jpg}
    \end{subfigure}
    \begin{subfigure}[ht!]{0.218\linewidth}
        \centering
        \includegraphics[width=\linewidth, height=0.75\linewidth]{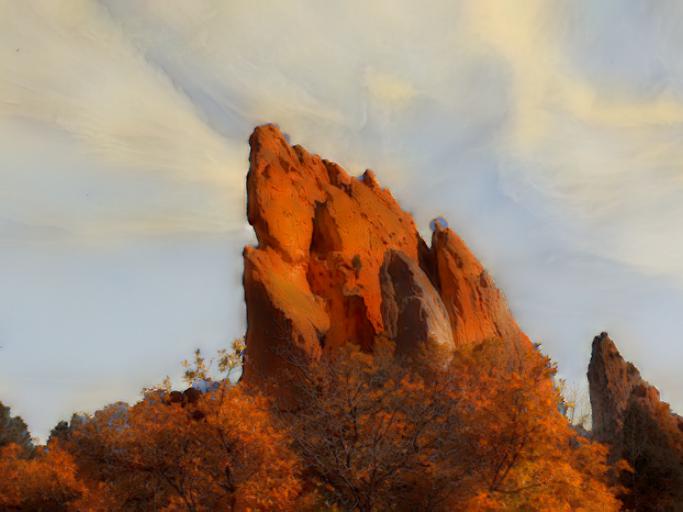}
    \end{subfigure}
    \begin{subfigure}[ht!]{0.218\linewidth}
        \centering
        \includegraphics[width=\linewidth, height=0.75\linewidth]{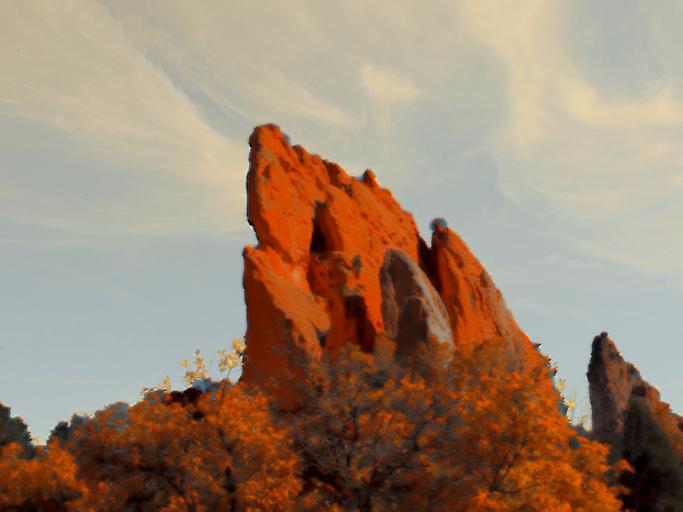}
    \end{subfigure}
    \begin{subfigure}[ht!]{0.218\linewidth}
        \centering
        \includegraphics[width=\linewidth, height=0.75\linewidth]{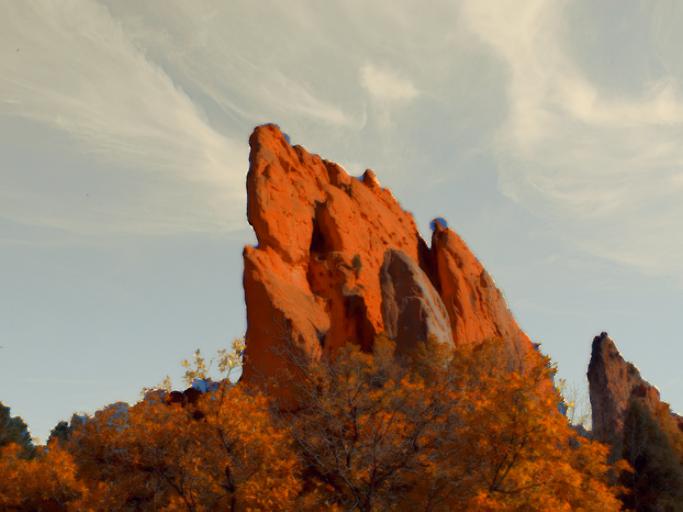}
    \end{subfigure}\\
    \begin{subfigure}[ht!]{0.109\linewidth}
        \centering
        \includegraphics[width=\linewidth]{./images/comparison/05_input2.jpg}
    \end{subfigure}
    \begin{subfigure}[ht!]{0.218\linewidth}
        \centering
        \includegraphics[width=\linewidth]{./images/comparison/05_ours.jpg}
    \end{subfigure}
    \begin{subfigure}[ht!]{0.218\linewidth}
        \centering
        \includegraphics[width=\linewidth, height=0.75\linewidth]{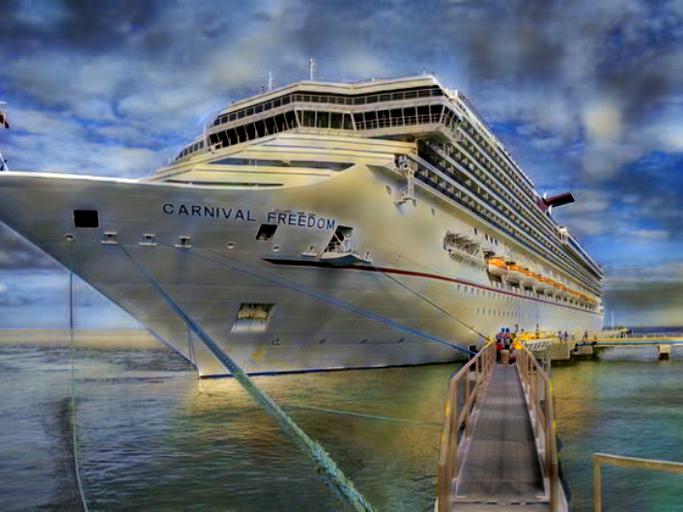}
    \end{subfigure}
    \begin{subfigure}[ht!]{0.218\linewidth}
        \centering
        \includegraphics[width=\linewidth, height=0.75\linewidth]{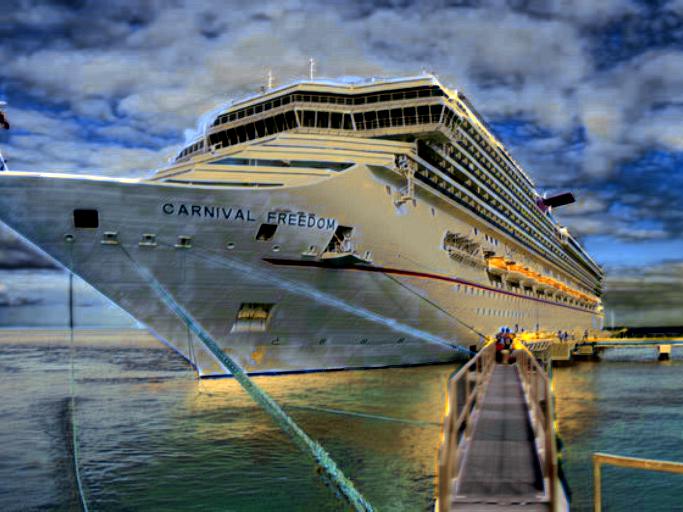}
    \end{subfigure}
    \begin{subfigure}[ht!]{0.218\linewidth}
        \centering
        \includegraphics[width=\linewidth, height=0.75\linewidth]{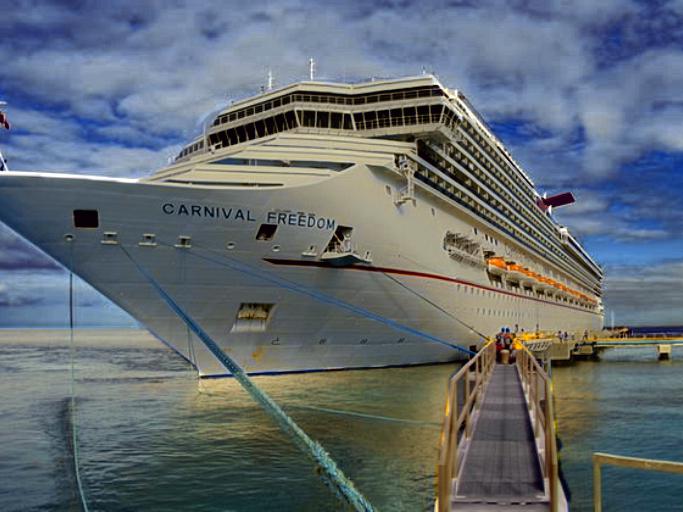}
    \end{subfigure}\\
\begin{subfigure}[ht!]{0.109\linewidth}
        \centering
        \includegraphics[width=\linewidth]{./images/comparison/06_input2.jpg}
        \caption{Input}
    \end{subfigure}
    \begin{subfigure}[ht!]{0.218\linewidth}
        \centering
        \includegraphics[width=\linewidth]{./images/comparison/06_ours.jpg}
        \caption{WCT$^2$}
    \end{subfigure}
    \begin{subfigure}[ht!]{0.218\linewidth}
        \centering
        \includegraphics[width=\linewidth, height=0.75\linewidth]{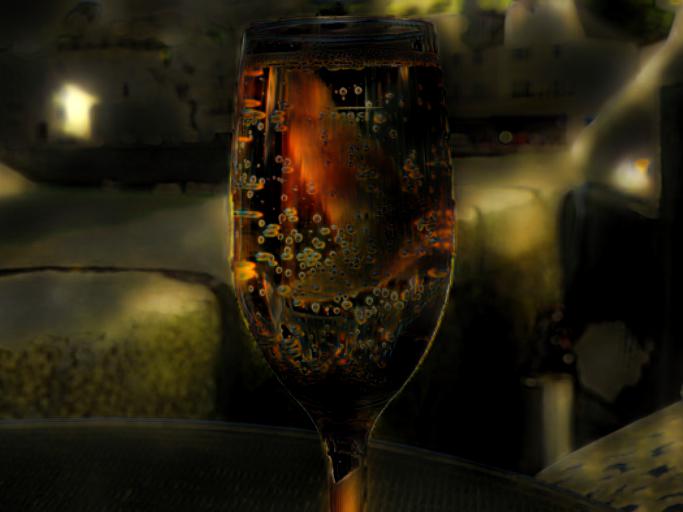}
        \caption{WCT$^2$ (sum) \cite{photowct}}
    \end{subfigure}
    \begin{subfigure}[ht!]{0.218\linewidth}
        \centering
        \includegraphics[width=\linewidth, height=0.75\linewidth]{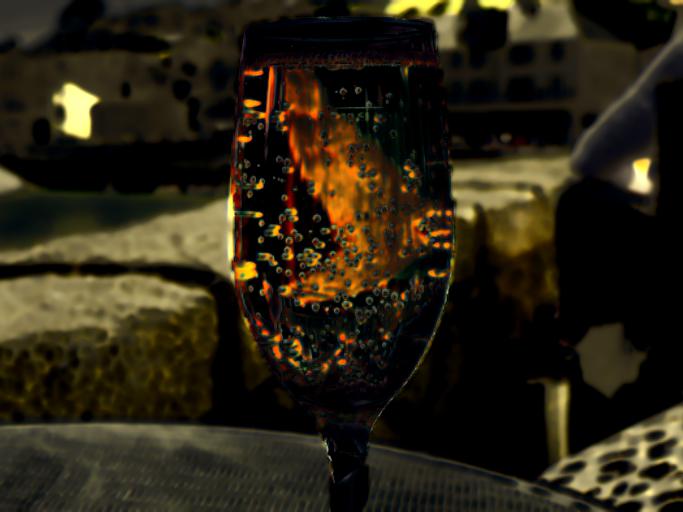}
        \caption{WCT$^2$ (+multi-level) \cite{photowct}}
    \end{subfigure}
    \begin{subfigure}[ht!]{0.218\linewidth}
        \centering
        \includegraphics[width=\linewidth, height=0.75\linewidth]{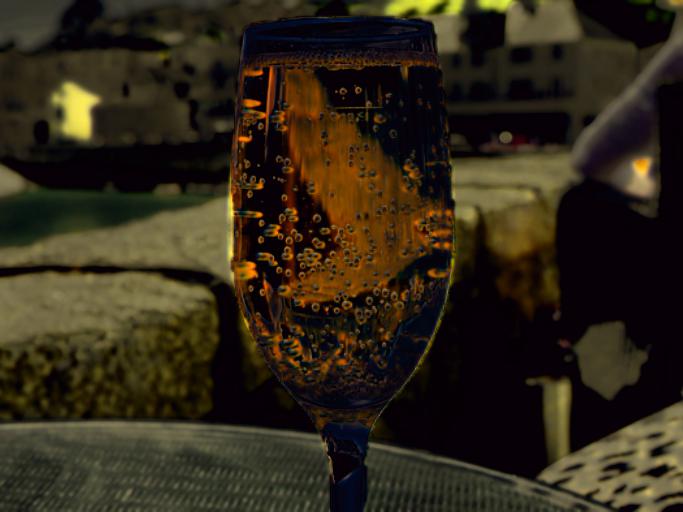}
        \caption{WCT$^2$ (+decoder)}
    \end{subfigure}
\caption{Photorealistic style transfer results. Given (a) an input pair (top: content, bottom: style), we compare the results of WCT$^2$ and its variants, \ie, (b) WCT$^2$, (c) WCT$^2$ (sum) (d) WCT$^2$ (+multi-level) and (e) WCT$^2$ (+decoder). }
\label{fig:photorealistic_comparison_suppl3}
\end{figure}